\documentclass[sigconf]{acmart}
\usepackage[ruled,linesnumbered,vlined]{algorithm2e}
\usepackage{enumitem}
\usepackage{amsmath}
\usepackage{ragged2e}
\usepackage{newfloat}
\usepackage{listings}
\usepackage{bm}
\usepackage{booktabs}
\usepackage{array}
\usepackage{multirow}
\usepackage{color}
\usepackage{soul}
\usepackage{makecell}
\usepackage{bbm}
\usepackage{subcaption}

\AtBeginDocument{%
  }

\copyrightyear{2025} 
\acmYear{2025} 
\setcopyright{cc}
\setcctype{by}
\acmConference[KDD '25]{Proceedings of the 31st ACM SIGKDD Conference on Knowledge Discovery and Data Mining V.2}{August 3--7, 2025}{Toronto, ON, Canada}
\acmBooktitle{Proceedings of the 31st ACM SIGKDD Conference on Knowledge Discovery and Data Mining V.2 (KDD '25), August 3--7, 2025, Toronto, ON, Canada}
\acmDOI{10.1145/3711896.3737140}
\acmISBN{979-8-4007-1454-2/2025/08}

\begin{document}

\title{SurvUnc: A Meta-Model Based Uncertainty Quantification Framework for Survival Analysis}

\author{Yu Liu}\authornote{Corresponding author.}
\email{yu.liu@eng.ox.ac.uk}
\affiliation{
	\institution{University of Oxford}
	\city{Oxford}
	\country{UK}
}

\author{Weiyao Tao}
\email{twy030819@gmail.com}
\affiliation{
	\institution{University of Oxford}
	\city{Oxford}
	\country{UK}
}

\author{Tong Xia}
\email{tx229@cam.ac.uk}
\affiliation{
	\institution{University of Cambridge}
	\city{Cambridge}
	\country{UK}
}

\author{Simon Knight}
\email{simon.knight@nds.ox.ac.uk}
\affiliation{
	\institution{University of Oxford}
	\city{Oxford}
	\country{UK}
}

\author{Tingting Zhu}
\email{tingting.zhu@eng.ox.ac.uk}
\affiliation{
	\institution{University of Oxford}
	\city{Oxford}
	\country{UK}
}

\begin{abstract}
Survival analysis, which estimates the probability of event occurrence over time from censored data, is fundamental in numerous real-world applications, particularly in high-stakes domains such as healthcare and risk assessment. Despite advances in numerous survival models, quantifying the uncertainty of predictions from these models remains underexplored and challenging. The lack of reliable uncertainty quantification limits the interpretability and trustworthiness of survival models, hindering their adoption in clinical decision-making and other sensitive applications. To bridge this gap, in this work, we introduce \emph{SurvUnc}, a novel meta-model based framework for post-hoc uncertainty quantification for survival models. SurvUnc introduces an anchor-based learning strategy that integrates concordance knowledge into meta-model optimization, leveraging pairwise ranking performance to estimate uncertainty effectively. Notably, our framework is model-agnostic, ensuring compatibility with any survival model without requiring modifications to its architecture or access to its internal parameters. Especially, we design a comprehensive evaluation pipeline tailored to this critical yet overlooked problem. Through extensive experiments on four publicly available benchmarking datasets and five representative survival models, we demonstrate the superiority of SurvUnc across multiple evaluation scenarios, including selective prediction, misprediction detection, and out-of-domain detection. Our results highlight the effectiveness of SurvUnc in enhancing model interpretability and reliability, paving the way for more trustworthy survival predictions in real-world applications.
\end{abstract}



\maketitle

\section{Introduction}\label{sec:intro}
Survival analysis, or equivalently time-to-event analysis, which aims to estimate when an event of interest is likely to occur, has received substantial attention from various fields, such as predicting patient death or disease risk in healthcare, product lifespan in manufacturing, and customer churn in finance \cite{wang2019machine}. Since the event of interest is not always observed (e.g., many patients are lost to follow-up), survival data are frequently right-censored (i.e., the event such as death occurs after the censoring) \cite{dey2022survival}, making survival analysis a more complex problem compared to traditional regression and classification tasks.

Driven by the advancements of machine learning and deep learning in recent years, several models have been proposed for survival analysis \cite{wiegrebe2024deep}. For example, RSF \cite{rsf} adapts random forest for this purpose. Built upon the linear Cox proportional hazard (CoxPH) model \cite{cox1972regression}, DeepSurv \cite{deepsurv} and SurvTRACE \cite{wang2022survtrace} apply fully connected networks and transformers to model the nonlinear relationship between covariates/features and the hazard rate of an event. DeepHit \cite{deephit} and DSM \cite{dsm} take a different approach by directly modeling the event occurrence probability using neural networks.

Despite their advancements, these survival models are subject to the inherent noise in the data and the lack of knowledge on the optimal modeling approach, making them inapplicable for unseen data in practice \cite{fakour2024structured}. The lack of uncertainty quantification for survival models significantly reduces their reliability in risk-sensitive applications such as healthcare \cite{loya2020uncertainty,Lu2024,xia2024uncertainty}. While much progress has been made in quantitatively measuring the reliability of a model in prediction \cite{ulmer2023prior}, most uncertainty quantification studies focus on classification and regression tasks. Bayesian approaches, such as Bayesian Neural Networks (BNNs) and Monte Carlo Dropout (MC-Dropout) \cite{mcdropout}, have been explored for uncertainty estimation in these domains \cite{sensoy2018evidential, Lu2024}. Additionally, meta-models have been developed to quantify classification uncertainty in a supervised setting \cite{Shen2023}. However, uncertainty quantification for survival models remains largely unexplored, presenting critical challenges that need to be addressed:
\begin{itemize}[leftmargin=10px]
    \item \textbf{Survival Model-Agnostic Integration.} The diversity of survival models requires a robust standalone uncertainty quantification framework, where the latter is capable of integrating into any model without modifications. For example, uncertainty quantification methods with BNNs and MC-Dropout are incompatible with survival models such as RSF \cite{rsf}. Thus, model-agnostic integration is essential to accommodate the wide range of existing and emerging survival models.
    
    \item \textbf{Absence of Ground-Truth Survival Curves.} Survival curves describe the survival probability of an event (e.g., death) not occurring by a specific time. However, in real-world settings, the ``true'' survival curves are unknown, and for censored samples, the actual event times remain uncertain. This absence of ground truth makes it infeasible to directly apply supervised uncertainty quantification methods to survival models, posing a significant challenge in reliability assessment.

    \item \textbf{Lack of Standardized Evaluation Protocols.} Despite the increasing focus on uncertainty quantification, there is no universally accepted evaluation framework for assessing the quality of uncertainty estimates in survival models. Beyond the methodological challenges in developing uncertainty quantification methods, establishing robust and systematic evaluation protocols remains an open problem. 
\end{itemize}

To address the aforementioned challenges, we propose \emph{SurvUnc}, a post-hoc meta-model based uncertainty quantification framework for survival models. More specifically, we develop a lightweight meta-model that acts as an ``observer'' on top of any existing survival model (base model). Sharing the same input covariates as the survival model, the meta-model is trained to estimate the uncertainty of the prediction of the survival model, which requires neither modifications to the survival model nor access to its architecture and parameters. Especially, motivated by the classic concordance idea in survival analysis \cite{harrell1982evaluating}, we design an anchor-based learning strategy to address the absence of ground-truth survival curves in meta-model optimization. Our strategy (i) selects a group of uncensored samples as anchors; (ii) evaluates whether the survival model correctly ranks the survival probabilities of samples relative to these anchors; and (iii) incorporates the evaluation outcome into the meta-model training to quantify the uncertainty of the survival model’s predictions. Furthermore, we refine and adapt existing evaluation protocols from uncertainty quantification studies in classification and regression to suit the survival analysis context, ensuring a more standardized assessment of uncertainty estimation in survival models.

In summary, our key contributions are as follows:
\begin{itemize}[leftmargin=10px]
    \item We propose SurvUnc, a post-hoc meta-model based uncertainty quantification framework for survival models. Notably, it requires no modifications or access to the model’s parameters. To the best of our knowledge, SurvUnc is the first model-agnostic framework capable of quantifying the uncertainty of predictions in any survival model.
    \item We develop an anchor-based learning strategy that leverages the concordance concept of survival analysis to construct the meta-model training set for optimization, thereby providing deeper insights into understanding the uncertainty in survival analysis.
    \item We design systematic evaluation protocols to assess the quality of quantified uncertainty in survival models, including selective prediction, misprediction detection, and out-of-domain (OOD) detection tasks. These protocols offer an evaluation reference for future research in uncertainty estimation for survival analysis.   
    \item We validate SurvUnc on multiple survival models and datasets. Our extensive experiments demonstrate that SurvUnc significantly outperforms baseline methods across various tasks, underscoring its effectiveness and robustness in uncertainty quantification for survival models.
\end{itemize}

\section{Related Work}\label{sec:related_work}
\subsection{Survival Analysis}
Survival analysis is concerned with modeling event occurrence in the presence of censoring, and it primarily involves two interrelated quantities: (i) \emph{hazard function}, which represents the instantaneous rate of event occurrence at a given time point, conditional on the event not having occurred prior to that time; and (ii) \emph{survival function}, (a.k.a. the \emph{survival curve}), which denotes the probability that the event has not occurred by a specific time. Survival probability is determined by the cumulative hazard up to that time. Survival models typically focus on estimating either the hazard function or the survival function \cite{wiegrebe2024deep}.

Traditionally, statistical methods have been widely explored for survival analysis. Kaplan-Meier estimator \cite{kaplan1958nonparametric} firstly defines the survival function based on empirical estimation of the survival data. CoxPH model \cite{cox1972regression} assumes that the log of the hazard is a linear combination of covariates, and the ratio between the hazards of two samples is constant. Furthermore, several machine learning methods have been adapted for survival analysis \cite{wang2019machine}. One of the most representative examples is RSF \cite{rsf}, which extends the random forest to survival analysis.

Owing to the ability to capture high-level non-linear interactions, neural network-based deep learning models have gained popularity in survival analysis \cite{wiegrebe2024deep}. Early models, such as the Farggi-Simon model \cite{faraggi1995neural} and DeepSurv \cite{deepsurv}, replace the linear combination in the Cox model with fully connected networks. Furthermore, recurrent neural networks and Transformers are also explored for modeling hazard functions in DRSA \cite{ren2019deep} and SurvTRACE \cite{wang2022survtrace}, respectively. On the other hand, DeepHit \cite{deephit} and DSM \cite{dsm} directly target the survival function using fully connected networks and parametric distributions. A recent work, MOTOR \cite{steinbergmotor} introduces a Transformer-based foundation model pretrained on time-to-event data for survival analysis. However, the uncertainty associated with predictions from such survival models remains unquantified. 

In addition, several Bayesian methods have been developed for survival analysis. For instance, the Gaussian process is firstly incorporated into the hazard function for a semi-parametric Bayesian model in \cite{fernandez2016gaussian}, while DSA \cite{ranganath2016deep,miscouridou2018deep} employs a deep latent variable model within a Bayesian framework to analyze survival data. BNNs \cite{loya2020uncertainty,lillelund2023uncertainty,lillelund2024efficient} and deep Gaussian processes \cite{alaa2017deep} further integrate neural network architectures with Bayesian techniques for survival prediction. Although Bayesian methods inherently provide uncertainty quantification through iterative sampling or posterior inference, they are typically confined to probabilistic models and often struggle with generalization or integration with other survival models. This limitation can reduce their practical reliability in real-world applications. Overall, accurately quantifying prediction uncertainty in survival models remains a significant challenge.

\subsection{Uncertainty Quantification}
Uncertainty quantification aims to quantitatively measure the reliability and confidence of a model's predictions, and can be categorized into \emph{intrinsic} and \emph{post-hoc (extrinsic)}, depending on whether uncertainty estimation is integrated within model training \cite{Shen2023}.

Intrinsic uncertainty quantification methods integrate uncertainty estimation directly into the model learning process, often leveraging Bayesian approaches to naturally generate uncertainty estimates as part of their predictions \cite{gawlikowski2023survey}. For instance, various studies employ BNNs with techniques such as variational inference and Laplace approximation to quantify uncertainty in both classification and regression tasks \cite{louizos2017multiplicative,posch2019variational}. In addition, evidential deep learning methods, which parameterize a Dirichlet distribution on the neural network outputs, have been used to quantify uncertainty in classification problems \cite{malinin2018predictive}. However, these methods are often computationally intensive and based on strong underlying assumptions. Furthermore, intrinsic methods cannot directly quantify the uncertainty in survival models.

In comparison, post-hoc uncertainty quantification methods separate the uncertainty estimation from the model prediction, without affecting the model learning process. Prominent examples include MC-Dropout\footnote{MC-Dropout requires a dropout layer to be designed in the quantified model.} \cite{mcdropout} and Deep Ensembles \cite{ensemble}, widely used in various domains \cite{he2023survey}. MC-Dropout estimates uncertainty by calculating the variance of multiple forward predictions with dropout layers activated during inference. Deep Ensembles define uncertainty as the prediction variance across multiple independent models trained from different random initializations, while Hyper-batch Ensembles \cite{wenzel2020hyperparameter} further consider different hyperparameters in the ensemble. Another promising paradigm is to build auxiliary or meta-models to quantify the uncertainty of the base model with respect to its original task, as studied in classification \cite{chen2019confidence,elder2021learning,Shen2023} and regression \cite{yu2024discretization}. However, the effectiveness of these methods in survival analysis remains unclear due to the unique challenges posed by censored data and the absence of ground-truth survival curves, which are fundamental to meta-model methods. Our work addresses this gap by introducing a novel post-hoc uncertainty quantification method specifically tailored for survival models.

\section{Proposed Framework}\label{sec:framework}

\subsection{Preliminaries \& Problem Definition}
\subsubsection{Survival Analysis.} Firstly, we define a survival dataset $\mathcal{D}=\{(\bm{x}_i, t_i, \delta_i)\}^N_{i=1}$, where $\bm{x}_i\in\mathbb{R}^d$ is the set of covariates/features of an individual; ${t}_i\in\mathbb{R}^{+}$ is the time to event or censoring as indicated by the indicator $\delta_i=1$ or $\delta_i=0$, respectively. $N$ refers to the number of samples and $d$ is the dimension of covariates. 

Moreover, the probability that an individual with covariates $\bm{x}$ will experience the event at time $t$, i.e., the probability density function of survival times, is denoted as $p(t|\bm{x})=\mathbb{P}(T\!=\!t|\bm{X}\!=\!\bm{x})$. Consequently, the survival function is represented as:
\begin{align}
    S(t|\bm{x})=\mathbb{P}(T>t|\bm{X}=\bm{x})=1-\int^t_0 p(z|\bm{x})\mathrm{d}z,
\end{align}
which signifies the probability that the event does not occur until time $t$. The hazard function is defined as:
\begin{align}
h(t|\bm{x})=\lim_{\Delta t\to0}\frac{\mathbb{P}(t\leq T<t+\Delta t|T\geq t, \bm{X}=\bm{x})}{\Delta t}\!=\!\frac{p(t|\bm{x})}{S(t|\bm{x})}, 
\end{align}
which indicates the probability that the event will occur at time $t$, given that the event has not occurred before. Consequently, the survival model $F(\cdot)$ is developed to learn $S(t|\bm{x})$ and $h(t|\bm{x})$ with covariates $\bm{x}$ as input, based on the survival dataset $\mathcal{D}$.

\subsubsection{Uncertainty Quantification.} In the context of machine learning and deep learning, the total uncertainty encompasses two types of uncertainty: (i) reducible \emph{epistemic uncertainty} (a.k.a. model uncertainty) caused by the model's limited knowledge due to insufficient training data, e.g., unseen data samples; and (ii) irreducible \emph{aleatoric uncertainty} (a.k.a. data uncertainty) caused by the inherent noise and stochastic nature in data \cite{hullermeier2021aleatoric}. Based on the preliminaries, we formally define the problem of uncertainty quantification in survival analysis as follows.
\newtheorem{problem}{Problem}
\begin{problem}{}
    \emph{\textbf{Uncertainty Quantification in Survival Analysis.}} Given a survival model $F(\cdot)$ trained on a survival dataset $\mathcal{D}$, the uncertainty quantification problem is to learn a function $U(\cdot):\mathbb{R}^d \to \mathbb{R}_{\geq0}$, such that based on $F(\cdot)$ and $\mathcal{D}$, it produces an uncertainty score that quantifies the predictive uncertainty of $F(\cdot)$ for a new coming sample.
\end{problem} 

\subsection{Framework Overview}

\begin{figure}[htbp]
    \centering
    \includegraphics[width=0.999\columnwidth]{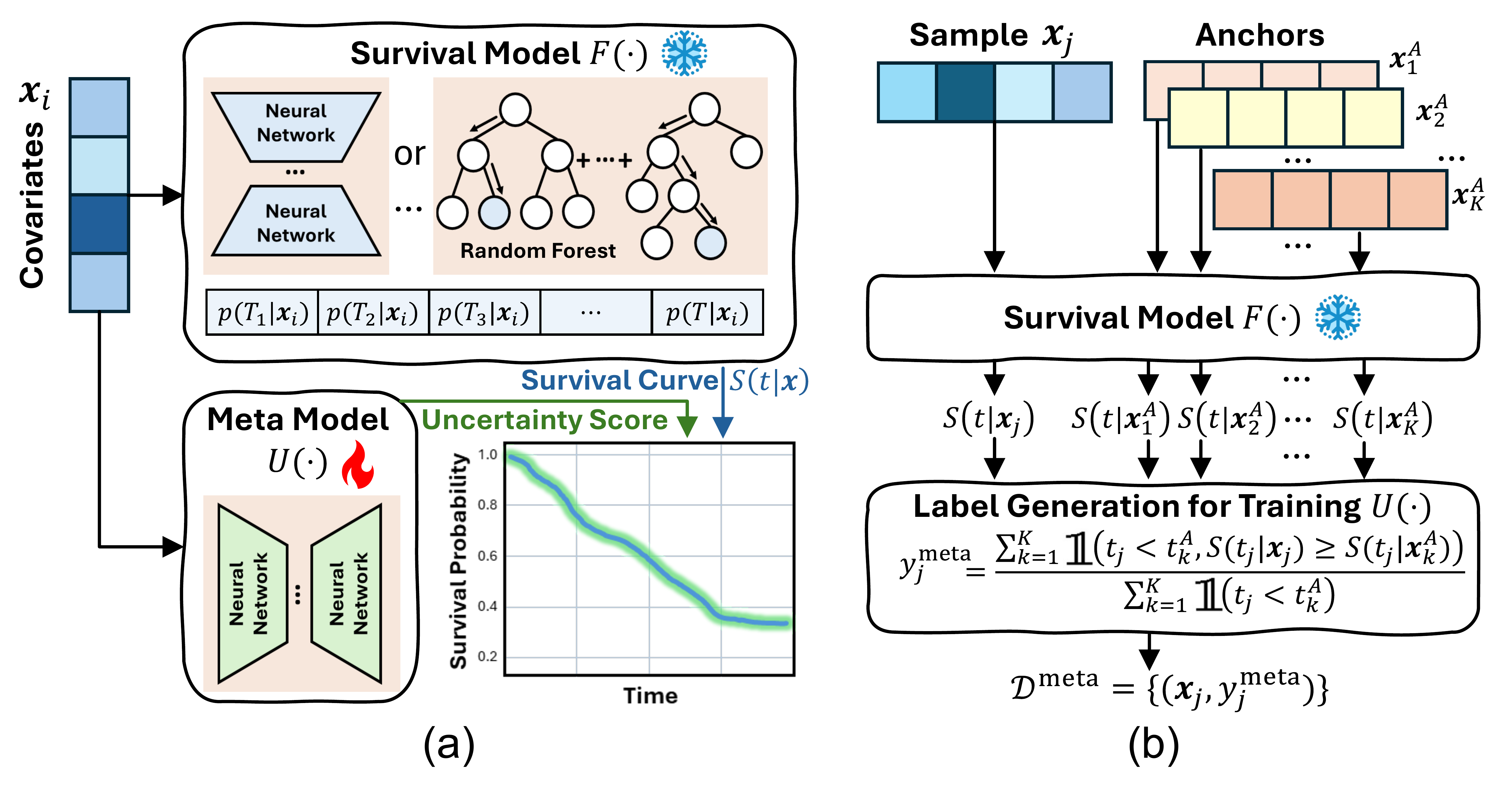}
    \vspace{-10px}
    \caption{Illustration of our proposed framework SurvUnc. (a) The pipeline of post-hoc meta-model based uncertainty quantification for survival models, and (b) the anchor-based learning strategy for meta-model optimization.}
    \label{fig:framework}
\end{figure}

Figure~\ref{fig:framework} illustrates an overview of our proposed framework, SurvUnc, for the uncertainty quantification problem in survival analysis. Specifically, we propose a meta-model based framework to quantify the uncertainty of survival models in a post-hoc manner. Given a pretrained survival model $F(\cdot)$ (e.g., neural network-based model, or random forest-based model), we further develop a meta-model $U(\cdot)$ to quantify its uncertainty. Notably, the meta-model shares the same covariate input as the survival model, and only the parameters of the meta-model are optimized in the learning process. In general, the architecture of the meta-model can be arbitrarily chosen; in our experiments, we explore two different variants of $U(\cdot)$, as shown in later experiments. To overcome the absence of ground truth of survival curves for meta-model training, we design an anchor-based learning strategy to construct a meta-model training set, $\mathcal{D}^{\text{meta}}$, derived from the survival model training set $\mathcal{D}$ and the pretrained survival model $F(\cdot)$, as illustrated in Figure~\ref{fig:framework}(b).

\subsection{Meta-Model Learning}

\begin{table*}[htbp]
\centering
\def\arraystretch{1.1}
\setlength\tabcolsep{6pt}
\caption{Dataset statistics. ``real'' and ``categorical'' denote the number of real-valued and categorical covariates in the datasets, respectively. Minimum, maximum, and mean values for event durations and censoring times are reported.}\label{tab:dataset}
\vspace{-10px}
\begin{tabular}{c|c|c|c|c|ccc|ccc}
\toprule
\multirow{2}{*}{\textbf{Dataset}} &  \multirow{2}{*}{\textbf{\#Instances}} & \multirow{2}{*}{\textbf{\#Events (\%)}} & \multirow{2}{*}{\textbf{\#Censored (\%)}} & \multirow{2}{*}{\begin{tabular}[c]{@{}c@{}}\textbf{\#Covariates}\\(real, categorical)\end{tabular}} & \multicolumn{3}{c|}{\textbf{Event Duration}} & \multicolumn{3}{c}{\textbf{Censoring Time}} \\
\cline{6-11}
 &  &  & & & \textbf{min} & \textbf{max} & \textbf{mean} & \textbf{min} & \textbf{max} & \textbf{mean} \\
\hline
FLCHAIN & 6,524 & 1,962 (30.1\%) & 4,562 (69.9\%) & 7 (3, 4) &  0 & 4998 & 2137.9 & 1 & 5166 & 4296.7  \\
\hline
SUPPORT & 8,873 & 6,036 (68.0\%) & 2,837 (32.0\%) & 14 (8, 6) & 3 & 1944 & 205.4 & 344 & 2029 & 1059.8 \\
\hline
SEER-BC & 323,772 & 85,258 (26.3\%) & 238,514 (73.7\%) & 18 (4, 14) & 1 & 121 & 41.7 & 1 & 121 & 85.4  \\
\hline
SAC3 & 100,000 & 62,798 (62.8\%) & 37,202 (37.2\%) & 49 (49, 0) & 0.1 & 100 & 24.6 & 0.1 & 100 & 45.5  \\
\bottomrule
\end{tabular}
\end{table*}

To construct a labeled meta-model training set, we leverage the concordance concept from survival analysis in the medical domain \cite{harrell1982evaluating} and pairwise ranking idea in information retrieval domain \cite{liu2009learning}. The core idea in medical language is that a patient living shorter should have a lower survival probability at any given time compared to a patient living longer. Therefore, the uncertainty in a survival model’s predictions can be quantified by evaluating its ability to correctly rank the survival probabilities of samples relative to each other.

Intuitively, comparing a sample against some referenced points enhances the quality of the quantified uncertainty. For efficiency consideration, we design an anchor-based learning strategy. Following the literature we assume that censoring occurs completely at random \cite{deephit, dsm}. Specifically, we randomly select $K$ uncensored samples from the survival model training set $\mathcal{D}$ as anchors, denoted by $\mathcal{D}^{A}\!\!=\!\!\{(\bm{x}^A_k,t^A_k, \delta^A_k\!\!=\!\!1)\}^K_{k=1}$. For the $j$-th uncensored sample $(\bm{x}_j, t_j, \delta_j=1)\in\mathcal{D}$, we treat it as a training instance for the meta-model, and calculate its uncertainty label as follows:
\begin{align}
y^\text{meta}_{j}=\frac{\sum^K_{k=1}\boldsymbol{\mathbbm{1}}\!\left(t_j<t^A_k,S\left(t_j|\bm{x}_j\right)\geq S\left(t_j|\bm{x}^A_k\right)\right)}{\sum^K_{k=1}\boldsymbol{\mathbbm{1}}\!\left(t_j<t^A_k\right)}, \label{eq:label}
\end{align}
which measures the proportion of incorrectly ordered pairs by the survival model $F(\cdot)$, resulting in an uncertainty score ranging from 0 to 1 for the predicted sample. A higher uncertainty score indicates a less reliable prediction. Here $\boldsymbol{\mathbbm{1}}\!(\cdot)$ is an indicator function that returns 1 if the specified conditions are met and 0 otherwise. $S(t_j|\bm{x}_j)$ and $S(t_j|\bm{x}^A_k)$ denote the survival probabilities estimated by $F(\cdot)$ for samples with covariates $\bm{x}_j$ and $\bm{x}^A_k$ at time $t_j$, respectively. As a result, we construct the meta-model training set $\mathcal{D}^\text{meta}=\{(\bm{x}_j, y^\text{meta}_j)\}_{j=1}^{N^{\prime}}$ with $N^\prime$ labeled samples. 

Based on the constructed meta-model training set $\mathcal{D}^{\text{meta}}$, we explore two different architectures of the meta-model $U(\cdot)$ with sample covariates as input, including a multi-layer perceptron (MLP) model and a random forest model, referred to as \textbf{SurvUnc-MLP} and \textbf{SurvUnc-RF}, respectively. After training on $\mathcal{D}^\text{meta}$, the trained SurvUnc models provide uncertainty scores for the predictions from the corresponding survival model. Algorithm~\ref{alg:framework} summarizes the learning procedure of SurvUnc. Note that during inference, the event or censoring time for a test sample is unknown, requiring the meta-model $U(\cdot)$ to estimate predictive uncertainty solely from the covariates.

\SetKwInput{KwInit}{Init}
\SetKwInput{KwUse}{Usage}
\begin{algorithm}[tbp] 
    \caption{The learning procedure of SurvUnc}\label{alg:framework}
    \KwIn{\RaggedRight{Survival data $\mathcal{D}=\{(\bm{x}_i, t_i, \delta_i)\}^N_{i=1}$;\\ 
    \hspace{3em}Pretrained survival model $F(\cdot)$; Anchor number $K$}}
    \KwOut{Uncertainty quantification model $U(\cdot)$}

    Randomly sample $K$ anchors from $\mathcal{D}$, i.e., $\mathcal{D}^{A}$

    Initialize the model $U(\cdot)$
    
    $\mathcal{D}^{\text{meta}}\leftarrow\emptyset$

    \For{$(\bm{x}_j, t_j,\delta_j=1)\in\mathcal{D}$}{
        Obtain label $y_i^\text{meta}$ based on \eqref{eq:label} and $\mathcal{D}^{A}$\\
        $\mathcal{D}^\text{meta}\leftarrow \mathcal{D}^\text{meta} \cup \{(\bm{x}_i, y^\text{meta}_i)\}$
    }
    Train the model $U(\cdot)$ with $\mathcal{D}^{\text{meta}}$ \\
    \KwUse{$u_i = U(\bm{x}_{i})$}
\end{algorithm}

\section{Experiments}\label{sec:exp}
In this section, we show that our proposed SurvUnc framework effectively quantifies uncertainty in survival models and is applicable in different scenarios. Specifically, we conduct experiments to answer the following questions:
\begin{itemize}[leftmargin=10px]
\item \textbf{RQ1:} How does SurvUnc perform w.r.t. total uncertainty quantification, particularly in distinguishing between correctly and incorrectly predicted samples, compared to existing methods?
\item \textbf{RQ2:} How does SurvUnc perform w.r.t. epistemic uncertainty quantification, particularly in detecting OOD data, when compared to existing methods?
\item \textbf{RQ3:} How do different hyperparameter settings, such as meta-model structure and the number of anchors, affect the performance of SurvUnc?
\end{itemize}

\subsection{Experimental Settings}\label{sec:exp_settings}
\subsubsection{Datasets.}
We experiment with three real-world and one synthetic, publicly available survival analysis datasets:
\begin{itemize}[leftmargin=10px]
    \item \textbf{FLCHAIN} is a study of the relationship between free light chains and mortality in an elderly population \cite{flchain}.
    \item \textbf{SUPPORT} contains a dataset that aims to improve the care for seriously ill patients by understanding their prognosis and treatment preferences \cite{support}.
    \item \textbf{SEER-BC} is derived from the SEER database \cite{seer}, which contains survival information for oncology patients in the U.S. Following the processing steps in \cite{wang2022survtrace}, we select breast cancer patients to create the SEER-BC dataset. For the later OOD detection experiments, we curate a SEER-HD dataset, comprising the same number of patients as SEER-BC but diagnosed with heart disease.
    \item \textbf{SAC3} is a synthetic dataset from \cite{kvamme2021continuous}, which simulates survival times by sampling from a combination of three hazard functions. Thus, this dataset provides the ground-truth survival curves.
\end{itemize} 
Table~\ref{tab:dataset} presents the basic statistics of all datasets. For the SUPPORT dataset, we retain the original training/test split and reserve 20\% of the training set for validation. For all other datasets, we randomly split into training, validation, and test sets in a ratio of 6:2:2. 

\subsubsection{Survival Models.} To evaluate the effectiveness as well as robustness of our proposed SurvUnc framework on survival models, we select five representative survival models for uncertainty quantification, including three deep learning models of \textbf{DeepSurv} \cite{deepsurv}, \textbf{DeepHit} \cite{deephit}, \textbf{DSM} \cite{dsm}, one machine learning model of \textbf{RSF} \cite{rsf}, and one Bayesian-based model of \textbf{BNNSurv} \cite{lillelund2023uncertainty}. For each survival model, we optimize hyperparameters so that the reported performance on datasets is comparable to those published in the literature, thereby validating the correctness of pretrained survival models. 

\subsubsection{Uncertainty Quantification Baselines.} Given the lack of prior studies on model-agnostic uncertainty quantification for survival models, we adapt two widely used uncertainty quantification methods of \textbf{MC-Dropout} \cite{mcdropout} and \textbf{Deep Ensembles (Ensemble)} \cite{ensemble} to survival analysis. For MC-Dropout, we set the number of forward passes to 100. For Ensemble, we train 10 survival models with different random initializations. Note that the RSF model does not support the dropout mechanism, making MC-Dropout incompatible with it. Unlike MC-Dropout and Ensemble, which are general uncertainty quantification methods applicable across models, {BNNSurv} is a specialized Bayesian survival model that inherently estimates uncertainty but cannot be integrated with other survival models for uncertainty quantification.

\subsubsection{Tasks \& Metrics.}
As discussed before, standardized evaluation protocols for uncertainty quantification methods with survival models remain lacking. To address this, we extend existing evaluation methodologies from uncertainty quantification studies to the survival analysis domain, incorporating multiple tasks and metrics.

\textbf{Selective prediction} evaluates how well an uncertainty quantification method differentiates between correct and incorrect predictions. A good uncertainty quantification method should improve survival model performance by selectively discarding samples with high uncertainty. To evaluate this, we use two widely adopted survival analysis metrics.
\begin{itemize}[leftmargin=10px]
    \item \textbf{Time Dependent Concordance Index ($C^{\text{td}}$).} This metric measures the model discrimination power by comparing the relative survival probabilities across all pairs of samples in the test set \cite{antolini2005time}, defined as 
    \begin{align}
    C^{\text{td}}=\frac{\sum_{i=1}^n\sum_{j=1}^n\delta_i\cdot \boldsymbol{\mathbbm{1}}\!\left(t_i<t_j, S(t_i|\bm{x}_i)<S(t_i|\bm{x}_j)\right)}{\sum_{i=1}^n\sum_{j=1}^n\delta_i\cdot \boldsymbol{\mathbbm{1}}\!\left(t_i<t_j\right)}. \nonumber
    \end{align}
The range of $C^{\text{td}}$ is [0, 1], and a larger value indicates better model prediction. $C^{\text{td}}$=0.5 corresponds to a random prediction.

\item \textbf{Integral Brier Score (IBS).} This metric is an integral of the Brier score at all time points, while the Brier score calculates the mean squared error between predicted survival probability and binary observation at a given time \cite{kvamme2023brier}, defined as 
    \begin{align}
    BS(t)\!=\!\frac{1}{n}\!\sum^n_{i=1}[\frac{(0-S(t|\bm{x}_i))^2\cdot\!\!\boldsymbol{\mathbbm{1}}\!({t\!\geq\! t_i, \delta_i=1})}{\hat{G}(t_i)}\!+\!\frac{(1-S(t|\bm{x}_i))^2\!\cdot\!\boldsymbol{\mathbbm{1}}\!(t\!<\!t_i)}{\hat{{G}}(t)}], \nonumber
    \end{align}
where $\hat{G}(t)$ is the censoring survival function using Kaplan-Meier estimator \cite{kaplan1958nonparametric}. Unlike $C^{\text{td}}$, which relies on pairwise comparisons, IBS assesses each sample independently and provides an overall measure of model accuracy, where lower values indicate better performance. The range of IBS is [0, 1].
\end{itemize}

\textbf{Misprediction detection} examines whether uncertainty estimates align with actual prediction errors. To assess this, we use the following metric:
\begin{itemize}[leftmargin=10px]
    \item \textbf{Pearson Correlation Coefficient $\rho$.} This metric measures the linear correlation between the quantified uncertainty scores and IBSs of test samples, i.e., to evaluate whether the uncertainty quantification method can distinguish correctly and incorrectly predicted samples by survival models. The range of $\rho$ is [-1, 1], while we expect a positive correlation ($\rho\geq0$) here, i.e., samples with larger uncertainty have larger IBS.
\end{itemize}

\textbf{OOD detection} evaluates the ability of uncertainty quantification methods to distinguish in-distribution (IND) from OOD samples. We use the following two metrics:
\begin{itemize}[leftmargin=10px]
    \item \textbf{AUROC and AUPRC.} We adopt the area under the receiver operating curve (AUROC) and the area under the precision-recall curve (AUPRC) to evaluate the performance in OOD detection experiments. Specifically, IND test samples are labeled as the negative class, while OOD samples are labeled as the positive class \cite{Shen2023}. Both metrics are in the range of [0, 1], and a higher value indicates better performance.
\end{itemize}

\begin{table*}[htbp]
    \centering
    \def\arraystretch{1.1}
    \setlength\tabcolsep{5.5pt}
    \caption{$C^{\text{td}}$ of survival models under different discarding percentages (10\%, 30\%, 50\%), determined by different uncertainty quantification (UQ) methods across datasets. The best results are in bold, and the last row of each group shows relative improvement compared with the best baseline by 100 runs of experiments. $p$-value$<0.001$ is true for all results without $^*$.}\label{tab:selective_cindex}
    \vspace{-10px}
    \begin{tabular}{c|c|ccc|ccc|ccc|ccc}
        \toprule
        \multirow{2}{*}{\makecell{\textbf{Survival}\\ \textbf{Models}}} & \multirow{2}{*}{\textbf{UQ Methods}} & \multicolumn{3}{c|}{\textbf{FLCHAIN}} & \multicolumn{3}{c|}{\textbf{SUPPORT}} & \multicolumn{3}{c|}{\textbf{SEER-BC}} & \multicolumn{3}{c}{\textbf{SAC3}} \\
        \cline{3-14}
        & & 10\% & 30\% & 50\% & 10\% & 30\% & 50\% & 10\% & 30\% & 50\% & 10\% & 30\% & 50\% \\
        \hline
        \multirow{5}{*}{DeepSurv} & MC-Dropout  & 0.782  & 0.745  & 0.685  & 0.600  & 0.576  & 0.563  & 0.857  & 0.846  & 0.832  & 0.772  & 0.778  & 0.793  \\
        & Ensemble & 0.786  & 0.755  & 0.710  & 0.607  & 0.598  & 0.576  & 0.863  & 0.857  & 0.842  & 0.783  & 0.795  & 0.807\\ 
        & SurvUnc-RF & \textbf{0.856}  & \textbf{0.907}  &\textbf{0.941}  & 0.635  & 0.690  & 0.757  & \textbf{0.904}  & \textbf{0.938}  & \textbf{0.961}  & 0.792  & 0.822  & 0.855\\
        & SurvUnc-MLP & 0.839  & 0.894  & 0.935  & \textbf{0.637}  & \textbf{0.695}  & \textbf{0.762}  & \textbf{0.904}  & \textbf{0.938}  & \textbf{0.961}  & \textbf{0.797}  & \textbf{0.830}  & \textbf{0.862}  \\ \cline{2-14}
        & Improv. & 8.9\%  & 20.1\%  & 32.5\%  & 4.9\%  & 16.2\%  & 32.3\%  & 4.8\%  & 9.5\%  & 14.1\%  & 1.8\%  & 4.4\%  & 6.8\%\\
        \hline
        \multirow{5}{*}{DeepHit} & MC-Dropout & 0.784  & 0.762  & 0.733  & \textbf{0.648}  & 0.657  & 0.656  & 0.908  & 0.937  & 0.953  & 0.797  & 0.811  & 0.836\\
        & Ensemble  & \textbf{0.855}  & 0.902  & 0.934  & 0.637  & 0.637  & 0.637  & 0.896  & 0.923  & 0.944  & 0.805  & 0.822  & 0.842 \\
        & SurvUnc-RF   & \textbf{0.855}$^*$  & 0.908  & \textbf{0.940}  & {0.639}  & \textbf{0.664}$^*$  & \textbf{0.693}  & 0.913  & \textbf{0.945}  & \textbf{0.961}  & 0.811  & 0.838  & 0.866\\
        & SurvUnc-MLP  & \textbf{0.855}$^*$  & \textbf{0.909}  & \textbf{0.940}  & {0.639}  & 0.660$^*$  & 0.692  & \textbf{0.914}  & \textbf{0.945}  & \textbf{0.961}  & \textbf{0.817}  & \textbf{0.845}  & \textbf{0.876} \\ \cline{2-14}
        & Improv. & 0.0\%  & 0.8\%  & 0.6\%  & -1.4\%  & 1.1\%  & 5.6\%  & 0.7\%  & 0.9\%  & 0.8\%  & 1.5\%  & 2.8\%  & 4.0\%\\
        \hline
        \multirow{5}{*}{DSM} & MC-Dropout & 0.780  & 0.741  & 0.687  & 0.613  & 0.614  & 0.613  & 0.866  & 0.854  & 0.832  & 0.793  & 0.776  & 0.770\\
        & Ensemble & 0.787  & 0.755  & 0.715  & 0.616  & 0.605  & 0.581  & 0.872  & 0.868  & 0.854  & 0.802  & 0.813  & 0.828 \\
        & SurvUnc-RF  & \textbf{0.854}  & \textbf{0.907}  & \textbf{0.941}  & 0.640  & \textbf{0.685}  & 0.725  & \textbf{0.910}  & \textbf{0.943}  & \textbf{0.960}  & 0.813  & 0.837  & 0.862\\
        & SurvUnc-MLP & 0.852  & 0.906  & 0.940  & \textbf{0.641}  & \textbf{0.685}  & \textbf{0.731}  & \textbf{0.910}  & \textbf{0.943}  & 0.959  & \textbf{0.818} & \textbf{0.844}  & \textbf{0.871} \\  \cline{2-14}
        & Improv. & 8.5\%  & 20.1\%  & 31.6\%  & 4.1\%  & 11.6\%  & 19.2\%  & 4.4\%  & 8.6\%  & 12.4\%  & 2.0\%  & 3.8\%  & 5.2\%\\
        \hline
        \multirow{4}{*}{RSF} & MC-Dropout & -& -& -&- & -& -& -&- &- &- &- & -\\
        & Ensemble   & 0.790  & 0.777  & 0.745  & 0.648  & 0.662  & 0.684  & 0.878  & 0.874  & 0.863  & 0.649  & 0.663  & 0.677    \\
        & SurvUnc-RF  & \textbf{0.840}  & \textbf{0.897}  & \textbf{0.933}  & \textbf{0.663}  & \textbf{0.708}  & \textbf{0.750}  & 0.908  & 0.941  & 0.959  & \textbf{0.658}  & \textbf{0.692}  & \textbf{0.738}  \\
        & SurvUnc-MLP  & 0.820  & 0.854  & 0.892  & 0.656  & 0.689  & 0.721  & \textbf{0.913}  & \textbf{0.943}  & \textbf{0.960}  & 0.653  & 0.685  & 0.731 \\ \cline{2-14}
        & Improv. & 6.3\%  & 15.4\%  & 25.2\%  & 2.3\%  & 6.9\%  & 9.6\%  & 4.0\%  & 7.9\%  & 11.2\%     & 1.4\%  & 4.4\%  & 9.0\%\\
         \hline 
        \multirow{3}{*}{BNNSurv} & Bayesian & 0.773 & 0.732 &  0.670 & 0.623 & 0.648 & 0.687 & 0.847 & 0.836 & 0.805 & 0.719 & 0.740 & 0.771\\
        & SurvUnc-RF  & \textbf{0.848}  & \textbf{0.902}  & \textbf{0.936}  & \textbf{0.650}  & \textbf{0.701}  & \textbf{0.764}  & 0.891  & \textbf{0.932}  & \textbf{0.955}  & \textbf{0.727}  & {0.765}  & {0.808}  \\
        & SurvUnc-MLP  & 0.843  & 0.899  & 0.934  & 0.648  & 0.699  & 0.760  & \textbf{0.893}  & \textbf{0.932}  & \textbf{0.955}  & \textbf{0.727}  & \textbf{0.767}  & \textbf{0.815} \\ \cline{2-14}
        & Improv. & 9.7\%  & 23.2\%  & 39.7\%  & 4.3\%  & 8.2\%  & 11.2\%  & 5.4\%  & 11.5\%  & 18.6\%  & 1.1\%  & 3.6\%  & 5.7\%\\
    
        \bottomrule
    \end{tabular}
\end{table*}

\subsubsection{Implementation Details}
For the SurvUnc-RF method, the hyperparameters n\_estimators, min\_samples\_leaf, min\_samples\_split are uniformly set to 100, 5 and 10, respectively, across all survival models and datasets. Similarly, for the SurvUnc-MLP method, the learning rate and the hidden layers are simply set to 0.001 and [32, 32], respectively. The number of anchors is usually set to 50 for robust performance. All methods undergo 100 bootstrap resampling iterations on the test set, with both the mean value and standard deviation reported. We also conduct the Wilcoxon signed-rank test to confirm the statistical significance of our results. All experiments were run on an RTX 6000 GPU with 32GB RAM. The implementation is done in PyTorch, and training the SurvUnc framework with a pretrained survival model on the largest SEER-BC dataset takes less than 5 minutes, making it efficient enough for deployment. The implementation code and dataset are available at the given link\footnote{\url{https://github.com/liuyuaa/SurvUnc}}. Details on the experimental settings can be found in Appendix~\ref{sec:app_exp}.

\begin{figure*}[!h]
    \centering
    \begin{subfigure}[b]{0.246\textwidth}
        \centering
        \includegraphics[width=\textwidth]{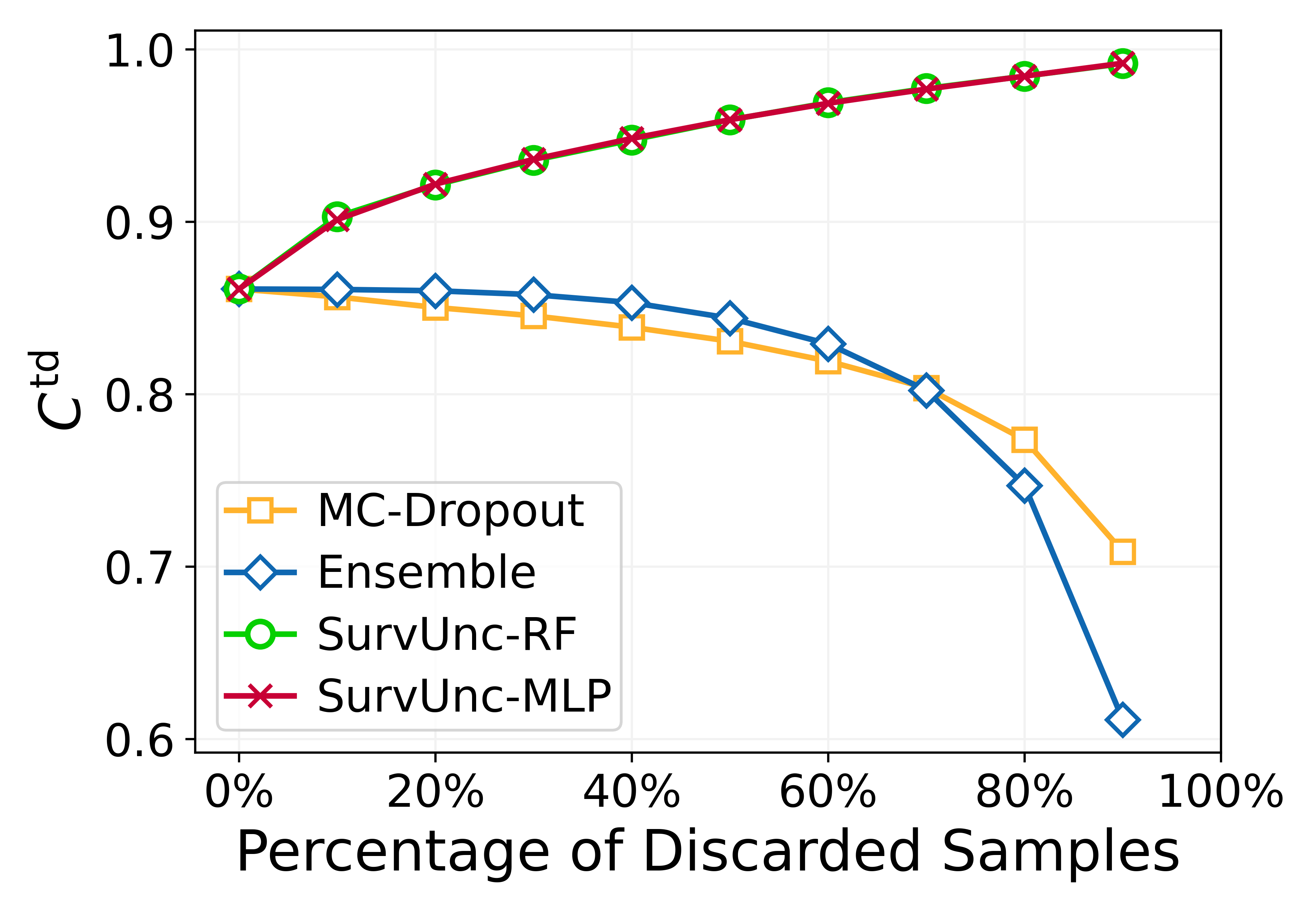}
        \caption{DeepSurv}
    \end{subfigure}
    \hfill
    \begin{subfigure}[b]{0.246\textwidth}
        \centering
        \includegraphics[width=\textwidth]{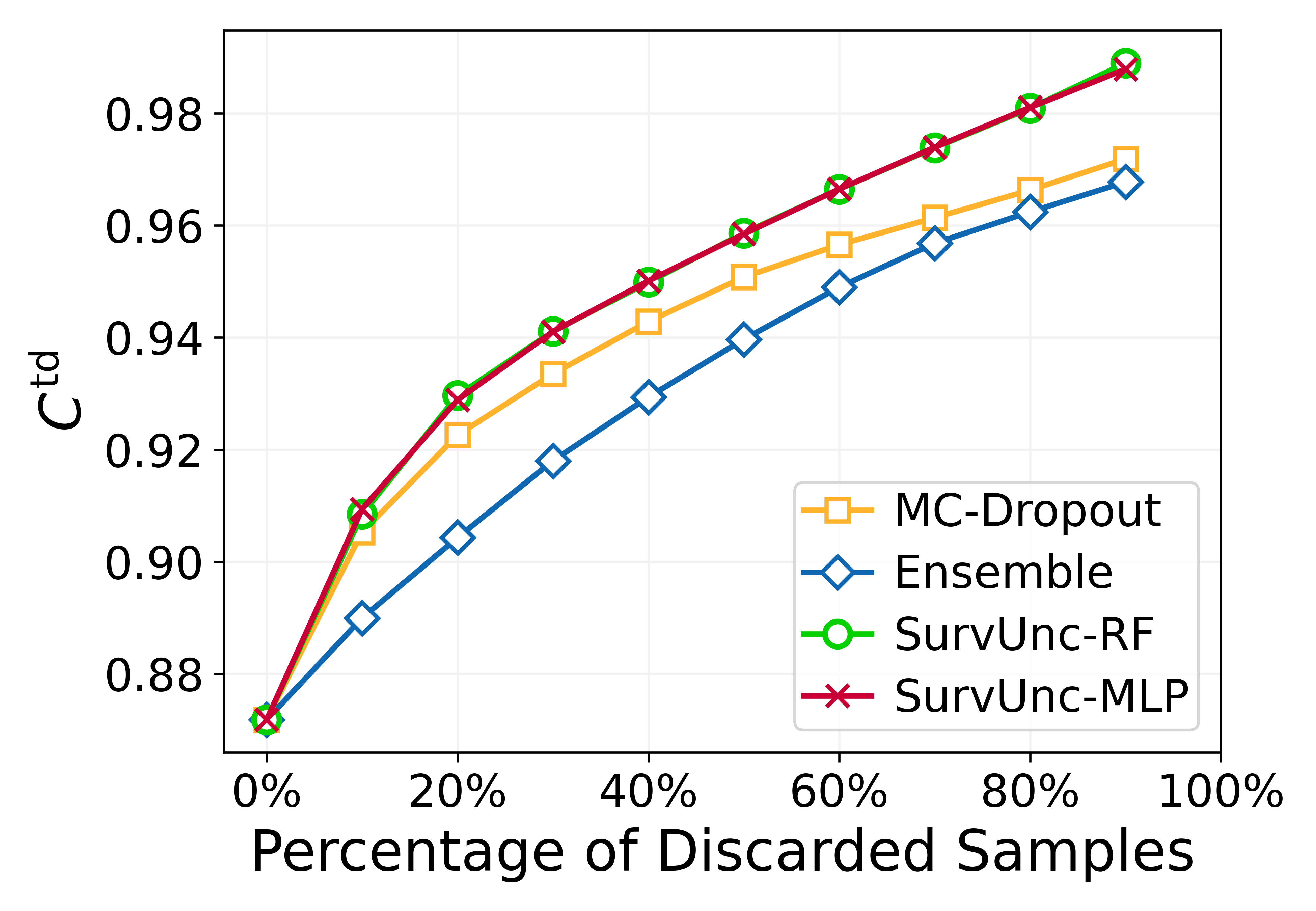}
        \caption{DeepHit}
    \end{subfigure}
    \hfill
    \begin{subfigure}[b]{0.246\textwidth}
        \centering
        \includegraphics[width=\textwidth]{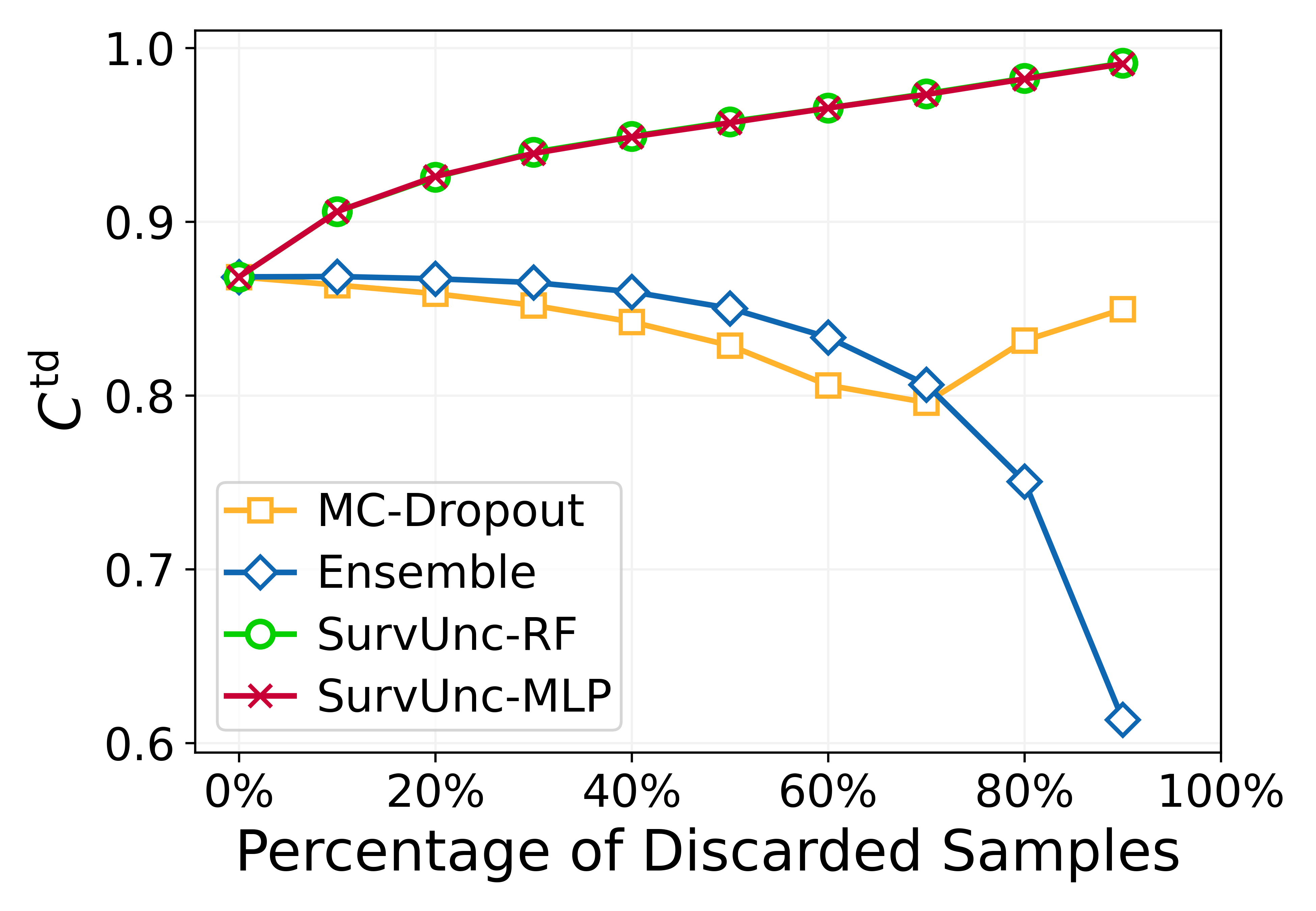}
        \caption{DSM}
    \end{subfigure}
    \hfill
    \begin{subfigure}[b]{0.246\textwidth}
        \centering
        \includegraphics[width=\textwidth]{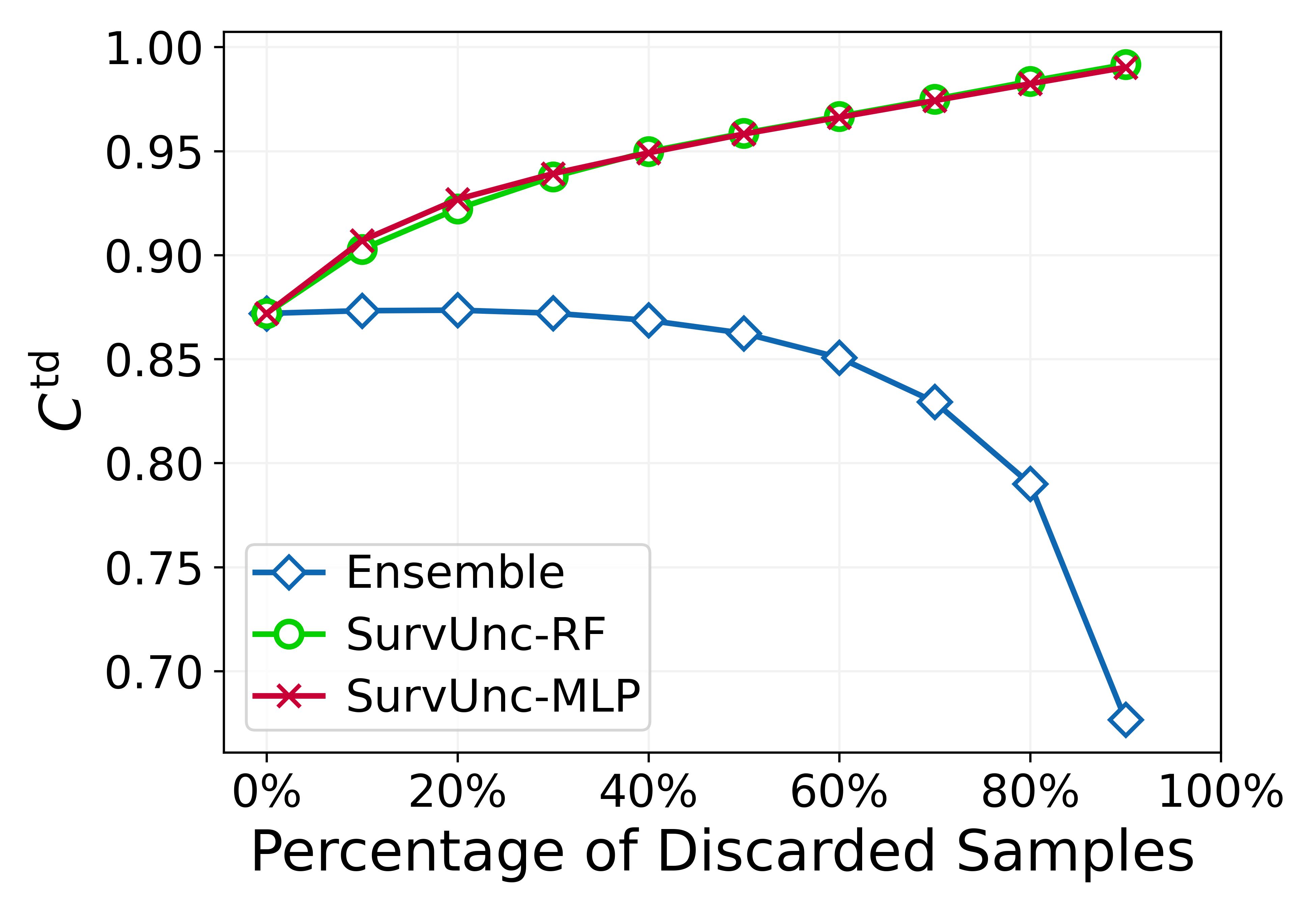}
        \caption{RSF}
    \end{subfigure}
    \vspace{-10px}
    \caption{$C^\text{td}$ of four survival models of (a) DeepSurv, (b) DeepHit, (c) DSM and (d) RSF on SEER-BC dataset with different percentages of samples discarded according to uncertainty scores from different UQ methods. A consistent upward trend is expected as the percentage of discarded samples increases. Error bars are omitted for better visualization.}
    \label{fig:selective_prediction}
\end{figure*}

\begin{table*}[!h]
\centering
\setlength\tabcolsep{1.pt}
\def\arraystretch{1.1}
\caption{Misprediction detection results, i.e., Pearson correlation coefficient between uncertainty scores and IBSs of samples.}\label{tab:misprediction}
\vspace{-10px}
\begin{tabular}{c|cccc|cccc|cccc|cccc}
\toprule
{\textbf{Datasets}} & \multicolumn{4}{c|}{\textbf{FLCHAIN}} & \multicolumn{4}{c|}{\textbf{SUPPORT}} & \multicolumn{4}{c|}{\textbf{SEER-BC}} & \multicolumn{4}{c}{\textbf{SAC3}}  \\
\hline
{\textbf{UQ Methods}} & DeepSurv & DeepHit & DSM & RSF  & DeepSurv  & DeepHit  & DSM & RSF   & DeepSurv & DeepHit & DSM & RSF & DeepSurv  & DeepHit  & DSM & RSF \\
\hline
MC-Dropout & -0.674 &-0.042 & -0.579& - & -0.299& 0.431& -0.123 & - & -0.293& \textbf{0.442} & -0.644 & - & -0.111&-0.157 &-0.329 & - \\ \hline
Ensemble & -0.361& \textbf{0.099} & -0.467&-0.258 &-0.109 & 0.511& -0.277& 0.203 & -0.168& 0.391 & -0.209 & -0.264 & -0.036& -0.017 & -0.085 & 0.129 \\
\hline
SurvUnc-RF & \textbf{0.688} & 0.084 & \textbf{0.718} & \textbf{0.687} & \textbf{0.657} & \textbf{0.590} & \textbf{0.516} & \textbf{0.544} & \textbf{0.683} & 0.331& \textbf{0.698} & 0.607 & 0.426 & 0.315& 0.342 & \textbf{0.436} \\ 
SurvUnc-MLP &0.590 & 0.053 & 0.700 & 0.352 & 0.652 & 0.509 &0.468 & 0.405& 0.671&0.323 & 0.677 & \textbf{0.666} & \textbf{0.638} & \textbf{0.510} & \textbf{0.554} & 0.391 \\
\bottomrule
\end{tabular}
\vspace{-10px}
\end{table*}

\subsection{Selective Prediction (RQ1)}
To investigate the effectiveness of total uncertainty quantification, we first conduct the selective prediction experiments \cite{van2020uncertainty,zhang2023uncertainty}. Specifically, we discard a portion of test samples by their uncertainty scores and then compute evaluation metrics on the remaining data. To better reflect real-world scenarios where testing samples have not yet been censored, we discard only uncensored samples, and subsequent analyses also focus on uncensored ones. It is expected that the performance will improve as more high-uncertainty samples are discarded. This evaluation is essential in realistic scenarios, where low-uncertainty predictions are retained, while high-uncertainty (less reliable) predictions are flagged for expert inspection.

Table~\ref{tab:selective_cindex} presents the selective prediction results for $C^\text{td}$ across five survival models, evaluated at different discarding percentages (10\%, 30\%, 50\%). Since BNNSurv inherently provides uncertainty estimation, we exclude MC-Dropout and Ensemble implementations for this model. Across all survival models, datasets and discarding percentages, our proposed SurvUnc framework generally outperforms the baselines with an average improvement of over 8\%, which demonstrates the effectiveness of the meta-model based uncertainty quantification framework and the anchor-based learning strategy. Especially, our approach aligns with the core objective of survival analysis, i.e., maintaining correct risk ordering, whose uncertainty scores better reflect this clinical need by focusing on ranking reliability. In comparison,  traditional methods focus on predictive variability, which do not account for the time-dependent nature of survival predictions. This leads to two main issues including that they may flag samples with high variance but correct rankings as ``uncertain'' and they often miss samples with small variance but incorrect relative rankings. Notably, the relatively weak performance of BNNSurv’s inherent uncertainty estimation suggests that Bayesian methods may struggle to capture meaningful uncertainty scores in survival analysis. The marginal improvement observed with DeepHit could be explained by its use of ranking loss, which also leverages the concordance concept, thereby making MC-Dropout and Ensemble methods based on its output more effective. Additionally, SurvUnc-RF and SurvUnc-MLP demonstrate comparable performance across different survival models and datasets, further validating the robustness and generalizability of SurvUnc.

Figure~\ref{fig:selective_prediction} shows the performance comparison with fine-grained discarding percentages on the largest dataset, SEER-BC. We observe that the performance of both SurvUnc-RF and SurvUnc-MLP improves as the discarding percentage increases, indicating that predictions become more reliable as high-uncertainty samples are excluded, consistent with the expectation of selective prediction experiments. However, for baselines quantifying the uncertainty of DeepSurv, DSM and RSF (see Figure~\ref{fig:selective_prediction}(a), (c) and (d)), the results are opposite, suggesting that these baselines fail to provide meaningful uncertainty quantification for these survival models. We also conduct the experiments using the IBS metric, which yields similar conclusions for most results (see Appendix~\ref{sec:app_results}).

\begin{figure}[htbp]
    \centering
    \begin{subfigure}[b]{0.232\textwidth}
        \centering
        \includegraphics[width=\textwidth]{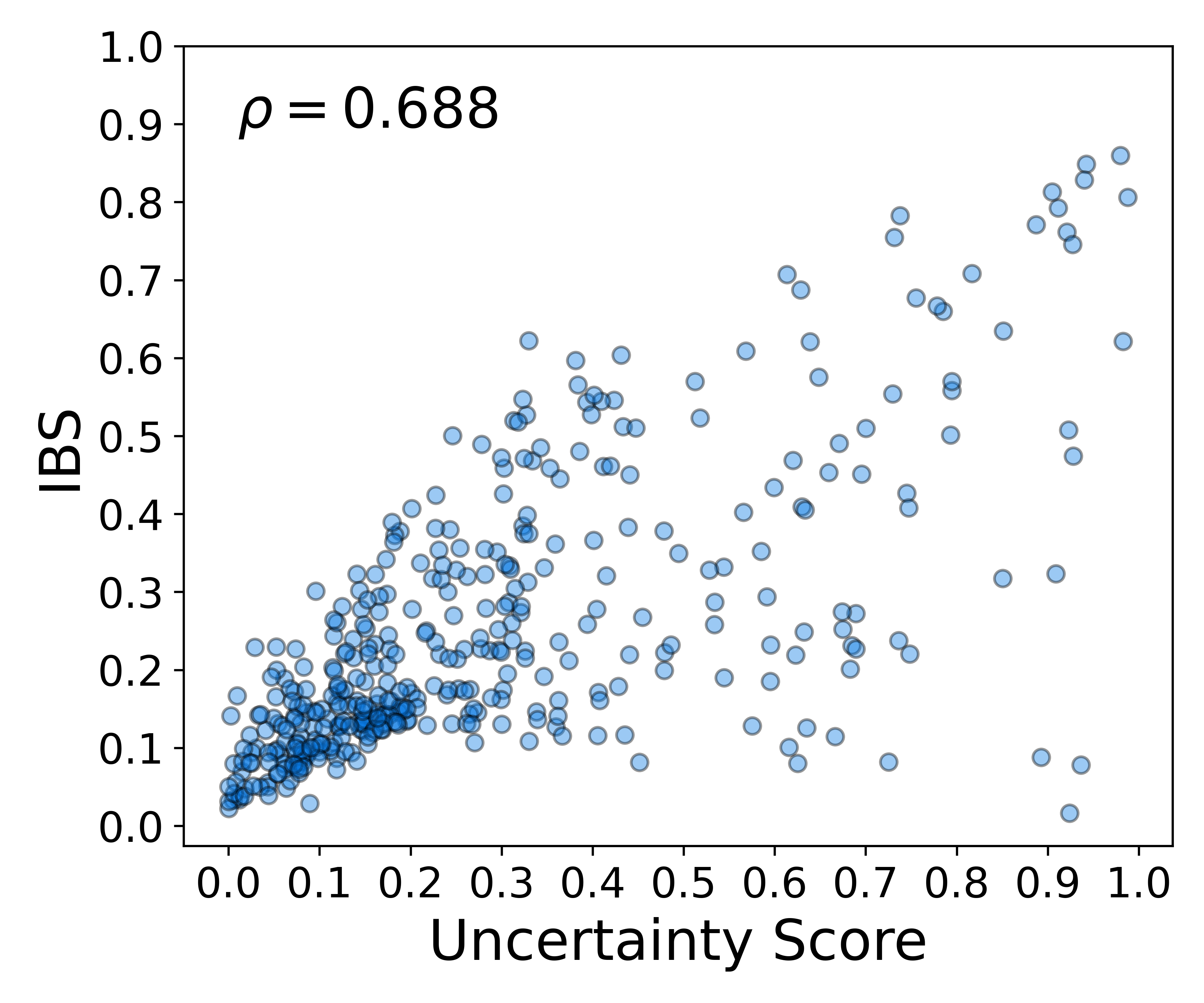}
        \caption{FLCHAIN}
    \end{subfigure}
    \begin{subfigure}[b]{0.232\textwidth}
        \centering
        \includegraphics[width=\textwidth]{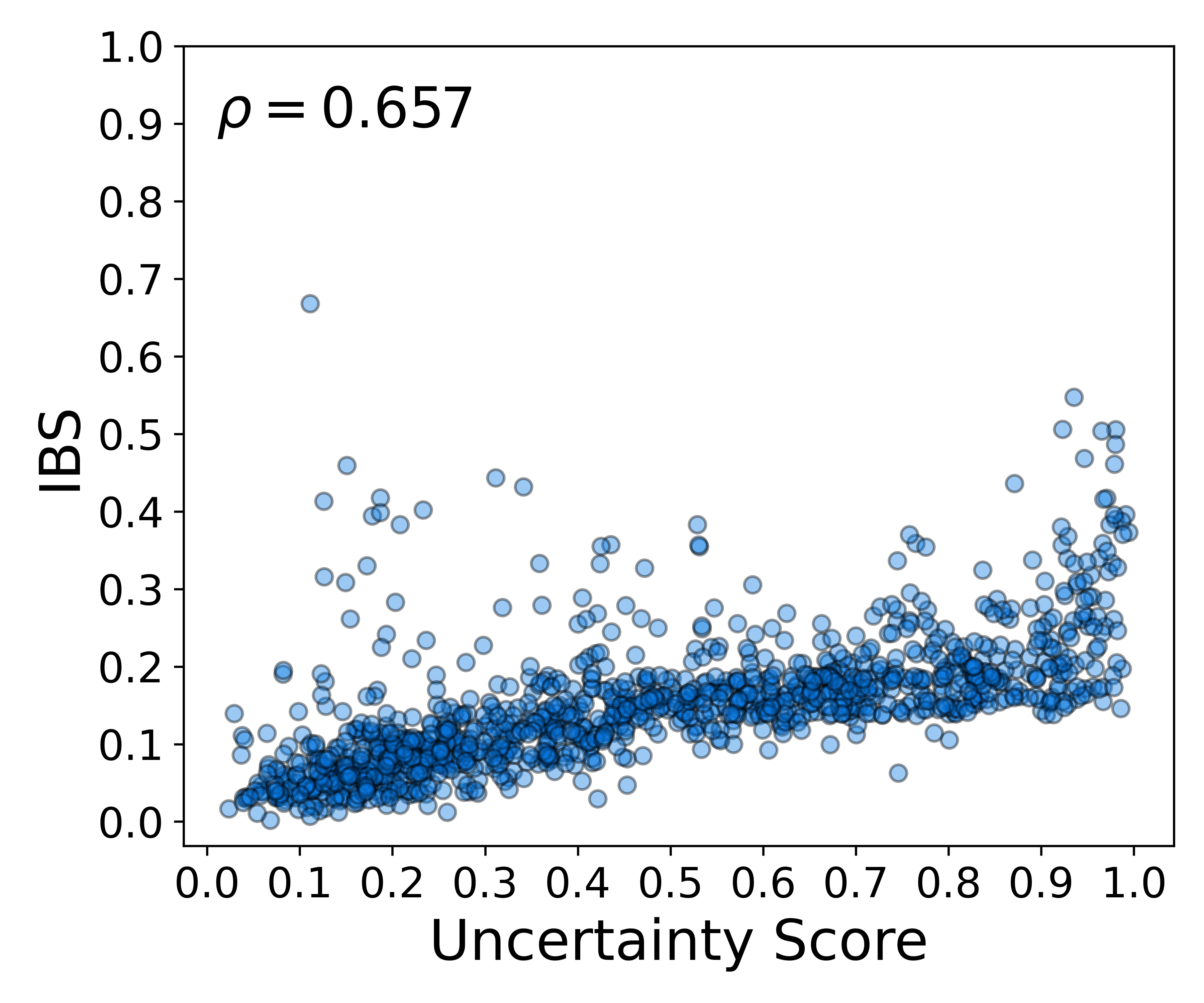}
        \caption{SUPPORT}
    \end{subfigure}
    \begin{subfigure}[b]{0.232\textwidth}
        \centering
        \includegraphics[width=\textwidth]{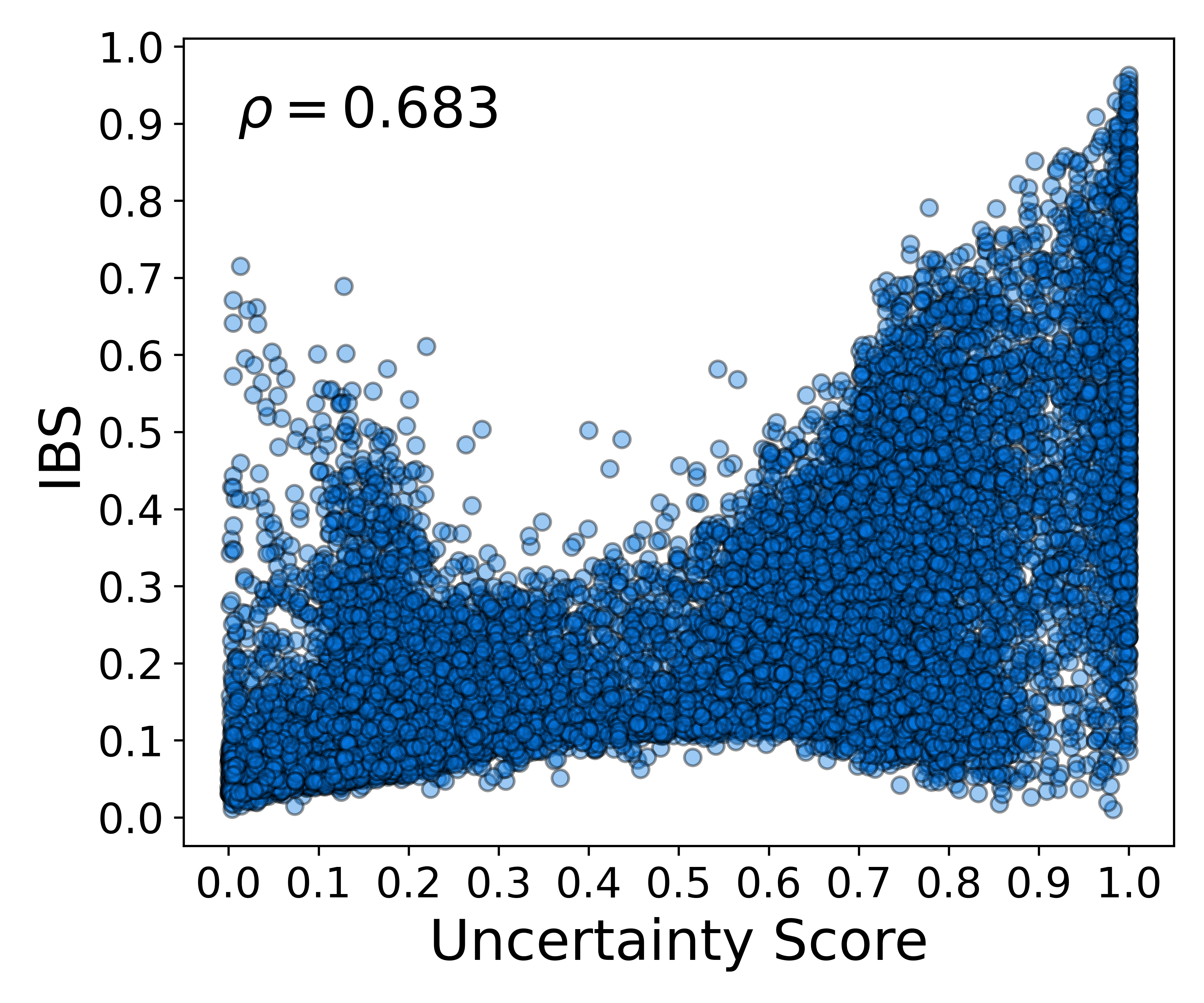}
        \caption{SEER-BC}
    \end{subfigure}
    \begin{subfigure}[b]{0.232\textwidth}
        \centering
        \includegraphics[width=\textwidth]{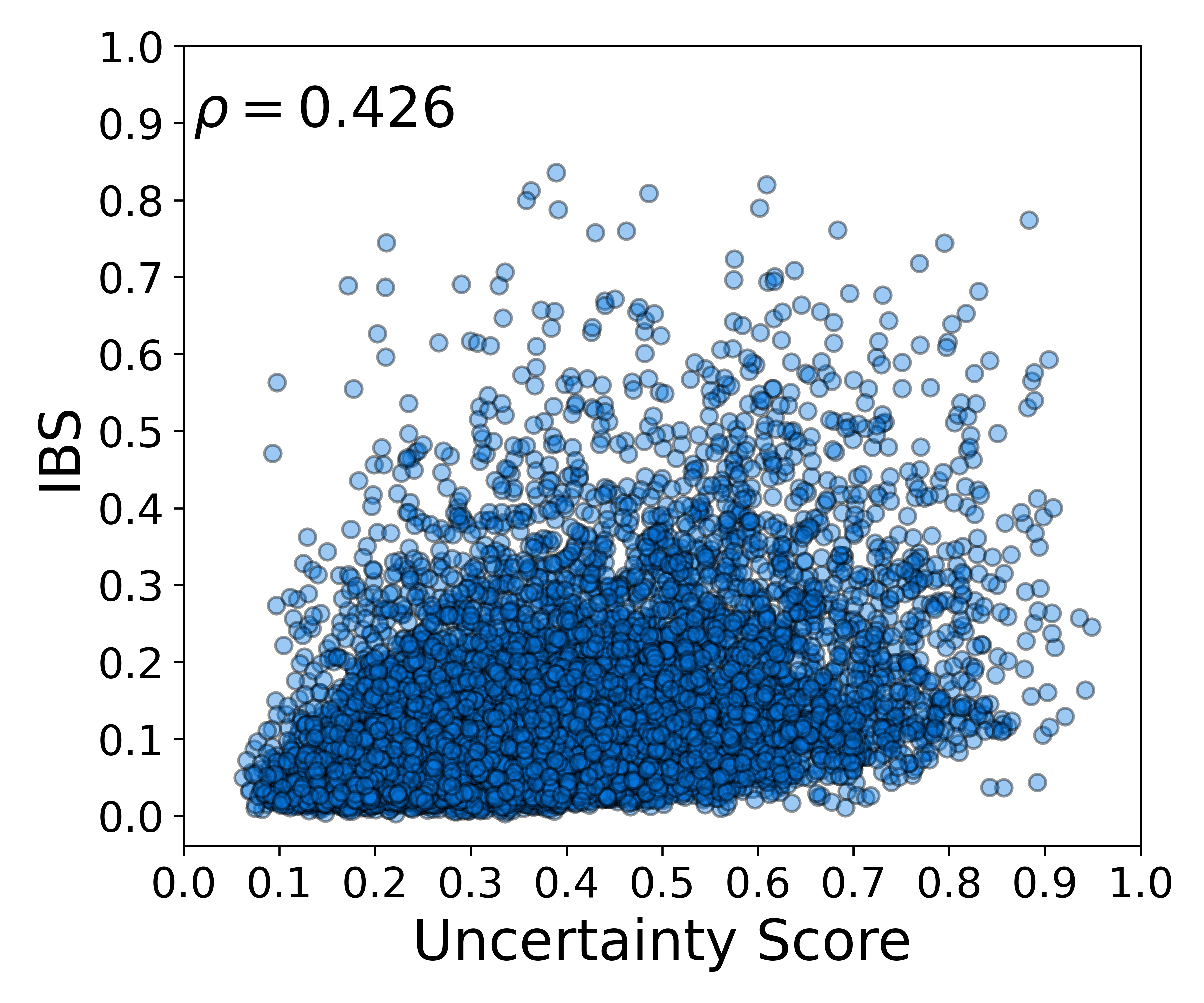}
        \caption{SAC3}
    \end{subfigure}
    \vspace{-5px}
    \caption{Predicted uncertainty scores versus IBSs from DeepSurv quantified by SurvUnc-RF across samples on (a) FLCHAIN, (b) SUPPORT, (c) SEER-BC and (d) SAC3 datasets.}
    \label{fig:misprediction_scatter}
\end{figure}

\subsection{Misprediction Detection (RQ1)}
We conduct the misprediction detection experiments to identify the correctly and incorrectly predicted samples using quantified uncertainty scores \cite{Shen2023}, where the mispredicted samples are viewed as in-distribution hard samples for survival models. 
Table~\ref{tab:misprediction} summarizes the Pearson correlation coefficients between the IBSs and quantified uncertainty scores from different uncertainty quantification methods for each survival model across four datasets. Results with BNNSurve are provided in Appendix~\ref{sec:app_results}. As observed, both SurvUnc-RF and SurvUnc-MLP achieve significantly higher and consistently positive correlations across all survival models and datasets, demonstrating their reliability in uncertainty quantification. In comparison, the performance of baselines is inconsistent, and several negative or weak correlation results are notable. The results indicate that traditional uncertainty quantification methods like MC-Dropout and Ensemble struggle to extend effectively to survival analysis, further demonstrating the effectiveness and importance of SurvUnc for survival analysis.

To further analyze the effectiveness of total uncertainty quantification, Figure~\ref{fig:misprediction_scatter} presents a comparison between predicted uncertainty scores and IBSs for the DeepSurv model using SurvUnc-RF on four datasets. The scatter plots show a clear positive correlation between quantified uncertainty and prediction performance. Notably, despite the different prediction distributions of IBSs across the four datasets, Our SurvUnc framework can effectively differentiate between relatively hard and easy in-distribution samples by assigning accurate uncertainty scores for each dataset.

\begin{table*}[htbp]
\centering
\setlength\tabcolsep{5pt}
\def\arraystretch{1.1}
\caption{OOD detection comparison with AUROC and AUPRC on the SEER dataset. Since the Bayesian-based model BNNSurv inherently provides uncertainty estimates and cannot be integrated with other survival models, the corresponding columns are left blank.}\label{tab:ood}
\vspace{-5px}
\begin{tabular}{c|cc|cc|cc|cc|cc}
\toprule
{\textbf{Models}} & \multicolumn{2}{c|}{\textbf{DeepSurv}} & \multicolumn{2}{c|}{\textbf{DeepHit}} & \multicolumn{2}{c|}{\textbf{DSM}} & \multicolumn{2}{c}{\textbf{RSF}} & \multicolumn{2}{c}{\textbf{BNNSurv}}  \\
\hline
{\textbf{UQ Methods}} & AUROC & AUPRC &  AUROC & AUPRC  &  AUROC & AUPRC &  AUROC & AUPRC &  AUROC & AUPRC \\
\hline
MC-Dropout & 0.358 & 0.398& 0.524 & 0.527 & 0.445 & 0.455& -& - & -& -\\ \hline
Ensemble & 0.493& 0.481& 0.499& 0.516& 0.558& 0.559& 0.516& 0.514 & -& -\\  \hline
Bayesian &  -  &   -    &  -   &    -  &   -   &  -    &  -   & -   &    0.483& 0.500 \\ \hline
SurvUnc-RF & 0.621 & 0.581 & 0.631 & 0.574 & 0.638 & 0.581 & 0.625 & \textbf{0.577} &  0.627 & 0.627\\ 
SurvUnc-MLP & \textbf{0.638} & \textbf{0.604} & \textbf{0.657} & \textbf{0.599} & \textbf{0.667} & \textbf{0.624} & \textbf{0.643} & \textbf{0.577}  & \textbf{0.636} &  \textbf{0.634}\\
\bottomrule
\end{tabular}
\end{table*}

\begin{figure}
    \centering
    \includegraphics[width=0.9\linewidth]{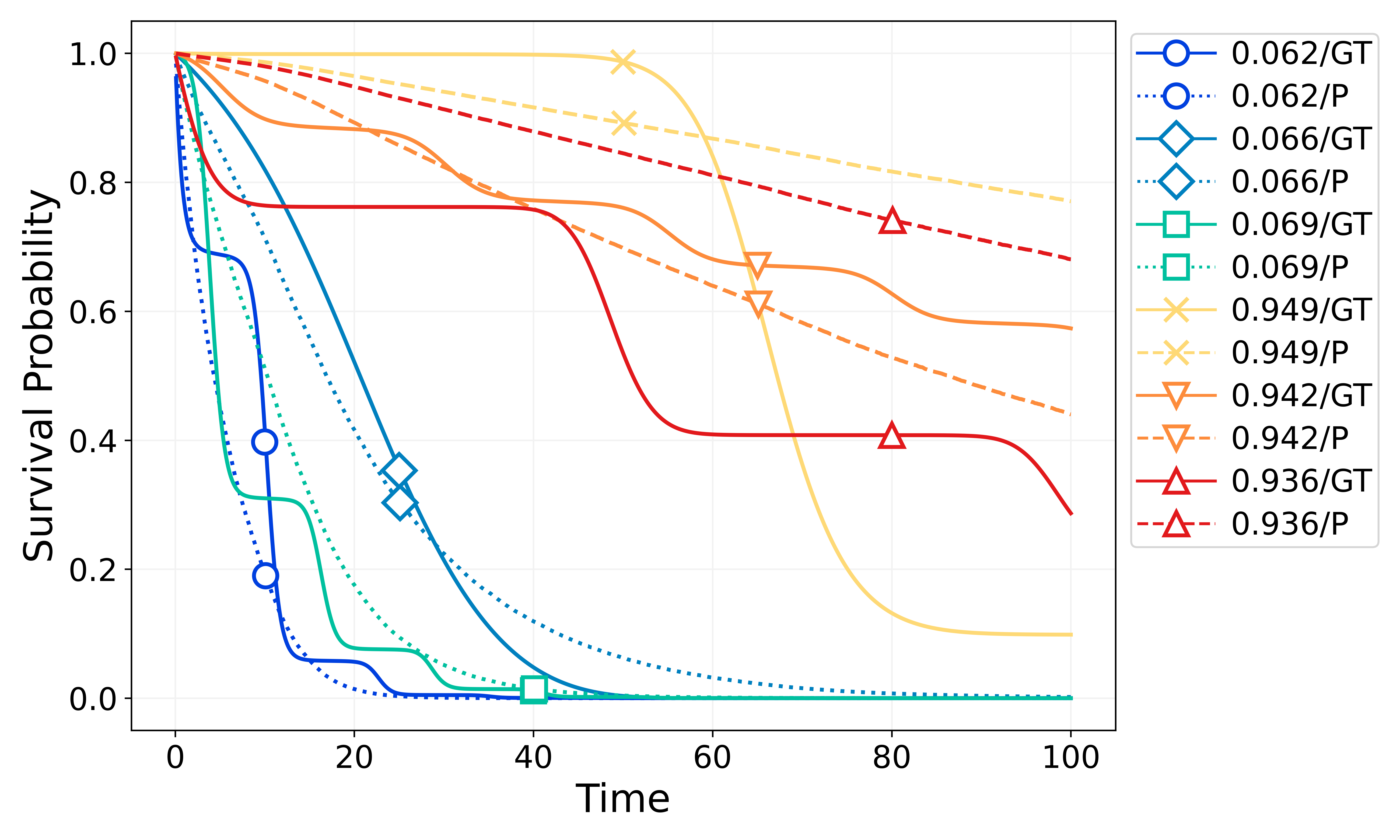}
    \vspace{-5px}
    \caption{Survival curve comparison of high-uncertainty and low-uncertainty samples on SAC3 dataset, quantified by SurvUnc-RF on DeepSurv. ``GT'' (with solid lines) and ``P'' (with dashed lines) denote ``Ground Truth'' and ``Predicted'', respectively, and the values in legend are uncertainty scores.}
    \label{fig:synthetic_surv_curves}
\end{figure}

As the absence of ground-truth survival curves poses a significant challenge to uncertainty quantification for survival models, we introduce the synthetic SAC3 dataset with ground-truth survival curves available for validation. Given a pretrained DeepSurv model on the SAC3 dataset, we select the three samples with the highest uncertainty and the three with the lowest uncertainty, as ordered by SurvUnc-RF, for visualization. As shown in Figure~\ref{fig:synthetic_surv_curves}, the predicted survival curves (in dashed lines with cold colors) for low-uncertainty samples closely match the ground-truth survival curves (in solid lines with cold colors). In contrast, for high-uncertainty samples, there is a marked discrepancy between the predicted and ground-truth survival curves. Our expanded analysis confirms that samples with high uncertainty predictions generally exhibit longer survival times compared to those with low uncertainty. The reason is current survival models tend to accumulate prediction errors when estimating long-term survival patterns, resulting in higher uncertainty estimates. These results further highlight the effectiveness of SurvUnc in quantifying total uncertainty associated with survival curves.

\subsection{OOD Detection (RQ2)}
In addition to the quantification of total uncertainty, the quantification of epistemic uncertainty is emphasized in reliable deployment to detect OOD data \cite{Shen2023,yu2024discretization}. Here, we conduct several ODD detection experiments to validate SurvUnc. As mentioned, we develop the SEER-HD dataset as the OOD counterpart to the SEER-BC dataset. The SEER-HD dataset includes the same number of uncensored patients as the SEER-BC dataset but comprises patients with a different condition, specifically heart disease. For age at diagnosis, SEER-BC patients' mean age is 60.52 (SD: 15.13) compared to SEER-HD patients with 74.84 (SD: 10.40). For time-to-event distribution, SEER-BC patients' mean duration is 41.67 (SD: 29.91), compared to SEER-HD patients with 55.44 (SD: 33.34). These differences are statistically significant (p<0.001) under Wilcoxon rank-sum test.

\begin{figure}[htbp]
    \centering
    \begin{subfigure}[b]{0.225\textwidth}
        \centering
        \includegraphics[width=\textwidth]{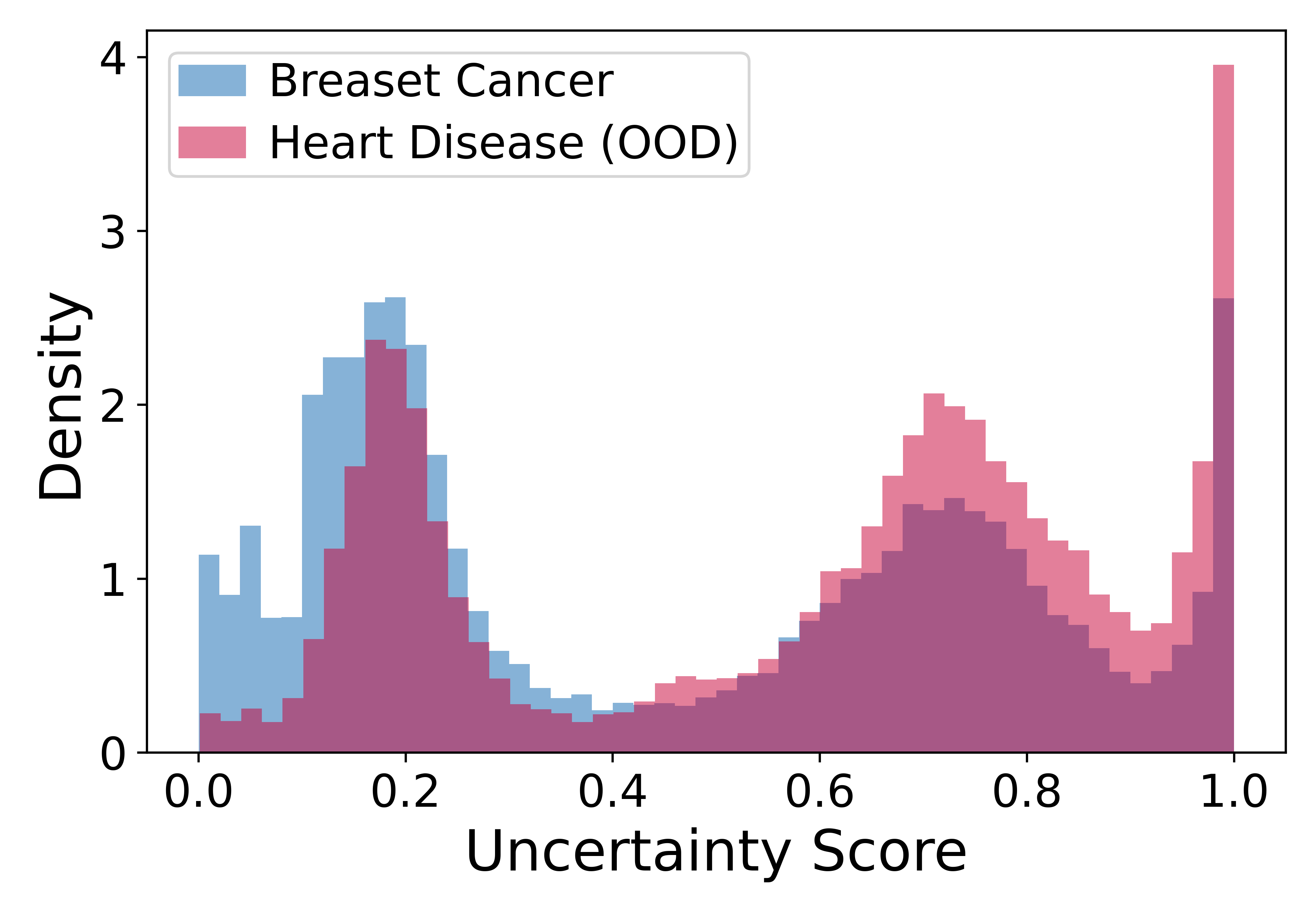}
        \caption{SurvUnc-RF}
    \end{subfigure}
    \begin{subfigure}[b]{0.225\textwidth}
        \centering
        \includegraphics[width=\textwidth]{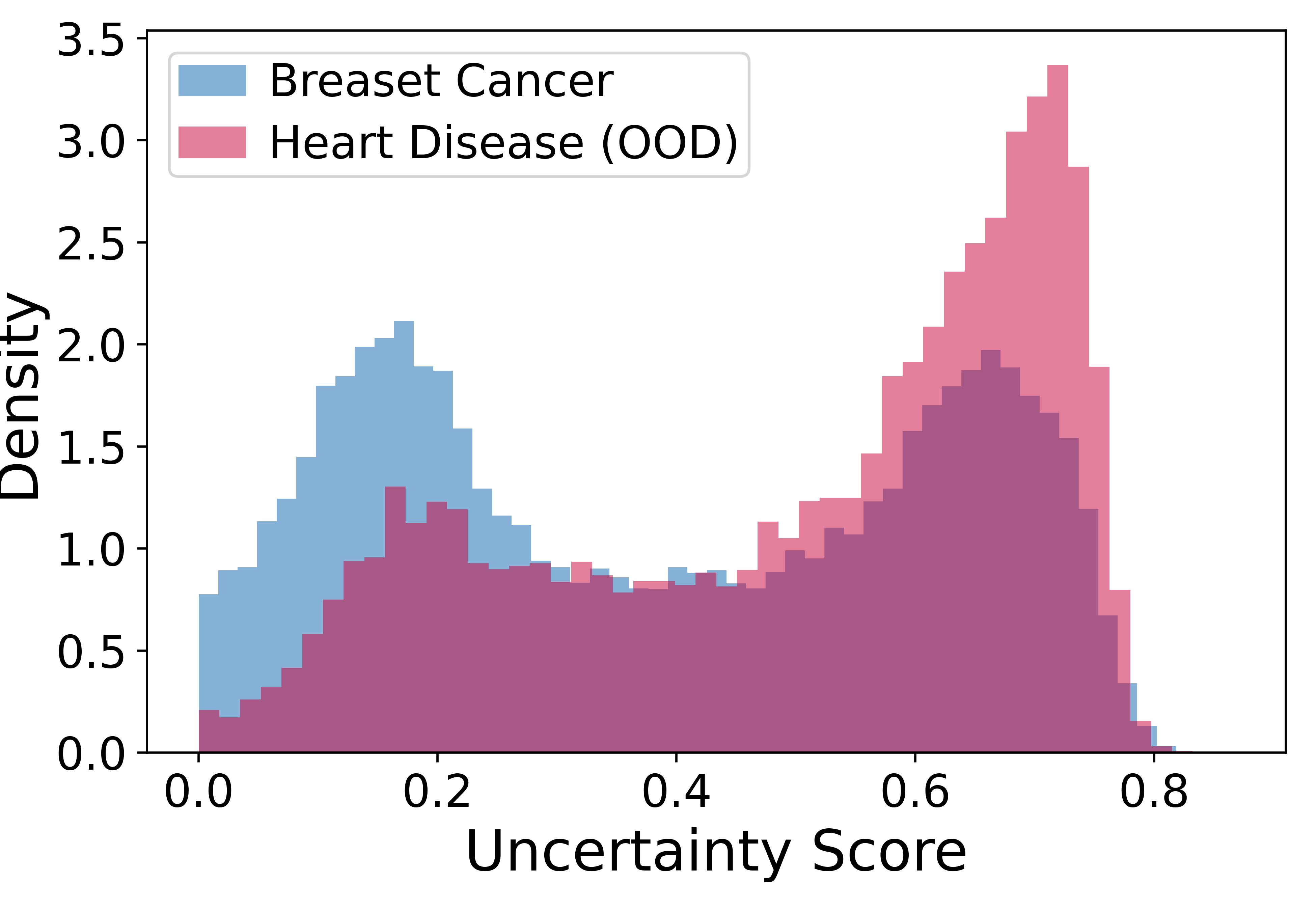}
        \caption{SurvUnc-MLP}
    \end{subfigure}
    \begin{subfigure}[b]{0.225\textwidth}
        \centering
        \includegraphics[width=\textwidth]{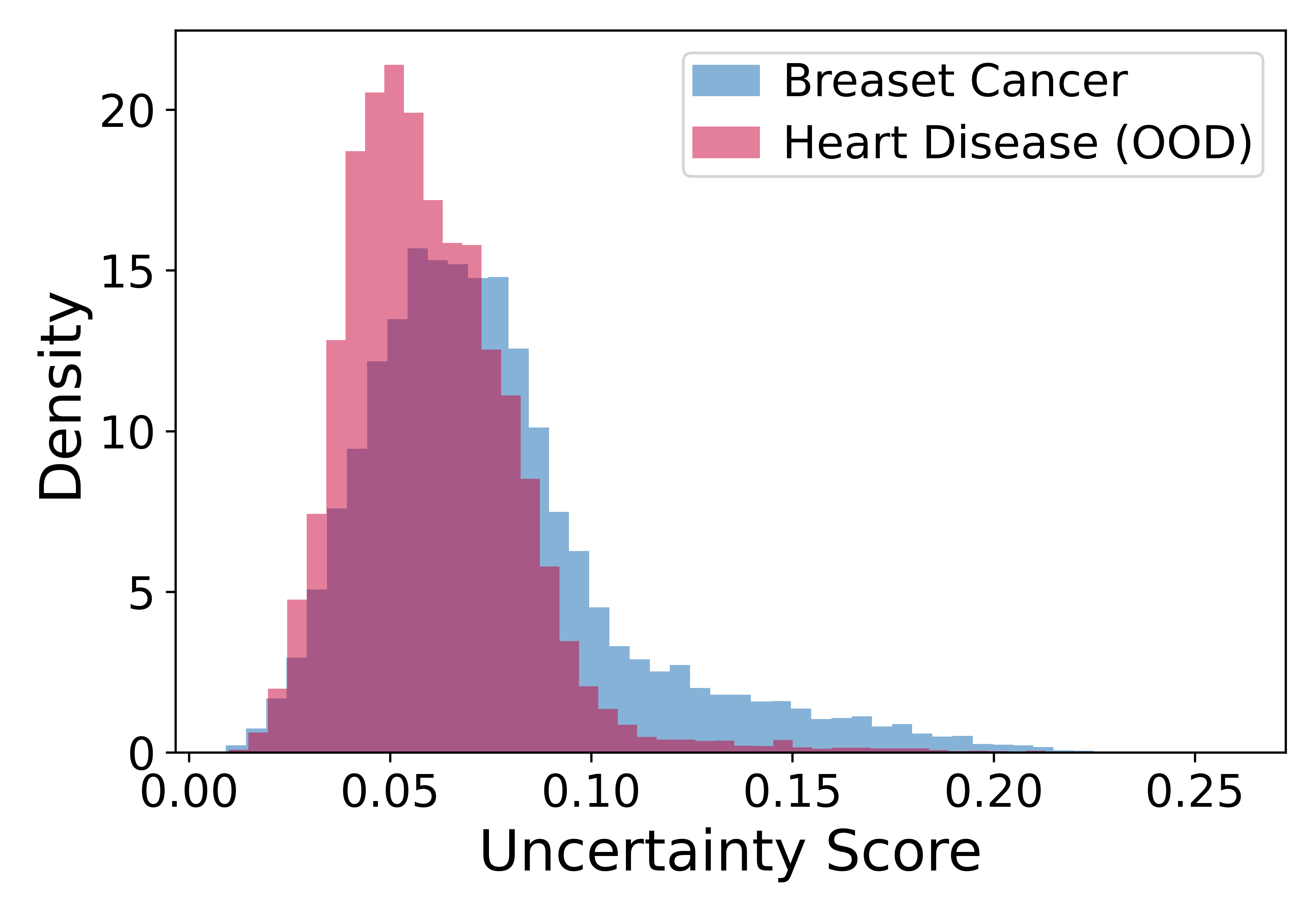}
        \caption{MC-Dropout}
    \end{subfigure}
    \begin{subfigure}[b]{0.225\textwidth}
        \centering
        \includegraphics[width=\textwidth]{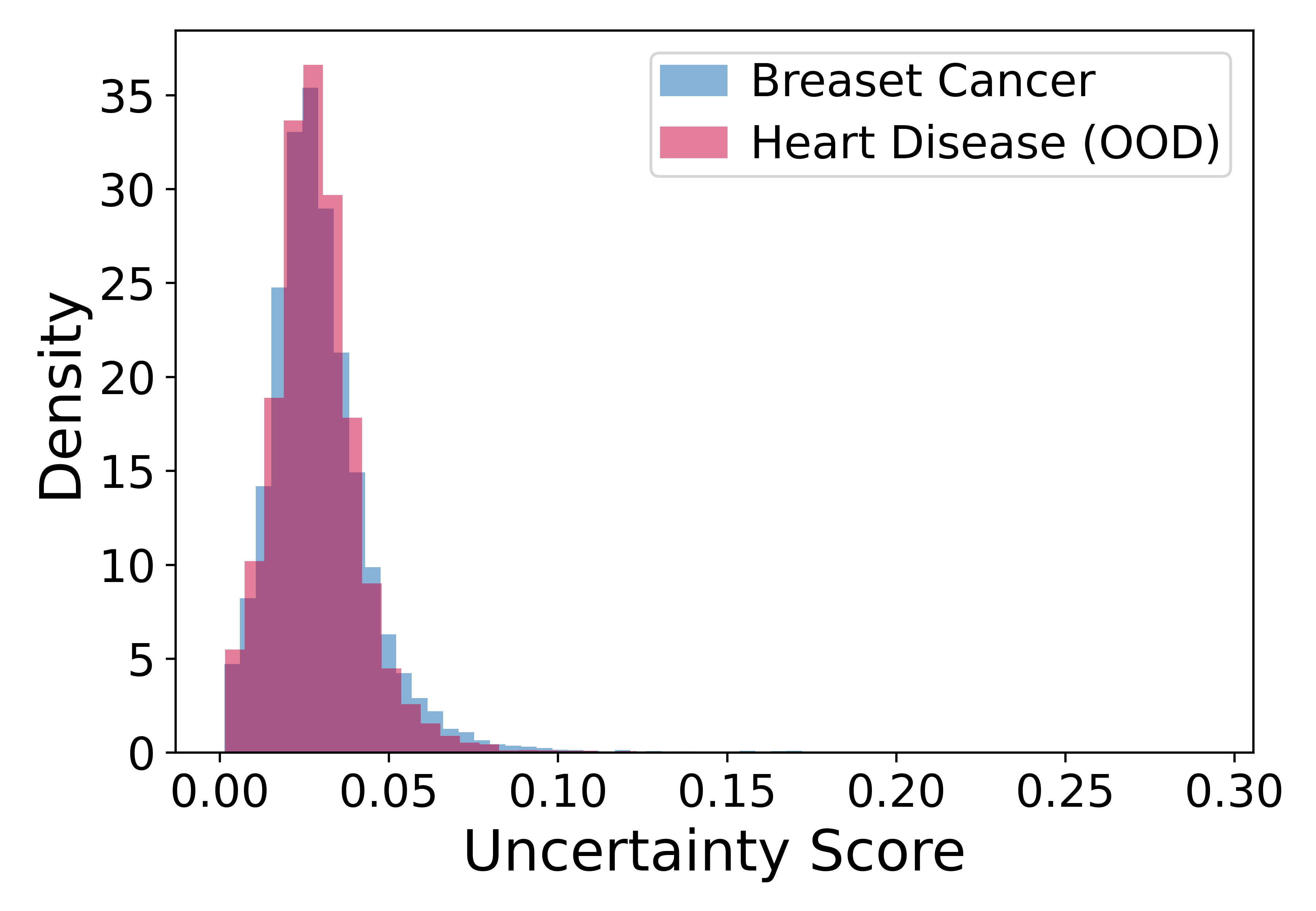}
        \caption{Ensemble}
    \end{subfigure}
    \vspace{-5px}
    \caption{Uncertainty score distribution comparison of DeepSurv between IND (BC) and OOD (HD) samples from SEER dataset, with uncertainty scores from (a) SurvUnC-RF, (b) SurvUnc-MLP, (c) MC-Dropout and (d) Ensemble.}
    \label{fig:ood_distribution}
\end{figure}

Firstly, we investigate whether the OOD data can be identified from the uncertainty score distribution. Based on DeepSurv, Figure~\ref{fig:ood_distribution} compares the uncertainty score distributions of both datasets across different quantification methods. The findings suggest that both SurvUnc-RF (Figure~\ref{fig:ood_distribution}(a)) and SurvUnc-MLP (Figure~\ref{fig:ood_distribution}(b)) can generally sense the OOD and IND data differently. Notably, the uncertainty score distribution for OOD data (shown in pink) is right-shifted in comparison to that of IND data (shown in blue), indicating higher uncertainty for more OOD samples. However, Ensemble (Figure~\ref{fig:ood_distribution}(d)) fails to distinguish between the two distributions, as they appear nearly identical. For MC-Dropout, IND data is generally quantified with even higher uncertainty scores than OOD data, highlighting its limitations in OOD detection.

Furthermore, we provide a quantitative evaluation of the OOD detection task in Table~\ref{tab:ood}. By using uncertainty scores as logits to classify IND and OOD samples, SurvUnc demonstrates a significant performance improvement over baseline methods across all survival models. In particular, SurvUnc-MLP shows notable enhancements, achieving an 19.5\% -31.7\% increase in AUROC and an 11.6\%-26.8\% improvement in AUPRC. This epistemic uncertainty quantification capability can be attributed to the anchor-based learning strategy, which successfully infuses in-domain knowledge into the meta-model learning process. 

\subsection{Hyperparameter Study (RQ3)}
Our proposed SurvUnc framework consists of two important hyperparameters, i.e., the meta-model structure and the number of anchors. In terms of the meta-model structure, we explore the random forest and MLP in former experiments, where both meta-models achieve consistently better performance than baselines with robustness achieved. 

We investigate the impact of varying the number of anchors on selective prediction performance ($C^\text{td}$), as shown in Figure~\ref{fig:param_1_selective}. Additional results with SurvUnc-MLP, provided in Appendix~\ref{sec:app_results}, exhibit similar trends. The anchor-based learning strategy suggests that increasing the number of anchors can enhance the robustness of label calculation for the meta-model training set construction, thereby improving performance. For each discarding percentage, the performance stabilizes when using 10 or more anchors\footnote{The slight variations are due to randomness in training.}. This indicates that the proposed SurvUnc framework can achieve efficient performance with a relatively small number of anchors.

\begin{figure}[htbp]
    \centering
    \includegraphics[width=0.5\textwidth]{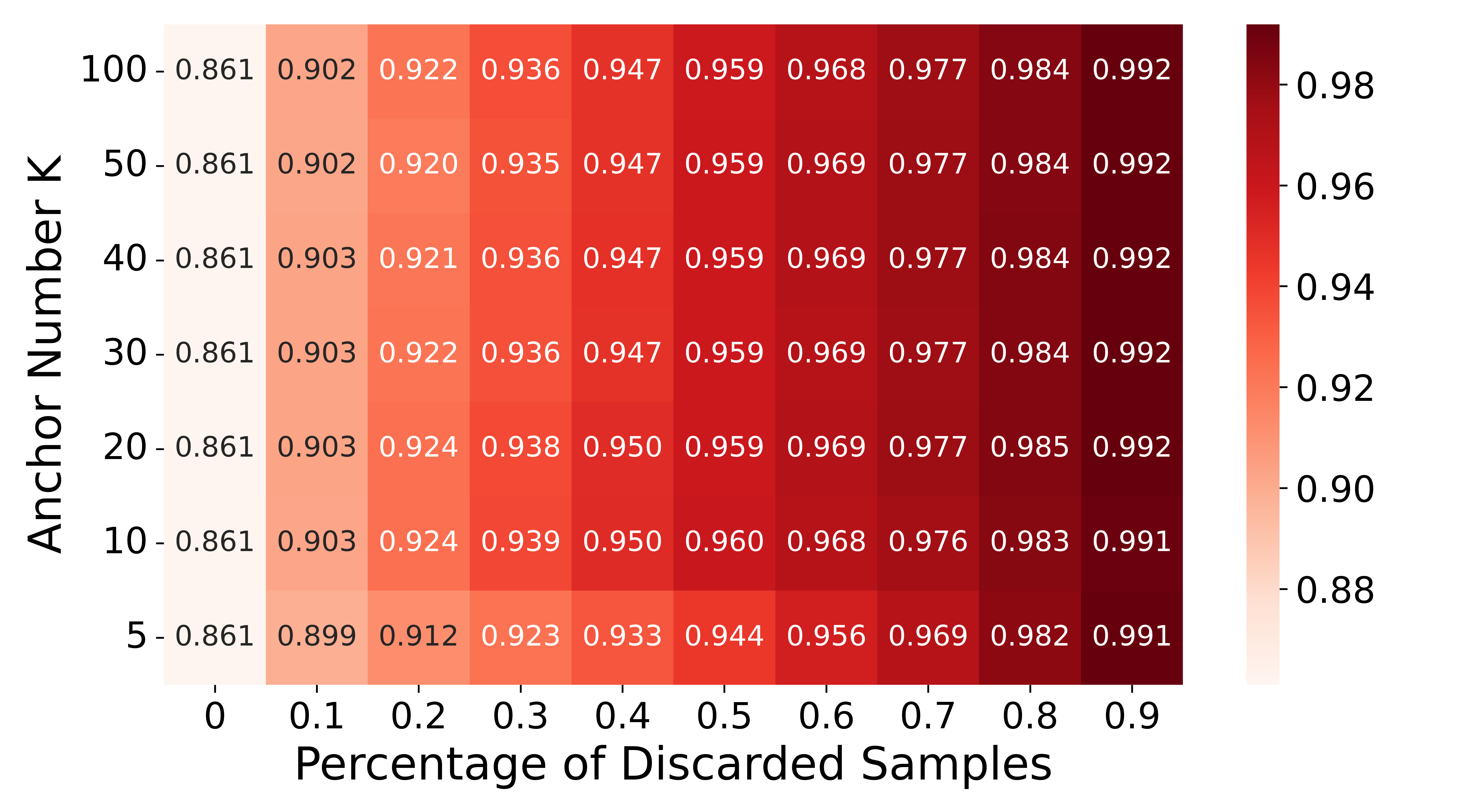}
    \vspace{-5px}
    \caption{Varying anchor number $K$ on selective prediction performance of SurvUnc-RF with DeepSurv on SEER-BC.}
    \label{fig:param_1_selective}
\end{figure}

\section{Conclusion}
In this paper, we address the novel challenge of uncertainty quantification in survival analysis and propose SurvUnc, a meta-model based framework that incorporates an anchor-based learning strategy. We establish systematic evaluation protocols to assess uncertainty quantification methods on survival models. Extensive experiments conducted across multiple datasets and survival models demonstrate the effectiveness and robustness of SurvUnc in uncertainty quantification. Furthermore, its model-agnostic design ensures compatibility with a wide range of survival models, offering valuable insights into uncertainty quantification for survival analysis. Future work will explore uncertainty quantification for survival analysis with competing events and time-varying covariates, as well as evaluate SurvUnc in the context of foundation model-based survival analysis \cite{jeanselme2024review,gu2025time}.

\begin{acks}
Some aspects of this work were funded through an award from the NIHR (AI award 02316). Tingting Zhu was supported by the Royal Academy of Engineering under the Research Fellowship scheme.
\end{acks}

\bibliographystyle{ACM-Reference-Format}
\bibliography{sample-base}

\appendix

\section{Experiment Setup}\label{sec:app_exp}

\subsection{Survival Model Details}
To quantify the uncertainty for survival models, we first pretrain several survival models to be quantified. Specifically, the implementations of DeepSurv \cite{deepsurv} and DeepHit \cite{deephit} are sourced from \texttt{pycox} package. The implementation of RSF is from \texttt{scikit-survival} package\footnote{\url{https://scikit-survival.readthedocs.io/en/latest/index.html}}, and the implementation of DSM is from \texttt{auton\_survival} package\footnote{\url{https://autonlab.org/auton-survival/}}. The implementation of BNNSurv is from \texttt{bnnsurv} package\footnote{\url{https://github.com/thecml/UE-BNNSurv}}. We summarize the hyperparameter settings of such models in Table~\ref{tab:hyper-deepsurv}-\ref{tab:hyper-rsf}. Detailed implementation codes are provided in the anonymous link. Since our primary focus is on uncertainty quantification rather than survival prediction, we tune the hyperparameters of the survival models to match the performance levels reported in existing works. 

\begin{table}[htbp]
    \centering
    \def\arraystretch{1.}
    \caption{Hyperparameters for DeepSurv model.}\label{tab:hyper-deepsurv}
    \vspace{-10px}
    \begin{tabular}{c|c|c|c}
     \toprule
     \textbf{Dataset}    & \textbf{learning rate} & \textbf{dropout} & \textbf{hidden layers} \\
     \hline
      FLCHAIN   & 0.01 & 0.1  & [32] \\
      \hline
      SUPPORT   & 0.1 &  0.1 & [32]\\
      \hline
      SEER-BC   & 0.01 & 0.1 & [32] \\
      \hline
       SAC3  & 0.01 & 0.1 & [32]\\
     \bottomrule
    \end{tabular}
\end{table}

\begin{table}[htbp]
\vspace{-10px}
    \centering
    \def\arraystretch{1.}
    \caption{Hyperparameters for DeepHit model. $d_\text{in}$ is determined by the dataset covariate dimensions.}\label{tab:hyper-deephit}
    \vspace{-10px}
    \begin{tabular}{c|c|c|c}
     \toprule
     \textbf{Dataset}    & \textbf{learning rate} & \textbf{dropout} & \textbf{hidden layers} \\
     \hline
      FLCHAIN   & 0.001 & 0.1  & [$3d_\text{in}$, $5d_\text{in}$, $3d_\text{in}$] \\
      \hline
      SUPPORT   & 0.005 &  0.1 & [$3d_\text{in}$, $5d_\text{in}$, $3d_\text{in}$]\\
      \hline
      SEER-BC   & 0.001 & 0.1 & [$3d_\text{in}$, $5d_\text{in}$, $3d_\text{in}$] \\
      \hline
       SAC3  & 0.001 & 0.1 & [$3d_\text{in}$, $5d_\text{in}$,$3d_\text{in}$]\\
     \bottomrule
    \end{tabular}
\end{table}

\begin{table}[htbp]
    \centering
    \def\arraystretch{1}
    \caption{Hyperparameters for DSM model.}\label{tab:hyper-dsm}
    \vspace{-10px}
    \begin{tabular}{c|c|c|c}
     \toprule
     \textbf{Dataset}  & \textbf{learning rate} & \textbf{Distribution, No.} & \textbf{hidden layers} \\
     \hline
      FLCHAIN   & 0.005 & Weibull, 4 & [32, 32] \\
      \hline
      SUPPORT   & 0.005 &  Weibull, 6 & [32, 32]\\
      \hline
      SEER-BC   & 0.005 & Weibull, 6 & [32, 32] \\
      \hline
       SAC3  & 0.005 & Weibull, 10 & [32, 32, 32]\\
     \bottomrule
    \end{tabular}
\end{table}

\begin{table}[htbp]
    \centering
    \setlength\tabcolsep{2pt}
    \def\arraystretch{1.}
    \caption{Hyperparameters for RSF model.}\label{tab:hyper-rsf}
    \vspace{-10px}
    \begin{tabular}{c|c|c|c}
     \toprule
     \textbf{Dataset}  & \textbf{n\_estimators} & \textbf{min\_samples\_split} & \textbf{min\_samples\_leaf} \\
     \hline
      FLCHAIN   & 100 & 20 &  5\\
      \hline
      SUPPORT   & 100 & 20 & 5\\
      \hline 
      SEER-BC   & 100 & 20 & 5 \\
      \hline
       SAC3  & 20 & 20 & 5\\
     \bottomrule
    \end{tabular}
\end{table}

\subsection{Baseline Details}
For each testing sample, MC-Dropout quantifies uncertainty by activating dropout layers in the survival models during inference and calculating the maximum standard deviation of predicted survival probabilities across time steps over 100 forward passes. For the Ensemble method, we train the survival model with 10 different random initializations and use the maximum standard deviation of predicted survival probabilities across time steps from the 10 models as the uncertainty measure. It's worth noting that we also explored alternative approaches, such as calculating the mean standard deviation and the mean Euclidean distance between predicted survival probability time vectors to assess the discrepancy among multiple predictions for the same sample. However, these alternatives performed relatively worse compared to using the maximum standard deviation.

\section{Additional Experiment Results}\label{sec:app_results}
\subsection{Survival Model Validation}
In Table~\ref{tab:survival_model}, we compare the performance of our implemented survival models with the reported performance on FLACHAIN and SUPPORT datasets from existing works. The results are sourced from papers \cite{deephit,deepsurv,Sun2021,Kvamme2019,kvamme2021continuous,zhong2021deep,dsm,Cui2024}. Due to variations in dataset splits, the reported performance in the literature is not exactly the same. To account for this, we present the available performance range for comparison.

\begin{table}[htbp]
\centering
\def\arraystretch{1.}
\caption{Performance comparison of survival models with reported results in literature (Lit.).}\label{tab:survival_model}
\vspace{-10px}
\begin{tabular}{c|c|cc|cc}
\toprule
\multirow{2}{*}{\textbf{Model}} & \multirow{2}{*}{\textbf{From}}  & \multicolumn{2}{c|}{\textbf{FLCHAIN}}  & \multicolumn{2}{c}{\textbf{SUPPORT}} \\
\cline{3-6}
 &  & \textbf{C-index} & \textbf{IBS} & \textbf{C-index} & \textbf{IBS} \\
\hline
\multirow{2}{*}{DeepHit} & Lit.  & 0.79-0.80 & 0.09-0.13  & 0.56-{0.64} &  0.20-0.23 \\
 & Ours  & 0.797 & 0.125  & 0.626 & 0.209  \\
\hline
\multirow{2}{*}{DeepSurv} & Lit.  & 0.79-0.80& 0.09-0.10 & 0.57-0.61 & 0.19-0.21\\
 & Ours    & 0.798 & 0.101 & 0.606 & 0.195 \\
\hline
\multirow{2}{*}{RSF} & Lit.  & 0.77-0.82 & 0.09-0.10  & 0.61-0.63 & 0.18-0.21 \\
 & Ours  & 0.795 & 0.100 & 0.631 & 0.189    \\
\hline
\multirow{2}{*}{DSM} & Lit.  & 0.79-0.80 &  0.10-0.11  & 0.60-0.61 & 0.20-0.21 \\
 & Ours     & 0.799 & 0.102  & 0.610 & 0.197\\
\bottomrule
\end{tabular}
\end{table}

\subsection{Additional Results on Selective Prediction}
\begin{figure*}[htbp]
    \centering
    \begin{subfigure}[b]{0.246\textwidth}
        \centering
        \includegraphics[width=\textwidth]{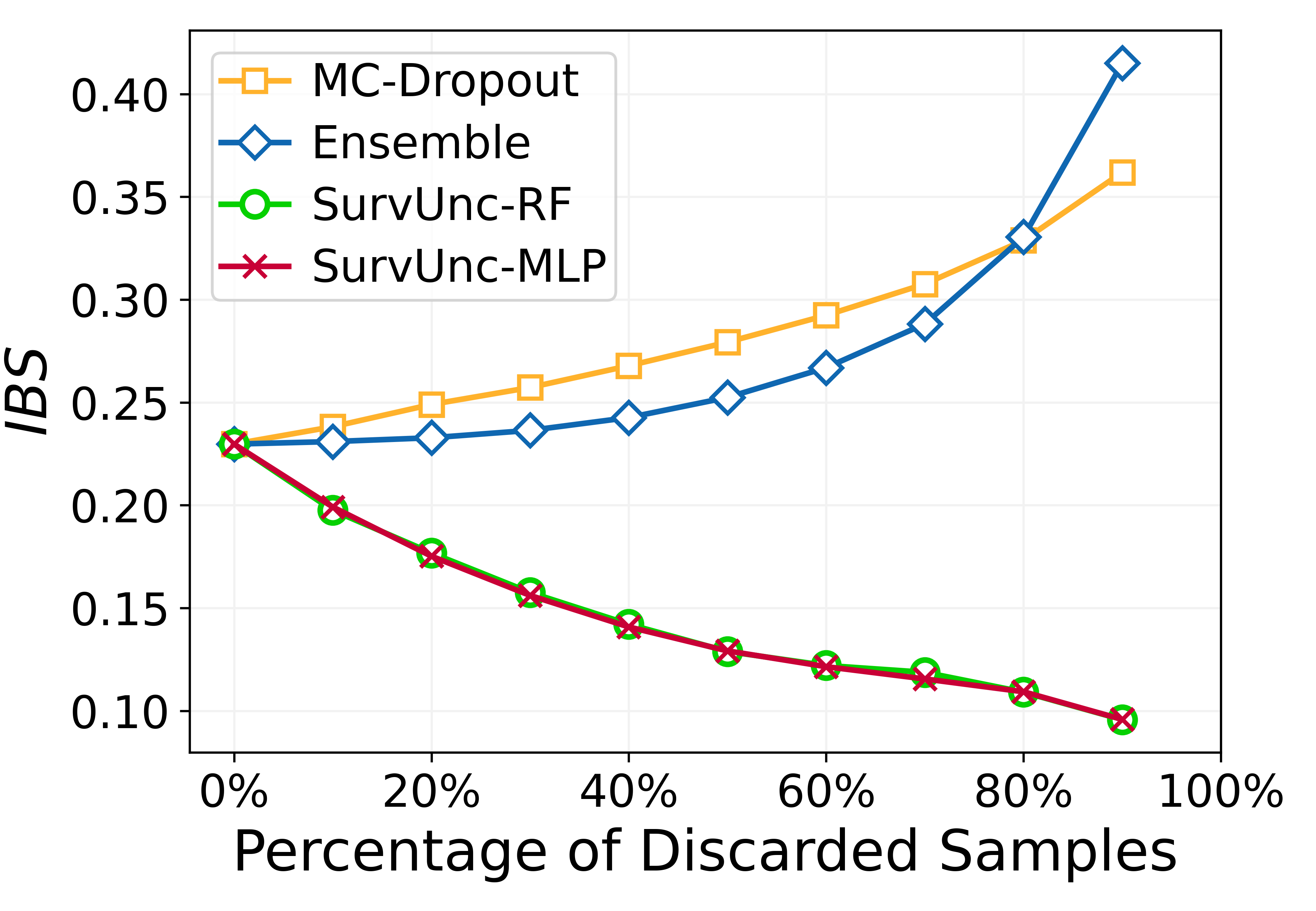}
        \caption{DeepSurv}
    \end{subfigure}
    \hfill
    \begin{subfigure}[b]{0.246\textwidth}
        \centering
        \includegraphics[width=\textwidth]{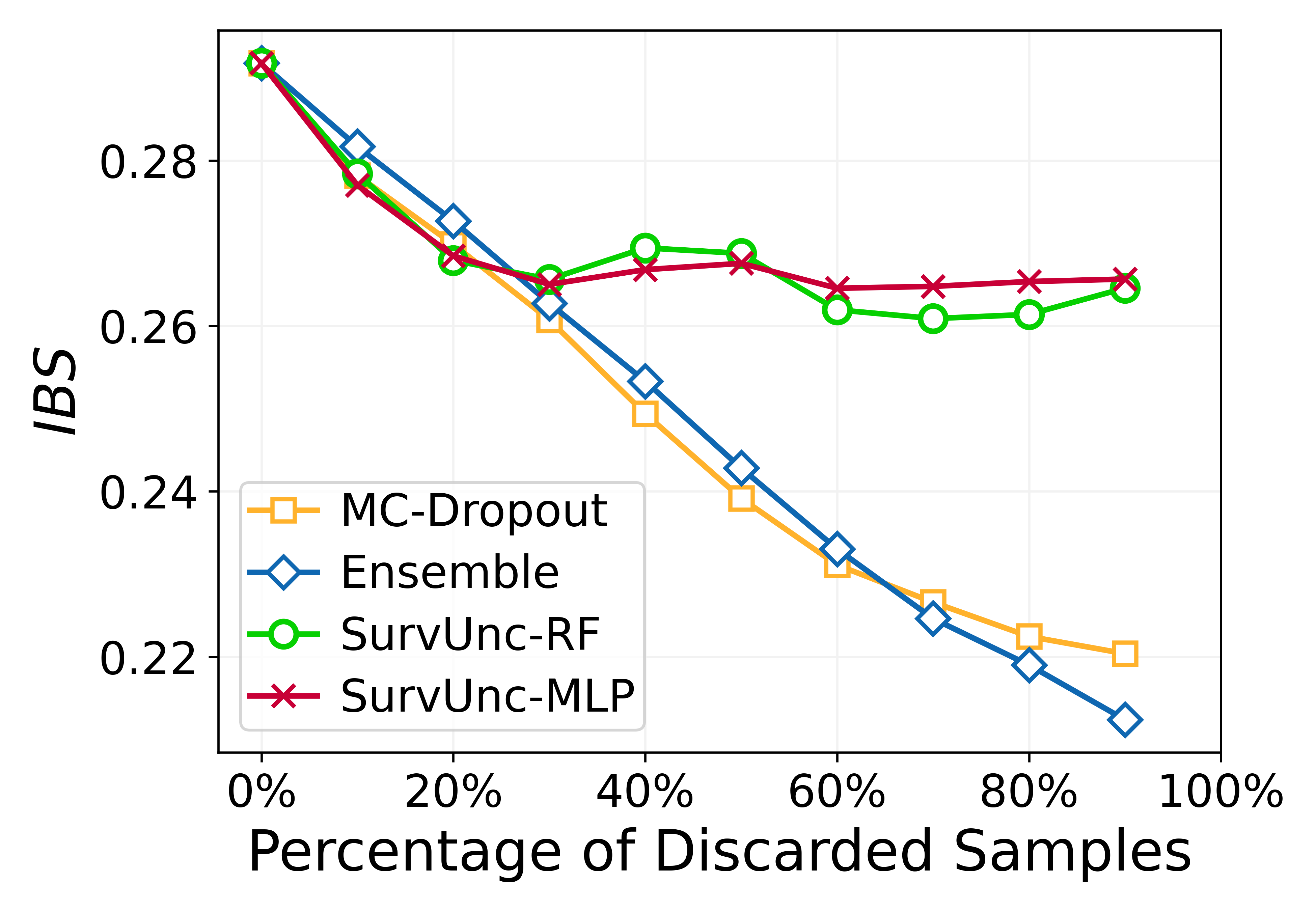}
        \caption{DeepHit}
    \end{subfigure}
    \hfill
    \begin{subfigure}[b]{0.246\textwidth}
        \centering
        \includegraphics[width=\textwidth]{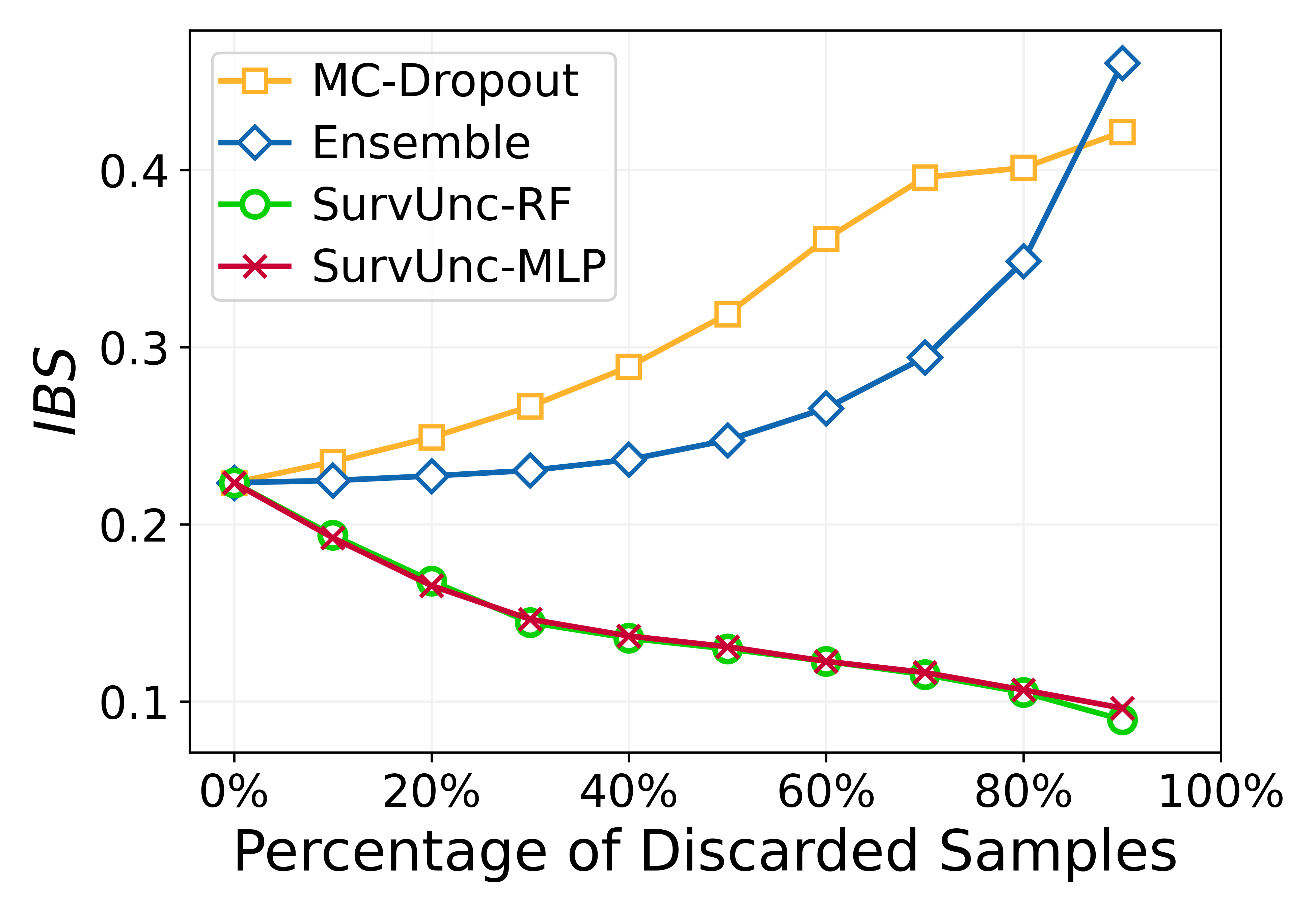}
        \caption{DSM}
    \end{subfigure}
    \hfill
    \begin{subfigure}[b]{0.246\textwidth}
        \centering
        \includegraphics[width=\textwidth]{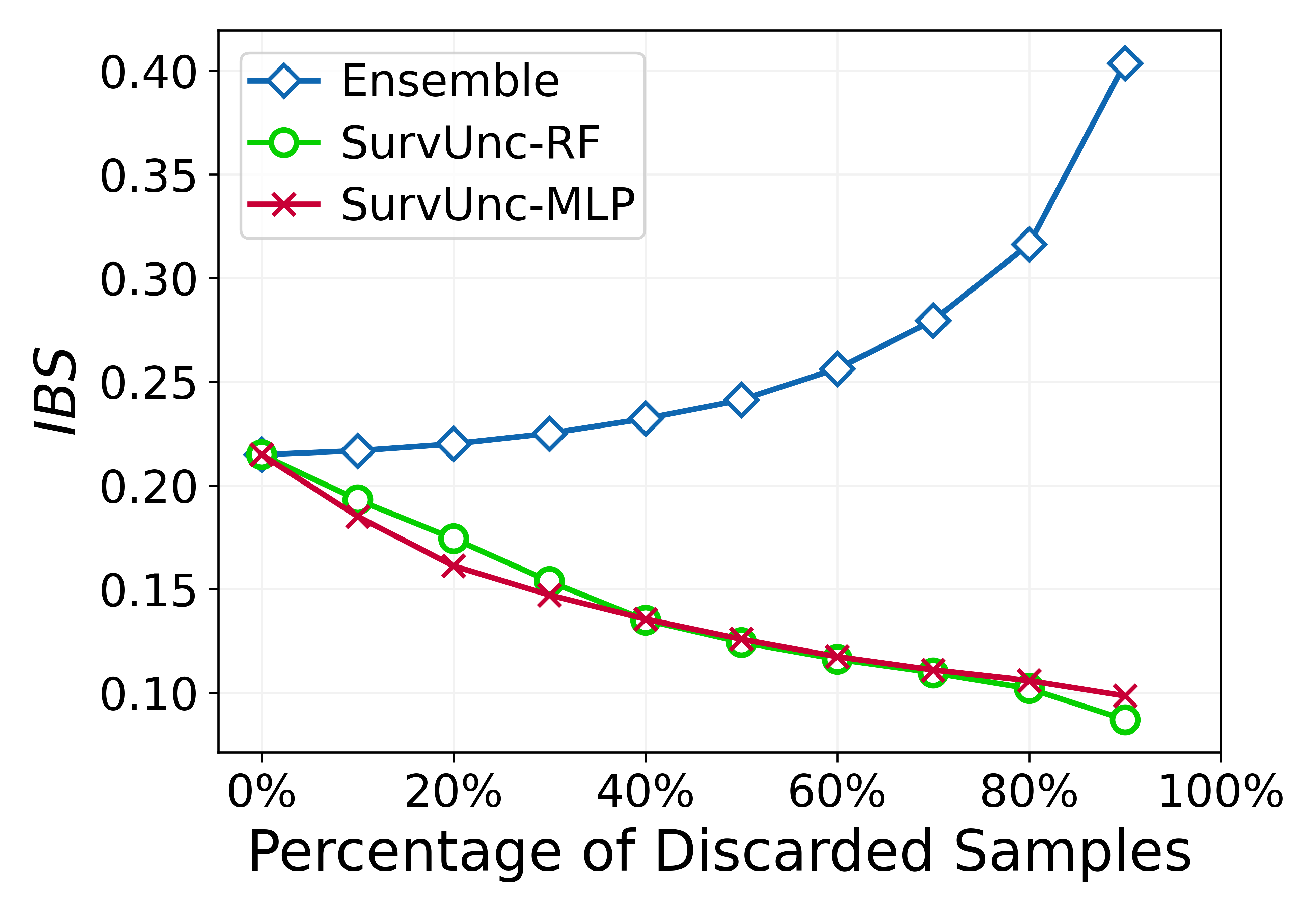}
        \caption{RSF}
    \end{subfigure}
    \vspace{-10px}
    \caption{IBS of four survival models of (a) DeepSurv, (b) DeepHit, (c) DSM and (d) RSF on SEER-BC dataset with different percentages of samples discarded according to uncertainty scores from different uncertainty quantification methods.}
    \label{fig:selective_prediction_ibs}
\end{figure*}

Table~\ref{tab:selective_cindex_app} is a supplemented table for Table~\ref{tab:selective_cindex}. We also report the IBS comparison in Table~\ref{tab:selective_ibs_app} and Figure~\ref{fig:selective_prediction_ibs}. Except the results with DeepHit model, similar trends to $C^\text{td}$ can be observed for IBS, which further demonstrate the superiority of SurvUnc framework. 

In addition, the Brier score is know to be imperfect in survival analysis since it approximates the ground truth survival curve as a step function that begins at 1 and then immediately becomes 0 at the time to event. Thus, we compute the absolute difference between predicted median survival ($\min\{t|S(t|\bm{x})\leq0.5\}$) and actual event times for uncensored patients in selective prediction on SEER-BC and SUPPORT, as shown in Table~\ref{tab:selective_time_diff}. Our proposed SurvUnc still achieves the best performance.

\begin{table}[htbp]
\centering
\def\arraystretch{1.}
\caption{Absolute difference between predicted median survival and actual event times of DeepSurv under different discarding percentages (10\%, 30\%, 50\%), determined by the SurvUnc framework with different UQ methods across datasets.}\label{tab:selective_time_diff}
\vspace{-10px}
\begin{tabular}{c|ccc|ccc}
\toprule
{\textbf{Datasets}} & \multicolumn{3}{c|}{\textbf{SUPPORT}} & \multicolumn{3}{c}{\textbf{SEER-BC}}   \\
\hline
{\textbf{UQ Methods}} & 10\% & 30\% & 50\% & 10\% & 30\% & 50\% \\
\hline
MC-Dropout & 289.50 & 320.36 & 323.45 & 41.21 & 44.18 & 47.22 \\
Ensemble & 282.52 & 299.62 & 314.78 & 39.64 & 39.73 & 41.29 \\
SurvUnc-RF & 242.70 & 203.24 & 166.85 & 36.29 & 28.39 & 21.75 \\
SurvUnc-MLP & 243.22 & 203.77 & 169.99 & 36.40 & 27.79 & 21.81 \\
\bottomrule
\end{tabular}
\end{table}

\subsection{Additional Results on Misprediction}
\begin{figure*}[htbp]
    \centering
    \begin{subfigure}[b]{0.232\textwidth}
        \centering
        \includegraphics[width=\textwidth]{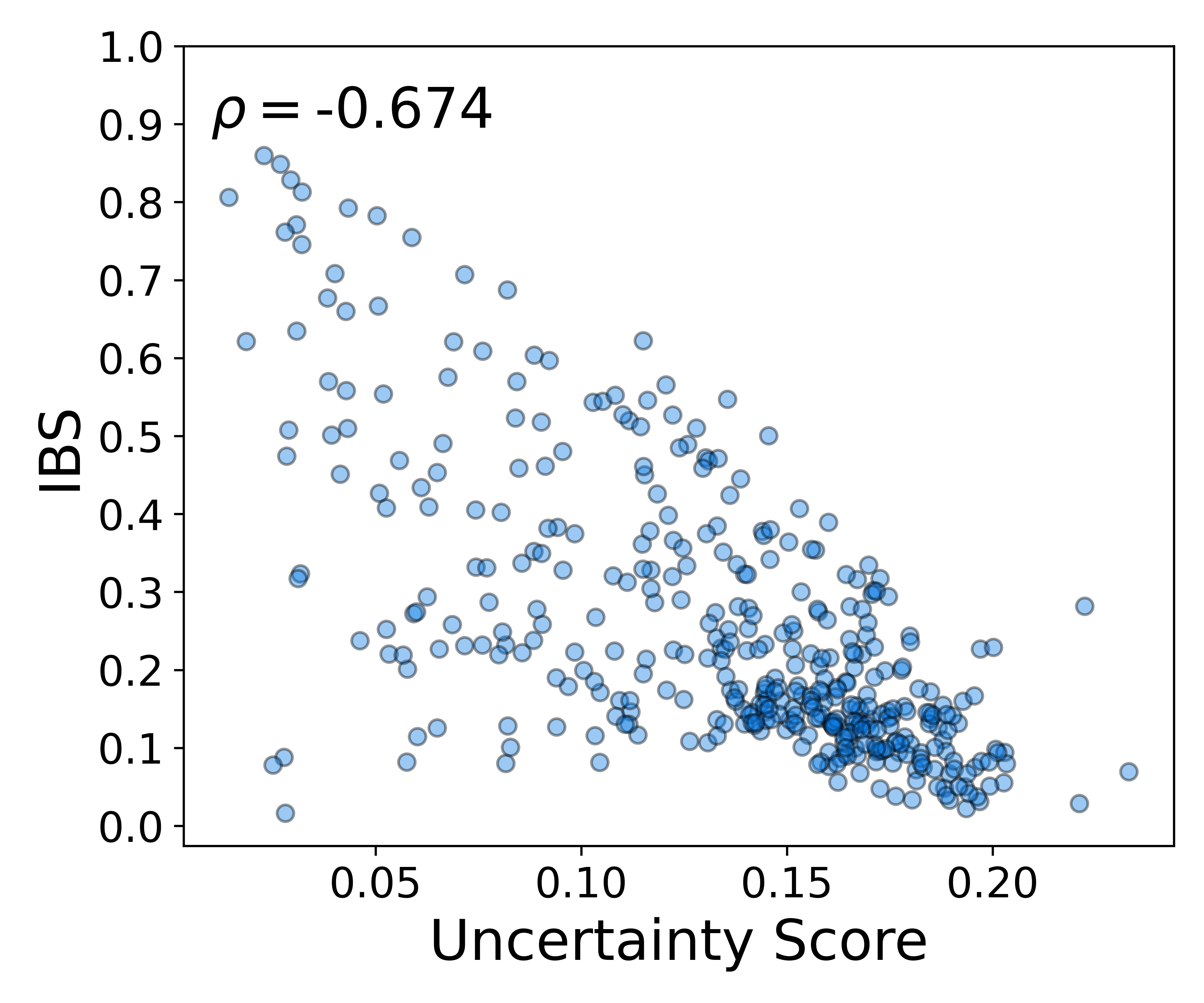}
        \caption{FLCHAIN}
    \end{subfigure}
    \begin{subfigure}[b]{0.232\textwidth}
        \centering
        \includegraphics[width=\textwidth]{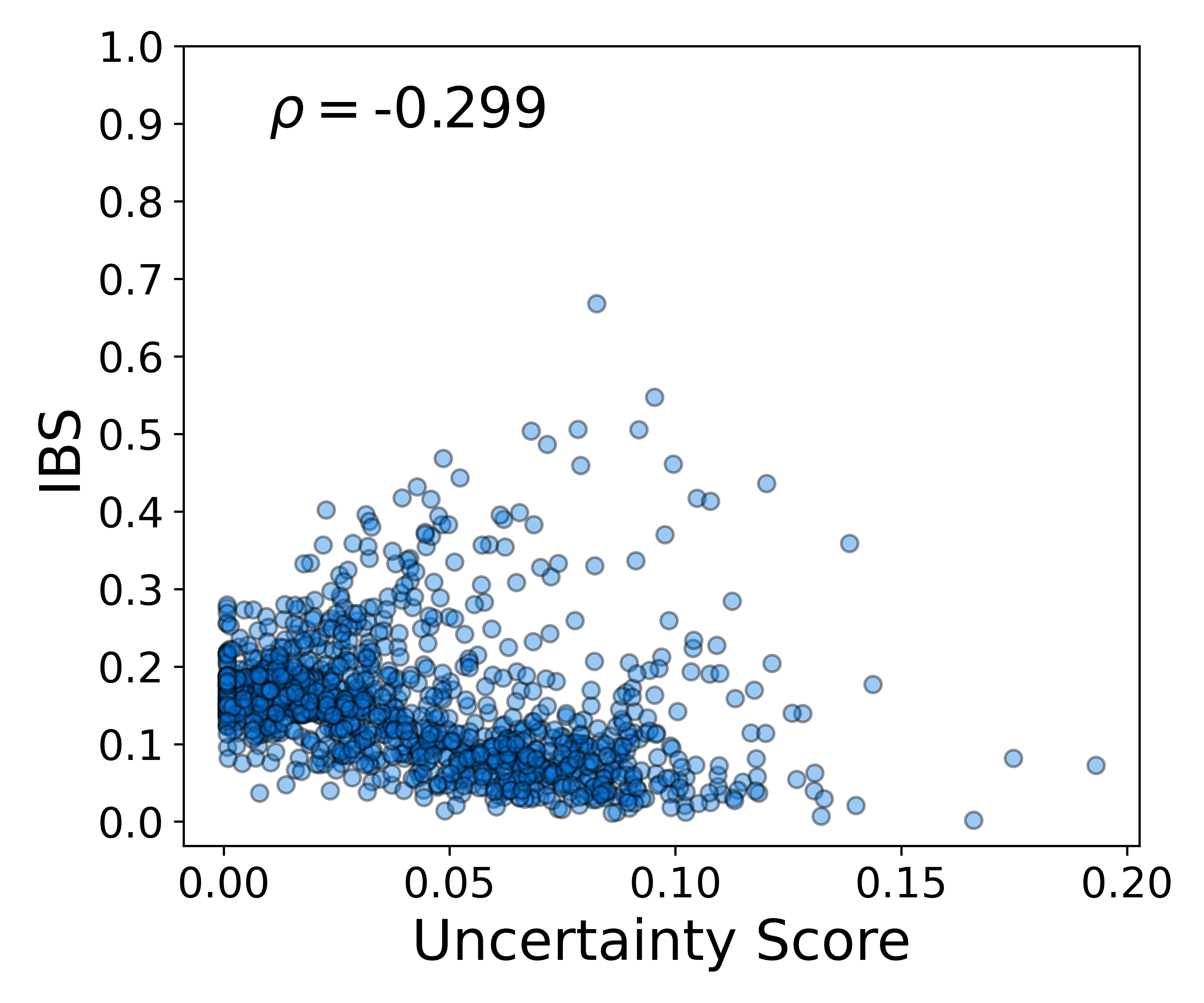}
        \caption{SUPPORT}
    \end{subfigure}
    \begin{subfigure}[b]{0.232\textwidth}
        \centering
        \includegraphics[width=\textwidth]{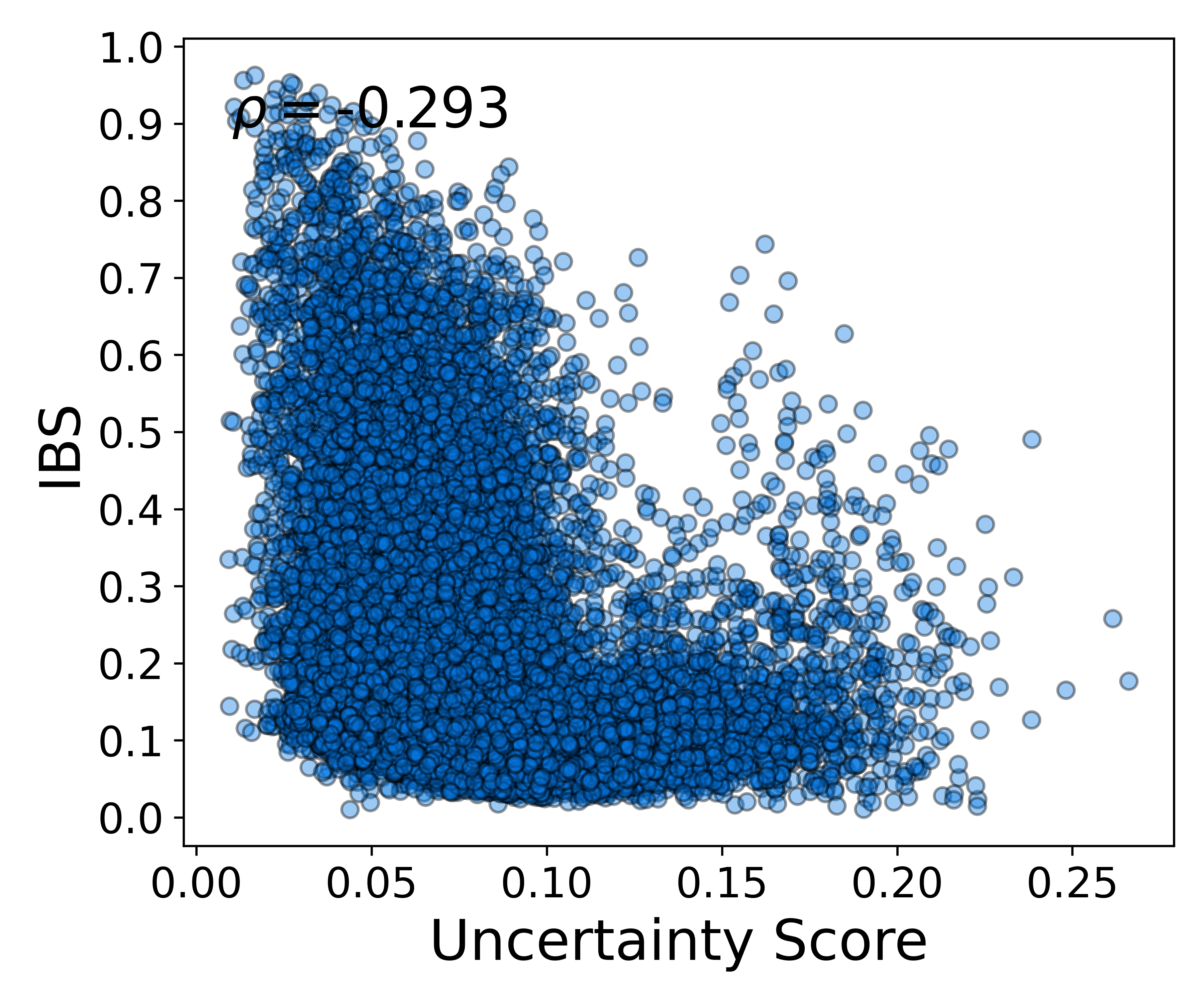}
        \caption{SEER-BC}
    \end{subfigure}
    \begin{subfigure}[b]{0.232\textwidth}
        \centering
        \includegraphics[width=\textwidth]{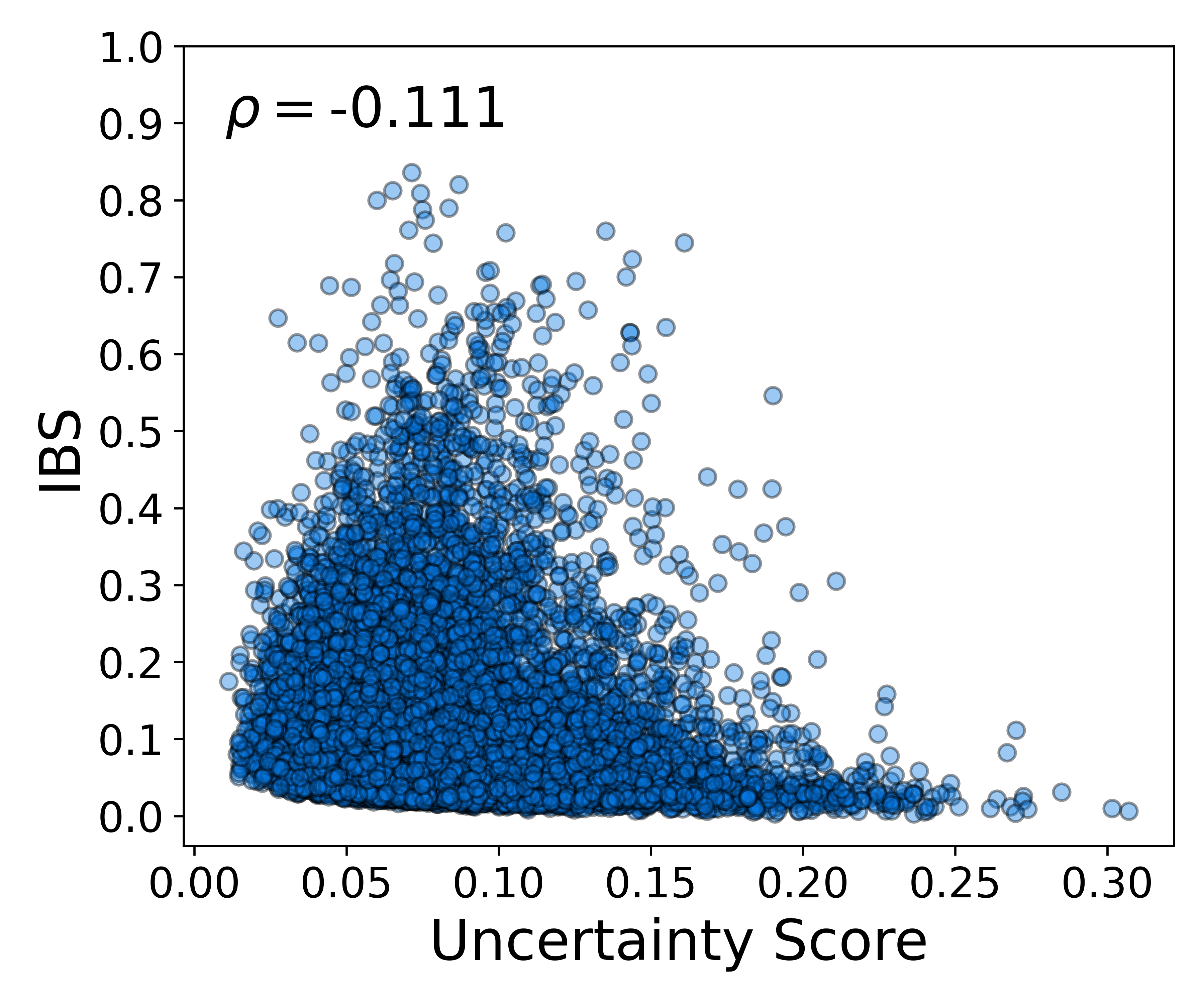}
        \caption{SAC3}
    \end{subfigure}
    \vspace{-10px}
    \caption{Predicted uncertainty scores versus IBSs from DeepSurv quantified by MC-Dropout across samples on (a) FLCHAIN, (b) SUPPORT, (c) SEER-BC and (d) SAC3 datasets.}
    \label{fig:misprediction_scatter_mcdropout}
\end{figure*}

\begin{figure*}[htbp]
    \centering
    \begin{subfigure}[b]{0.232\textwidth}
        \centering
        \includegraphics[width=\textwidth]{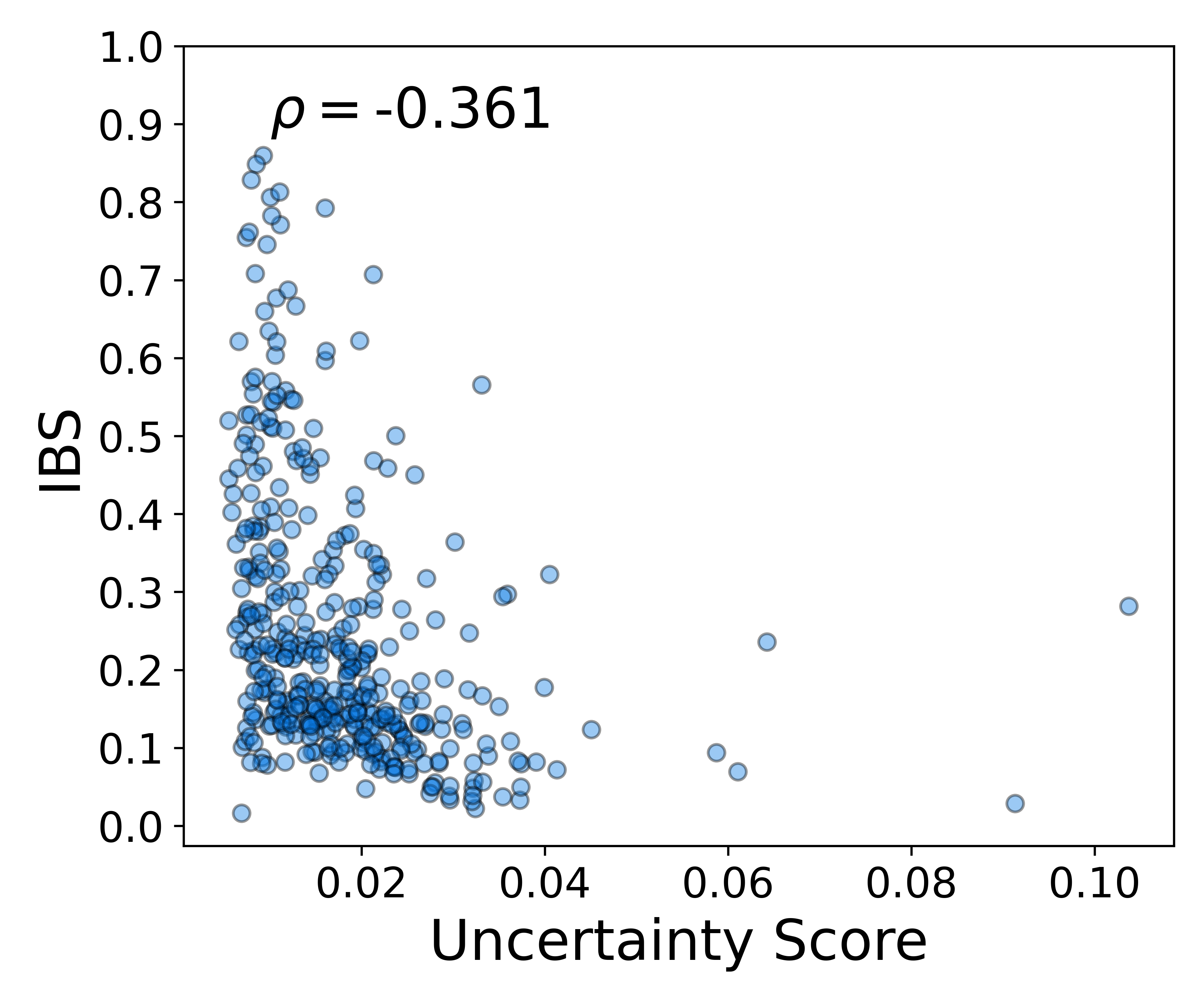}
        \caption{FLCHAIN}
    \end{subfigure}
    \begin{subfigure}[b]{0.232\textwidth}
        \centering
        \includegraphics[width=\textwidth]{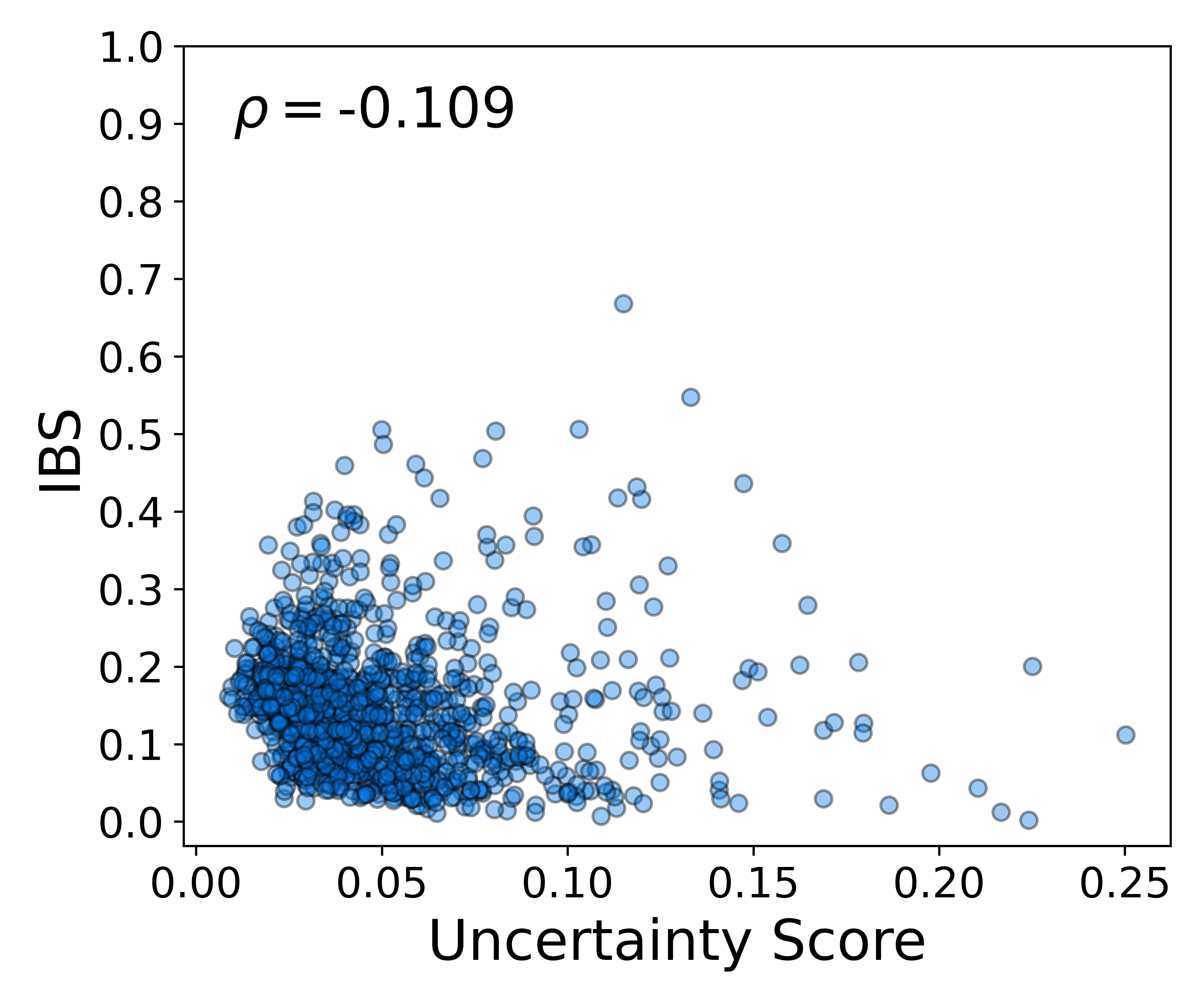}
        \caption{SUPPORT}
    \end{subfigure}
    \begin{subfigure}[b]{0.232\textwidth}
        \centering
        \includegraphics[width=\textwidth]{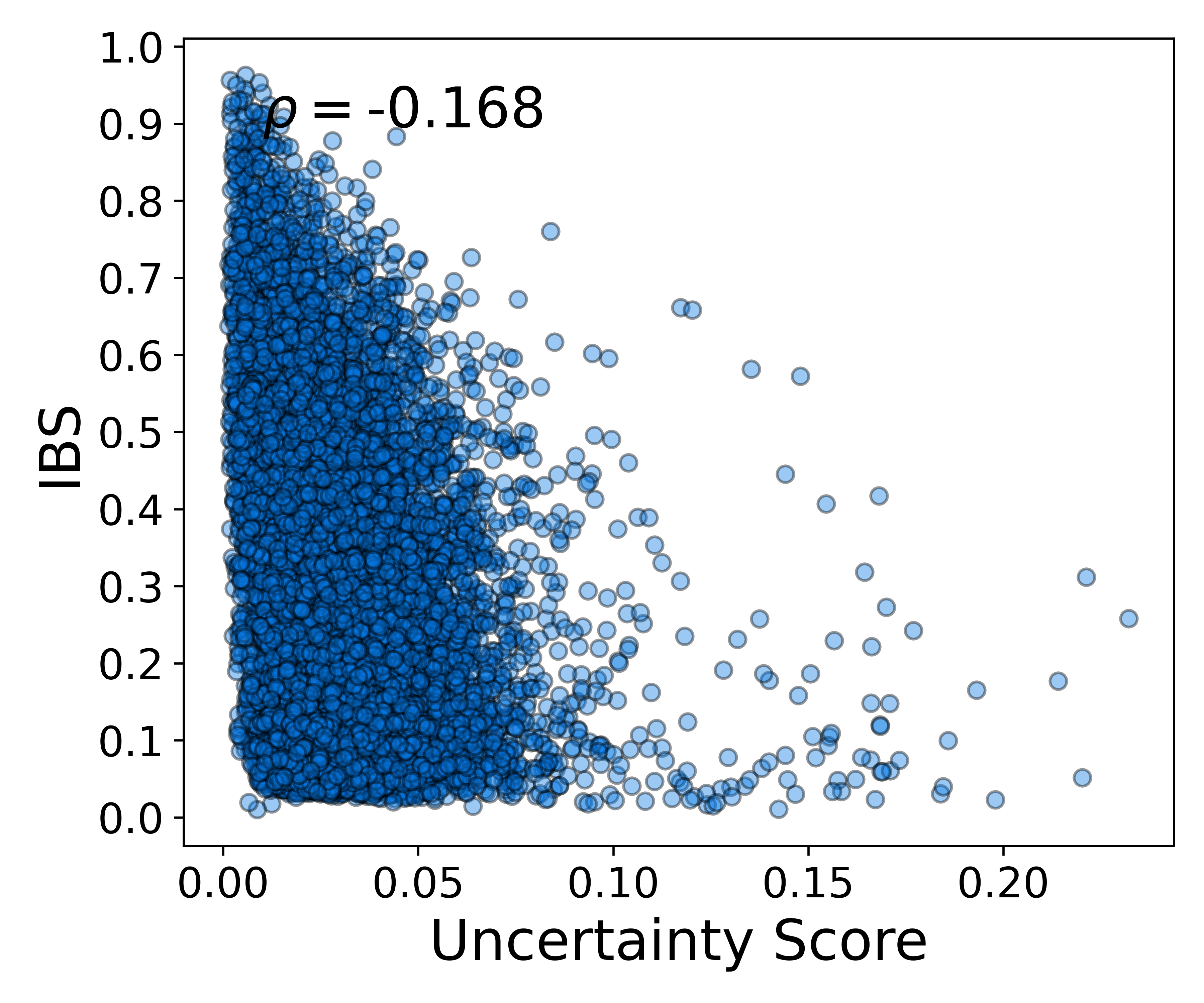}
        \caption{SEER-BC}
    \end{subfigure}
    \begin{subfigure}[b]{0.232\textwidth}
        \centering
        \includegraphics[width=\textwidth]{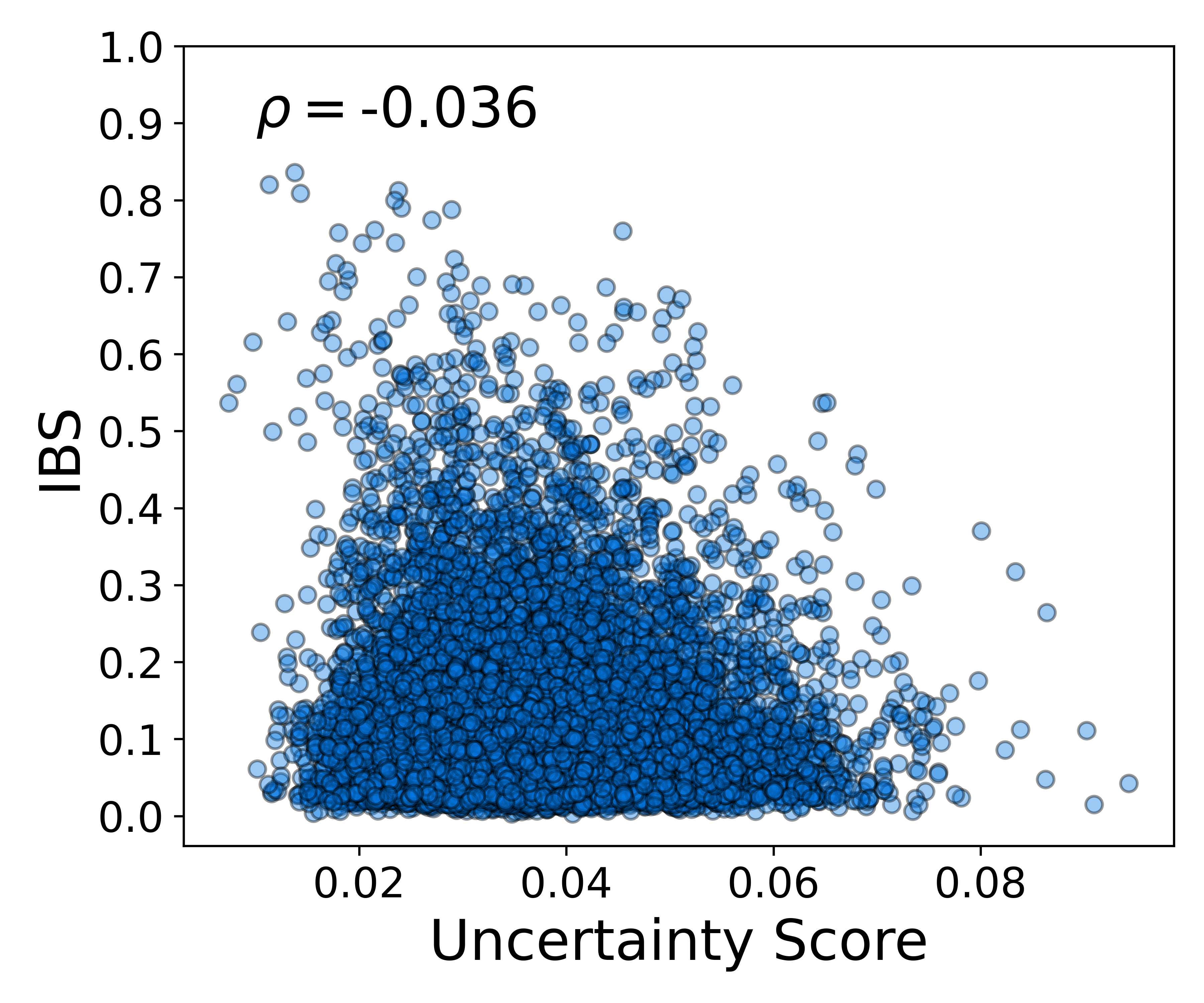}
        \caption{SAC3}
    \end{subfigure}
    \vspace{-10px}
    \caption{Predicted uncertainty scores versus IBSs from DeepSurv quantified by Ensemble across samples on (a) FLCHAIN, (b) SUPPORT, (c) SEER-BC and (d) SAC3 datasets.}
    \label{fig:misprediction_scatter_ensemble}
\end{figure*}

We present the misprediction results of BNNSurv model in Table~\ref{tab:misprediction_bnnsurv}. It can be observed our proposed SurvUnc framework outperforms the inherent uncertainty estimation in Bayesian-based survival model, BNNSurv. Besides, the comparison between predicted uncertainty scores and IBSs for the DeepSurv model using MC-Dropout and Ensemble are shown in Figure~\ref{fig:misprediction_scatter_mcdropout} and Figure~\ref{fig:misprediction_scatter_ensemble}, respectively. Similar to selective prediction experiments, we also present misprediction results with the absolute difference between predicted median survival and actual event times for uncensored patients, as shown in Table~\ref{tab:misprediction_time_diff}.

\begin{table}[htbp]
\centering
\def\arraystretch{1.}
\caption{Misprediction detection results with BNNSurv, i.e., Pearson correlation coefficient between uncertainty scores and IBSs of samples.}\label{tab:misprediction_bnnsurv}
\vspace{-10px}
\begin{tabular}{c|c|c|c|c}
\toprule
\textbf{Datasets} & \textbf{FLCHAIN} & \textbf{SUPPORT} & \textbf{SEER-BC} & \textbf{SAC3}  \\
\hline
Bayesian & -0.426  & 0.321& -0.680& 0.254  \\
\hline
SurvUnc-RF  & 0.440 & 0.776& 0.669 & 0.542\\ 
SurvUnc-MLP  & 0.410 & 0.749 & 0.629 & 0.619  \\
\bottomrule
\end{tabular}
\end{table}

\subsection{Hyperparameter Study}
\begin{figure*}[htbp]
    \centering
    \begin{subfigure}[b]{0.33\textwidth}
        \centering
        \includegraphics[width=\textwidth]{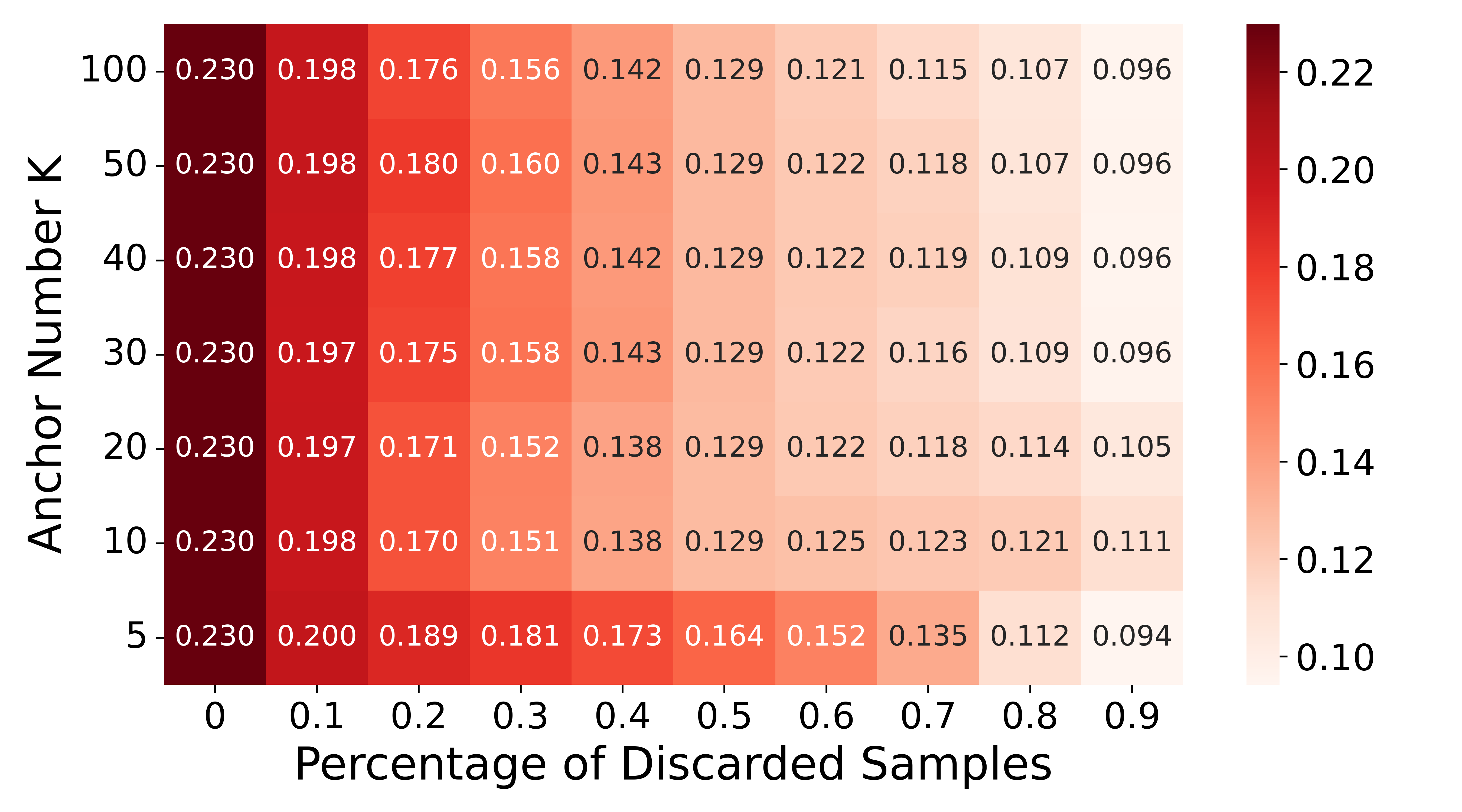}
        \caption{SurvUnc-RF (IBS)}
    \end{subfigure}
    \begin{subfigure}[b]{0.33\textwidth}
        \centering
        \includegraphics[width=\textwidth]{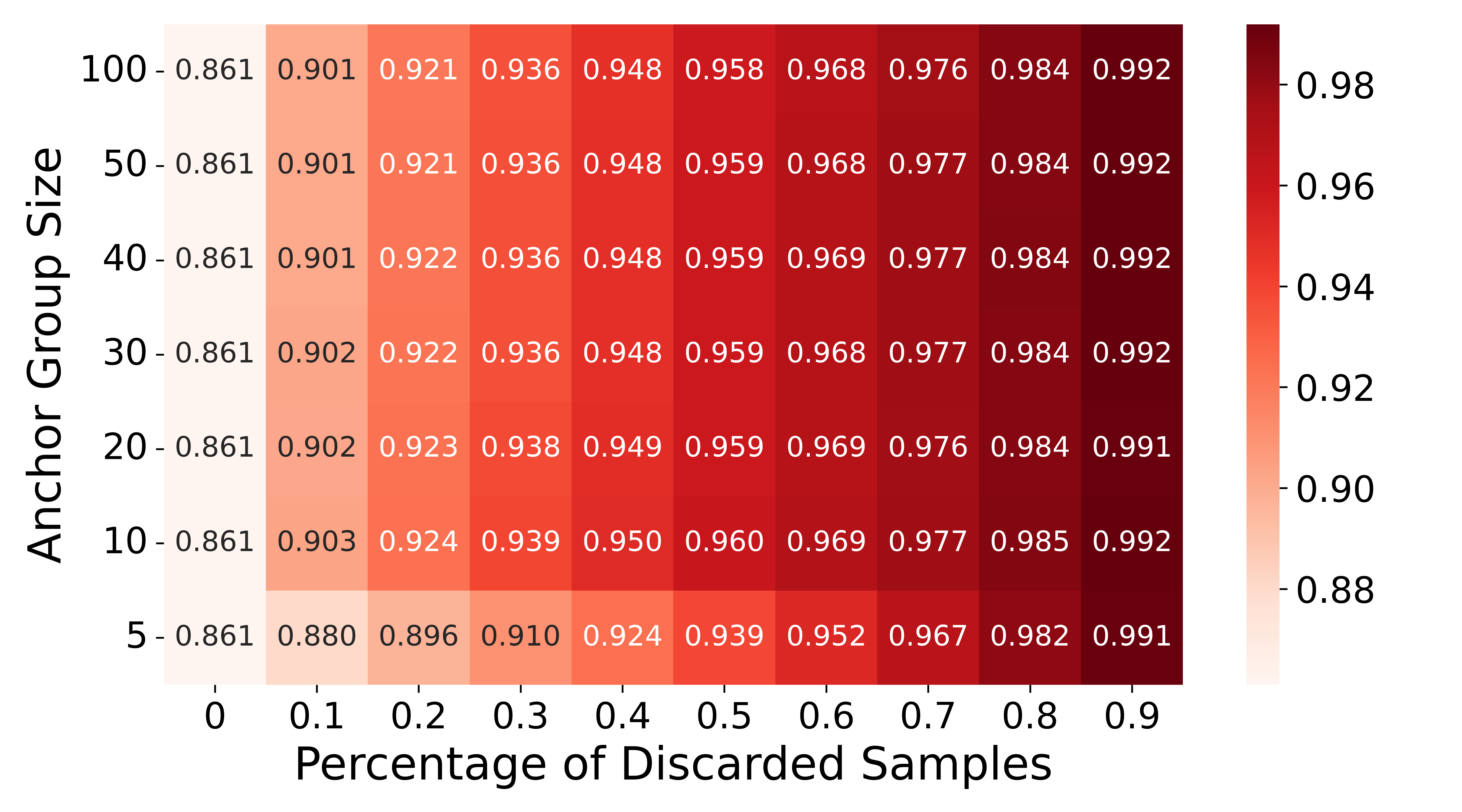}
        \caption{SurvUnc-MLP ($C^\text{td}$)}
    \end{subfigure}
    \begin{subfigure}[b]{0.33\textwidth}
        \centering
        \includegraphics[width=\textwidth]{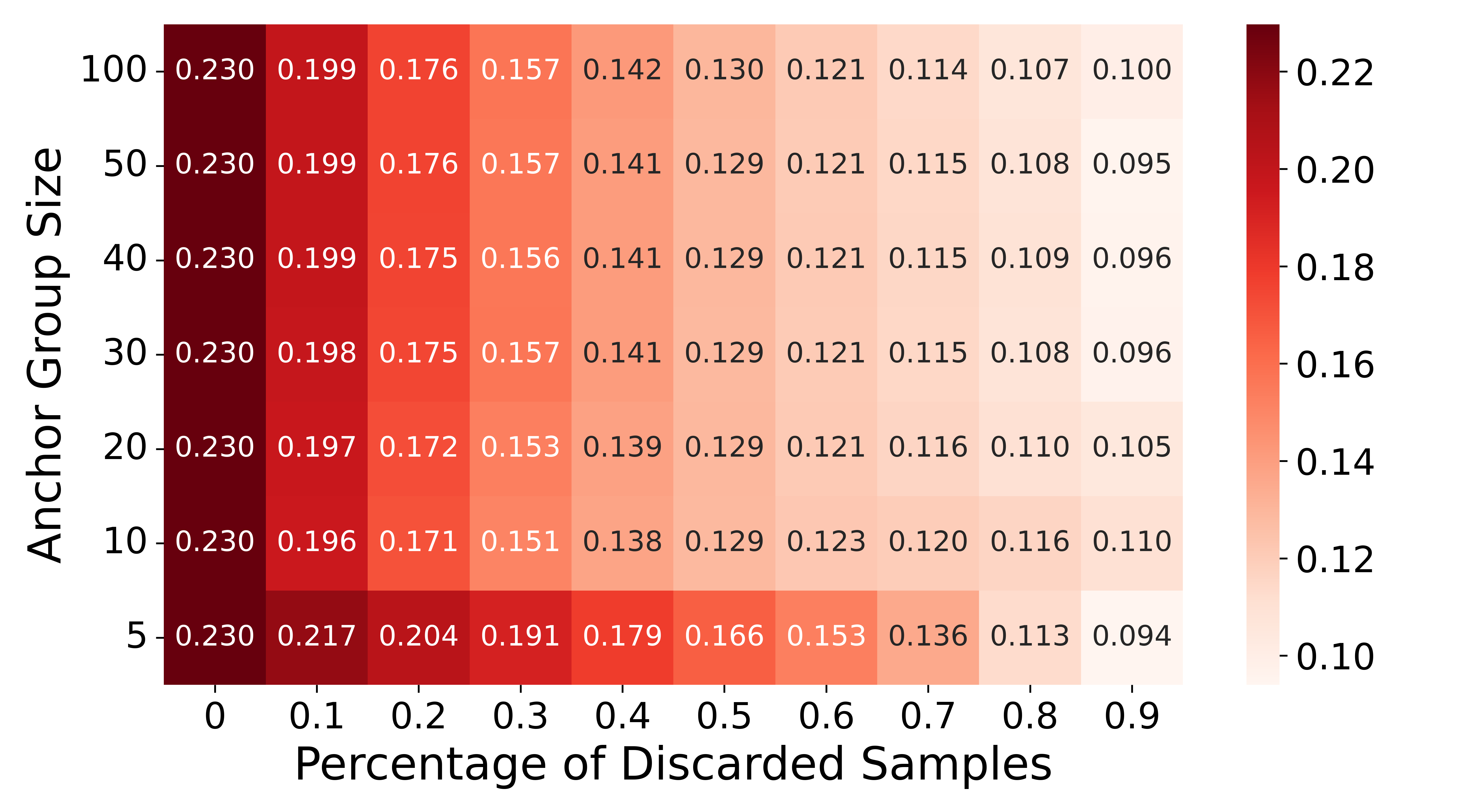}
        \caption{SurvUnc-MLP (IBS)}
    \end{subfigure}

    \vspace{-10px}
    \caption{Varying anchor number $K$ on selective prediction performance (a) IBS: SurvUnc-RF, (b) $C^\text{td}$: SurvUnc-MLP and (c) IBS: SurvUnc-MLP with DeepSurv on SEER-BC.}
    \label{fig:anchor_app}
\end{figure*}

For the impact of anchor number to the uncertainty quantification in Figure~\ref{fig:param_1_selective}, we further present its impact with IBS in Figure~\ref{fig:anchor_app}(a). Moreover, the corresponding studies for SurvUnc-MLP are presented in Figure~\ref{fig:anchor_app}(b) and Figure~\ref{fig:anchor_app}(c) for $C^\text{td}$ and IBS, respectively. The results demonstrate that the SurvUnc framework is robust with respect to the number of anchors, allowing for flexible selection.

\subsection{Case Study}
To validate the robustness of the proposed SurvUnc framework, we also evaluate two state-of-the-art models TabPFN \cite{hollmann2025accurate} and TabNet \cite{arik2021tabnet} as meta models with selective prediction, as shown in Table|\ref{tab:case_meta_model}. While TabPFN achieves marginally better results, the improvements are minimal considering the substantial computational and implementation overhead.

\begin{table}[h]
\centering
\def\arraystretch{1.}
\caption{$C^{\text{td}}$ of DeepSurv under different discarding percentages (10\%, 30\%, 50\%), determined by the SurvUnc framework with different meta models across datasets.}\label{tab:case_meta_model}
\vspace{-10px}
\begin{tabular}{c|ccc|ccc}
\toprule
{\textbf{Datasets}} & \multicolumn{3}{c|}{\textbf{SUPPORT}} & \multicolumn{3}{c}{\textbf{SEER-BC}}   \\
\hline
{\textbf{Meta Models}} & 10\% & 30\% & 50\% & 10\% & 30\% & 50\% \\
\hline
RF & 0.635 & 0.690 & 0.757 & 0.904 & 0.938 & 0.961 \\
MLP & 0.637 & 0.695 & 0.762 & 0.904 & 0.938 & 0.961 \\
TabNet & 0.639 & 0.695 & 0.762 & 0.904 & 0.938 & 0.961\\
TabPFN & 0.638 & 0701 & 0.768 & 0.903 & 0.934 & 0.959 \\
\bottomrule
\end{tabular}
\end{table}

\begin{table*}[htbp]
    \centering
    \def\arraystretch{1.}
    \setlength\tabcolsep{5.5pt}
    \caption{$C^{\text{td}}$ of survival models under different discarding percentages (10\%, 30\%, 50\%), determined by different uncertainty quantification methods across datasets. The best results are in bold. The last row of each group shows relative improvement compared with the best baseline by 100 runs of experiments. $p$-value$<0.001$ is true for all results without $^*$.}\label{tab:selective_cindex_app}
    \vspace{-10px}
   \rotatebox{0}{
    \begin{tabular}{c|c|ccc|ccc|ccc|ccc}
        \toprule
        \multirow{2}{*}{\makecell{\textbf{Survival}\\ \textbf{Models}}} & \multirow{2}{*}{\textbf{UQ Methods}} & \multicolumn{3}{c|}{\textbf{FLCHAIN}} & \multicolumn{3}{c|}{\textbf{SUPPORT}} & \multicolumn{3}{c|}{\textbf{SEER-BC}} & \multicolumn{3}{c}{\textbf{SAC3}} \\
        \cline{3-14}
        & & 10\% & 30\% & 50\% & 10\% & 30\% & 50\% & 10\% & 30\% & 50\% & 10\% & 30\% & 50\% \\
        \hline
        \multirow{5}{*}{DeepSurv} & MC-Dropout  & \makecell{0.782\\ \footnotesize${\pm 0.048}$}  & \makecell{0.745\\ \footnotesize${\pm 0.061}$}  & \makecell{0.685\\ \footnotesize${\pm 0.081}$}  & \makecell{0.600\\ \footnotesize${\pm 0.038}$}  & \makecell{0.576\\ \footnotesize${\pm 0.040}$}  & \makecell{0.563\\ \footnotesize${\pm 0.048}$}  & \makecell{0.857\\ \footnotesize${\pm 0.040}$}  & \makecell{0.846\\ \footnotesize${\pm 0.051}$}  & \makecell{0.832\\ \footnotesize${\pm 0.065}$}  & \makecell{0.772\\ \footnotesize${\pm 0.034}$}  & \makecell{0.778\\ \footnotesize${\pm 0.040}$}  & \makecell{0.793\\ \footnotesize${\pm 0.042}$}    \\
        & Ensemble & \makecell{0.786\\ \footnotesize${\pm 0.047}$}  & \makecell{0.755\\ \footnotesize${\pm 0.059}$}  & \makecell{0.710\\ \footnotesize${\pm 0.074}$}  & \makecell{0.607\\ \footnotesize${\pm 0.038}$}  & \makecell{0.598\\ \footnotesize${\pm 0.040}$}  & \makecell{0.576\\ \footnotesize${\pm 0.048}$}  & \makecell{0.863\\ \footnotesize${\pm 0.041}$}  & \makecell{0.857\\ \footnotesize${\pm 0.050}$}  & \makecell{0.842\\ \footnotesize${\pm 0.067}$}  & \makecell{0.783\\ \footnotesize${\pm 0.033}$}  & \makecell{0.795\\ \footnotesize${\pm 0.035}$}  & \makecell{0.807\\ \footnotesize${\pm 0.040}$}  \\ 
        & SurvUnc-RF & \makecell{\textbf{0.856}\\ \footnotesize${\pm 0.036}$}  & \makecell{\textbf{0.907}\\ \footnotesize${\pm 0.028}$}  & \makecell{\textbf{0.941}\\ \footnotesize${\pm 0.020}$}  & \makecell{0.635\\ \footnotesize${\pm 0.036}$}  & \makecell{0.690\\ \footnotesize${\pm 0.040}$}  & \makecell{0.757\\ \footnotesize${\pm 0.039}$}  & \makecell{\textbf{0.904}\\ \footnotesize${\pm 0.028}$}  & \makecell{\textbf{0.938}\\ \footnotesize${\pm 0.020}$}  & \makecell{\textbf{0.961}\\ \footnotesize${\pm 0.015}$}  & \makecell{0.792\\ \footnotesize${\pm 0.031}$}  & \makecell{0.822\\ \footnotesize${\pm 0.032}$}  & \makecell{0.855\\ \footnotesize${\pm 0.035}$}  \\
        & SurvUnc-MLP & \makecell{0.839\\ \footnotesize${\pm 0.042}$}  & \makecell{0.894\\ \footnotesize${\pm 0.034}$}  & \makecell{0.935\\ \footnotesize${\pm 0.026}$}  & \makecell{\textbf{0.637}\\ \footnotesize${\pm 0.037}$}  & \makecell{\textbf{0.695}\\ \footnotesize${\pm 0.041}$}  & \makecell{\textbf{0.762}\\ \footnotesize${\pm 0.041}$}  & \makecell{\textbf{0.904}\\ \footnotesize${\pm 0.029}$}  & \makecell{\textbf{0.938}\\ \footnotesize${\pm 0.020}$}  & \makecell{\textbf{0.961}\\ \footnotesize${\pm 0.014}$}  & \makecell{\textbf{0.797}\\ \footnotesize${\pm 0.032}$}  & \makecell{\textbf{0.830}\\ \footnotesize${\pm 0.033}$}  & \makecell{\textbf{0.862}\\ \footnotesize${\pm 0.030}$}    \\ \cline{2-14}
        & Improv. & 8.9\%  & 20.1\%  & 32.5\%  & 4.9\%  & 16.2\%  & 32.3\%  & 4.8\%  & 9.5\%  & 14.1\%  & 1.8\%  & 4.4\%  & 6.8\%\\
        \hline
        \multirow{5}{*}{DeepHit} & MC-Dropout & \makecell{0.784\\ \footnotesize${\pm 0.046}$}  & \makecell{0.762\\ \footnotesize${\pm 0.057}$}  & \makecell{0.733\\ \footnotesize${\pm 0.075}$}  & \makecell{\textbf{0.648}\\ \footnotesize${\pm 0.033}$}  & \makecell{0.657\\ \footnotesize${\pm 0.040}$}  & \makecell{0.656\\ \footnotesize${\pm 0.046}$}  & \makecell{0.908\\ \footnotesize${\pm 0.028}$}  & \makecell{0.937\\ \footnotesize${\pm 0.019}$}  & \makecell{0.953\\ \footnotesize${\pm 0.018}$}  & \makecell{0.797\\ \footnotesize${\pm 0.035}$}  & \makecell{0.811\\ \footnotesize${\pm 0.036}$}  & \makecell{0.836\\ \footnotesize${\pm 0.034}$}  \\
        & Ensemble  & \makecell{\textbf{0.855}\\ \footnotesize${\pm 0.035}$}  & \makecell{0.902\\ \footnotesize${\pm 0.027}$}  & \makecell{0.934\\ \footnotesize${\pm 0.022}$}  & \makecell{0.637\\ \footnotesize${\pm 0.034}$}  & \makecell{0.637\\ \footnotesize${\pm 0.038}$}  & \makecell{0.637\\ \footnotesize${\pm 0.047}$}  & \makecell{0.896\\ \footnotesize${\pm 0.033}$}  & \makecell{0.923\\ \footnotesize${\pm 0.029}$}  & \makecell{0.944\\ \footnotesize${\pm 0.023}$}  & \makecell{0.805\\ \footnotesize${\pm 0.033}$}  & \makecell{0.822\\ \footnotesize${\pm 0.030}$}  & \makecell{0.842\\ \footnotesize${\pm 0.034}$}  \\
        & SurvUnc-RF   & \makecell{\textbf{0.855}$^*$\\ \footnotesize${\pm 0.036}$}  & \makecell{0.908\\ \footnotesize${\pm 0.027}$}  & \makecell{\textbf{0.940}\\ \footnotesize${\pm 0.020}$}  & \makecell{{0.639}\\ \footnotesize${\pm 0.032}$}  & \makecell{\textbf{0.664}$^*$\\ \footnotesize${\pm 0.041}$}  & \makecell{\textbf{0.693}\\ \footnotesize${\pm 0.047}$}  & \makecell{0.913\\ \footnotesize${\pm 0.027}$}  & \makecell{\textbf{0.945}\\ \footnotesize${\pm 0.018}$}  & \makecell{\textbf{0.961}\\ \footnotesize${\pm 0.015}$}  & \makecell{0.811\\ \footnotesize${\pm 0.031}$}  & \makecell{0.838\\ \footnotesize${\pm 0.035}$}  & \makecell{0.866\\ \footnotesize${\pm 0.036}$}  \\
        & SurvUnc-MLP  & \makecell{\textbf{0.855}$^*$\\ \footnotesize${\pm 0.035}$}  & \makecell{\textbf{0.909}\\ \footnotesize${\pm 0.027}$}  & \makecell{\textbf{0.940}\\ \footnotesize${\pm 0.020}$}  & \makecell{{0.639}\\ \footnotesize${\pm 0.033}$}  & \makecell{0.660$^*$\\ \footnotesize${\pm 0.040}$}  & \makecell{0.692\\ \footnotesize${\pm 0.045}$}  & \makecell{\textbf{0.914}\\ \footnotesize${\pm 0.027}$}  & \makecell{\textbf{0.945}\\ \footnotesize${\pm 0.018}$}  & \makecell{\textbf{0.961}\\ \footnotesize${\pm 0.014}$}  & \makecell{\textbf{0.817}\\ \footnotesize${\pm 0.031}$}  & \makecell{\textbf{0.845}\\ \footnotesize${\pm 0.031}$}  & \makecell{\textbf{0.876}\\ \footnotesize${\pm 0.031}$}   \\ \cline{2-14}
        & Improv. & 0.0\%  & 0.8\%  & 0.6\%  & -1.4\%  & 1.1\%  & 5.6\%  & 0.7\%  & 0.9\%  & 0.8\%  & 1.5\%  & 2.8\%  & 4.0\%\\
        \hline
        \multirow{5}{*}{DSM} & MC-Dropout & \makecell{0.780\\ \footnotesize${\pm 0.049}$}  & \makecell{0.741\\ \footnotesize${\pm 0.062}$}  & \makecell{0.687\\ \footnotesize${\pm 0.079}$}  & \makecell{0.613\\ \footnotesize${\pm 0.035}$}  & \makecell{0.614\\ \footnotesize${\pm 0.041}$}  & \makecell{0.613\\ \footnotesize${\pm 0.049}$}  & \makecell{0.866\\ \footnotesize${\pm 0.037}$}  & \makecell{0.854\\ \footnotesize${\pm 0.045}$}  & \makecell{0.832\\ \footnotesize${\pm 0.057}$}  & \makecell{0.793\\ \footnotesize${\pm 0.033}$}  & \makecell{0.776\\ \footnotesize${\pm 0.040}$}  & \makecell{0.770\\ \footnotesize${\pm 0.051}$}  \\
        & Ensemble & \makecell{0.787\\ \footnotesize${\pm 0.047}$}  & \makecell{0.755\\ \footnotesize${\pm 0.059}$}  & \makecell{0.715\\ \footnotesize${\pm 0.075}$}  & \makecell{0.616\\ \footnotesize${\pm 0.036}$}  & \makecell{0.605\\ \footnotesize${\pm 0.044}$}  & \makecell{0.581\\ \footnotesize${\pm 0.054}$}  & \makecell{0.872\\ \footnotesize${\pm 0.036}$}  & \makecell{0.868\\ \footnotesize${\pm 0.043}$}  & \makecell{0.854\\ \footnotesize${\pm 0.058}$}  & \makecell{0.802\\ \footnotesize${\pm 0.034}$}  & \makecell{0.813\\ \footnotesize${\pm 0.037}$}  & \makecell{0.828\\ \footnotesize${\pm 0.035}$}   \\
        & SurvUnc-RF  & \makecell{\textbf{0.854}\\ \footnotesize${\pm 0.038}$}  & \makecell{\textbf{0.907}\\ \footnotesize${\pm 0.026}$}  & \makecell{\textbf{0.941}\\ \footnotesize${\pm 0.020}$}  & \makecell{0.640\\ \footnotesize${\pm 0.035}$}  & \makecell{\textbf{0.685}\\ \footnotesize${\pm 0.041}$}  & \makecell{0.725\\ \footnotesize${\pm 0.044}$}  & \makecell{\textbf{0.910}\\ \footnotesize${\pm 0.027}$}  & \makecell{\textbf{0.943}\\ \footnotesize${\pm 0.017}$}  & \makecell{\textbf{0.960}\\ \footnotesize${\pm 0.013}$}  & \makecell{0.813\\ \footnotesize${\pm 0.030}$}  & \makecell{0.837\\ \footnotesize${\pm 0.033}$}  & \makecell{0.862\\ \footnotesize${\pm 0.035}$}  \\
        & SurvUnc-MLP& \makecell{0.852\\ \footnotesize${\pm 0.038}$}  & \makecell{0.906\\ \footnotesize${\pm 0.028}$}  & \makecell{0.940\\ \footnotesize${\pm 0.020}$}  & \makecell{\textbf{0.641}\\ \footnotesize${\pm 0.034}$}  & \makecell{\textbf{0.685}\\ \footnotesize${\pm 0.039}$}  & \makecell{\textbf{0.731}\\ \footnotesize${\pm 0.043}$}  & \makecell{\textbf{0.910}\\ \footnotesize${\pm 0.027}$}  & \makecell{\textbf{0.943}\\ \footnotesize${\pm 0.018}$}  & \makecell{0.959\\ \footnotesize${\pm 0.014}$}  & \makecell{\textbf{0.818}\\ \footnotesize${\pm 0.029}$}  & \makecell{\textbf{0.844}\\ \footnotesize${\pm 0.031}$}  & \makecell{\textbf{0.871}\\ \footnotesize${\pm 0.031}$}   \\  \cline{2-14}
        & Improv. & 8.5\%  & 20.1\%  & 31.6\%  & 4.1\%  & 11.6\%  & 19.2\%  & 4.4\%  & 8.6\%  & 12.4\%  & 2.0\%  & 3.8\%  & 5.2\%\\
        \hline
        \multirow{5}{*}{RSF} & MC-Dropout & -& -& -&- & -& -& -&- &- &- &- & -\\
        & Ensemble    & \makecell{0.790\\ \footnotesize${\pm 0.049}$}  & \makecell{0.777\\ \footnotesize${\pm 0.062}$}  & \makecell{0.745\\ \footnotesize${\pm 0.079}$}  & \makecell{0.648\\ \footnotesize${\pm 0.037}$}  & \makecell{0.662\\ \footnotesize${\pm 0.043}$}  & \makecell{0.684\\ \footnotesize${\pm 0.043}$}  & \makecell{0.878\\ \footnotesize${\pm 0.034}$}  & \makecell{0.874\\ \footnotesize${\pm 0.042}$}  & \makecell{0.863\\ \footnotesize${\pm 0.057}$}  & \makecell{0.649\\ \footnotesize${\pm 0.043}$}  & \makecell{0.663\\ \footnotesize${\pm 0.045}$}  & \makecell{0.677\\ \footnotesize${\pm 0.050}$}    \\
        & SurvUnc-RF & \makecell{\textbf{0.840}\\ \footnotesize${\pm 0.044}$}  & \makecell{\textbf{0.897}\\ \footnotesize${\pm 0.029}$}  & \makecell{\textbf{0.933}\\ \footnotesize${\pm 0.023}$}  & \makecell{\textbf{0.663}\\ \footnotesize${\pm 0.037}$}  & \makecell{\textbf{0.708}\\ \footnotesize${\pm 0.037}$}  & \makecell{\textbf{0.750}\\ \footnotesize${\pm 0.041}$}  & \makecell{0.908\\ \footnotesize${\pm 0.026}$}  & \makecell{0.941\\ \footnotesize${\pm 0.019}$}  & \makecell{0.959\\ \footnotesize${\pm 0.015}$}  & \makecell{\textbf{0.658}\\ \footnotesize${\pm 0.044}$}  & \makecell{\textbf{0.692}\\ \footnotesize${\pm 0.048}$}  & \makecell{\textbf{0.738}\\ \footnotesize${\pm 0.053}$}   \\
        & SurvUnc-MLP  & \makecell{0.820\\ \footnotesize${\pm 0.044}$}  & \makecell{0.854\\ \footnotesize${\pm 0.043}$}  & \makecell{0.892\\ \footnotesize${\pm 0.041}$}  & \makecell{0.656\\ \footnotesize${\pm 0.036}$}  & \makecell{0.689\\ \footnotesize${\pm 0.038}$}  & \makecell{0.721\\ \footnotesize${\pm 0.040}$}  & \makecell{\textbf{0.913}\\ \footnotesize${\pm 0.026}$}  & \makecell{\textbf{0.943}\\ \footnotesize${\pm 0.019}$}  & \makecell{\textbf{0.960}\\ \footnotesize${\pm 0.014}$}  & \makecell{0.653\\ \footnotesize${\pm 0.045}$}  & \makecell{0.685\\ \footnotesize${\pm 0.050}$}  & \makecell{0.731\\ \footnotesize${\pm 0.053}$}  \\ \cline{2-14}
        & Improv. & 6.3\%  & 15.4\%  & 25.2\%  & 2.3\%  & 6.9\%  & 9.6\%  & 4.0\%  & 7.9\%  & 11.2\%      & 1.4\%  & 4.4\%  & 9.0\%\\

        \hline
        \multirow{4}{*}{BNNSurv} & Bayesian & \makecell{\textbf{0.773}\\ \footnotesize${\pm 0.049}$}  & \makecell{{0.732}\\ \footnotesize${\pm 0.063}$}  & \makecell{{0.670}\\ \footnotesize${\pm 0.082}$}  & \makecell{{0.623}\\ \footnotesize${\pm 0.036}$}  & \makecell{{0.648}\\ \footnotesize${\pm 0.039}$}  & \makecell{{0.687}\\ \footnotesize${\pm 0.044}$}  & \makecell{0.847\\ \footnotesize${\pm 0.044}$}  & \makecell{0.836\\ \footnotesize${\pm 0.055}$}  & \makecell{0.805\\ \footnotesize${\pm 0.076}$}  & \makecell{{0.719}\\ \footnotesize${\pm 0.037}$}  & \makecell{{0.740}\\ \footnotesize${\pm 0.040}$}  & \makecell{{0.771}\\ \footnotesize${\pm 0.043}$}   \\
        & SurvUnc-RF & \makecell{\textbf{0.848}\\ \footnotesize${\pm 0.039}$}  & \makecell{\textbf{0.902}\\ \footnotesize${\pm 0.029}$}  & \makecell{\textbf{0.936}\\ \footnotesize${\pm 0.023}$}  & \makecell{\textbf{0.650}\\ \footnotesize${\pm 0.033}$}  & \makecell{\textbf{0.701}\\ \footnotesize${\pm 0.037}$}  & \makecell{\textbf{0.764}\\ \footnotesize${\pm 0.038}$}  & \makecell{0.891\\ \footnotesize${\pm 0.034}$}  & \makecell{\textbf{0.932}\\ \footnotesize${\pm 0.022}$}  & \makecell{\textbf{0.955}\\ \footnotesize${\pm 0.016}$}  & \makecell{\textbf{0.727}\\ \footnotesize${\pm 0.036}$}  & \makecell{{0.765}\\ \footnotesize${\pm 0.034}$}  & \makecell{{0.808}\\ \footnotesize${\pm 0.037}$}   \\
        & SurvUnc-MLP  & \makecell{0.843\\ \footnotesize${\pm 0.040}$}  & \makecell{0.899\\ \footnotesize${\pm 0.030}$}  & \makecell{0.934\\ \footnotesize${\pm 0.023}$}  & \makecell{0.648\\ \footnotesize${\pm 0.034}$}  & \makecell{0.699\\ \footnotesize${\pm 0.037}$}  & \makecell{0.760\\ \footnotesize${\pm 0.038}$}  & \makecell{\textbf{0.893}\\ \footnotesize${\pm 0.033}$}  & \makecell{\textbf{0.932}\\ \footnotesize${\pm 0.023}$}  & \makecell{\textbf{0.955}\\ \footnotesize${\pm 0.016}$}  & \makecell{\textbf{0.727}\\ \footnotesize${\pm 0.036}$}  & \makecell{\textbf{0.767}\\ \footnotesize${\pm 0.037}$}  & \makecell{\textbf{0.815}\\ \footnotesize${\pm 0.038}$}  \\ \cline{2-14}
        & Improv. & 9.7\%  & 23.2\%  & 39.7\%  & 4.3\%  & 8.2\%  & 11.2\%  & 5.4\%  & 11.5\%  & 18.6\%  & 1.1\%  & 3.6\%  & 5.7\%\\
        \bottomrule
    \end{tabular}}
\end{table*}

\begin{table*}[htbp]
    \centering
    \def\arraystretch{1.}
    \setlength\tabcolsep{5pt}
    \caption{IBS of survival models under different discarding percentages (10\%, 30\%, 50\%), determined by different uncertainty quantification methods across datasets. The best results are in bold. The last row of each group shows relative improvement compared with the best baseline by 100 runs of experiments. $p$-value$<0.001$ is true for all results without $^*$.}\label{tab:selective_ibs_app}
    \vspace{-10px}
    \rotatebox{0}{
    \begin{tabular}{c|c|ccc|ccc|ccc|ccc}
        \hline
        \multirow{2}{*}{\makecell{\textbf{Survival}\\ \textbf{Models}}} & \multirow{2}{*}{\textbf{UQ Methods}} & \multicolumn{3}{c|}{\textbf{FLCHAIN}} & \multicolumn{3}{c|}{\textbf{SUPPORT}} & \multicolumn{3}{c|}{\textbf{SEER-BC}} & \multicolumn{3}{c}{\textbf{SAC3}} \\
        \cline{3-14}
        & & 10\% & 30\% & 50\% & 10\% & 30\% & 50\% & 10\% & 30\% & 50\% & 10\% & 30\% & 50\% \\
        \hline
        \multirow{5}{*}{DeepSurv} & MC-Dropout   & \makecell{0.268\\ \footnotesize${\pm 0.038}$}  & \makecell{0.302\\ \footnotesize${\pm 0.046}$}  & \makecell{0.344\\ \footnotesize${\pm 0.057}$}  & \makecell{0.151\\ \footnotesize${\pm 0.010}$}  & \makecell{0.164\\ \footnotesize${\pm 0.010}$}  & \makecell{0.169\\ \footnotesize${\pm 0.009}$}  & \makecell{0.240\\ \footnotesize${\pm 0.042}$}  & \makecell{0.259\\ \footnotesize${\pm 0.047}$}  & \makecell{0.282\\ \footnotesize${\pm 0.057}$}  & \makecell{0.120\\ \footnotesize${\pm 0.014}$}  & \makecell{0.123\\ \footnotesize${\pm 0.016}$}  & \makecell{0.122\\ \footnotesize${\pm 0.017}$}    \\
        & Ensemble  & \makecell{0.265\\ \footnotesize${\pm 0.037}$}  & \makecell{0.289\\ \footnotesize${\pm 0.043}$}  & \makecell{0.318\\ \footnotesize${\pm 0.051}$}  & \makecell{0.147\\ \footnotesize${\pm 0.009}$}  & \makecell{0.154\\ \footnotesize${\pm 0.009}$}  & \makecell{0.161\\ \footnotesize${\pm 0.010}$}  & \makecell{0.232\\ \footnotesize${\pm 0.040}$}  & \makecell{0.242\\ \footnotesize${\pm 0.048}$}  & \makecell{0.257\\ \footnotesize${\pm 0.058}$}  & \makecell{0.115\\ \footnotesize${\pm 0.014}$}  & \makecell{0.115\\ \footnotesize${\pm 0.017}$}  & \makecell{0.119\\ \footnotesize${\pm 0.022}$}  \\
        & SurvUnc-RF  & \makecell{\textbf{0.219}\\ \footnotesize${\pm 0.032}$}  & \makecell{\textbf{0.184}\\ \footnotesize${\pm 0.033}$}  & \makecell{\textbf{0.152}\\ \footnotesize${\pm 0.029}$}  & \makecell{\textbf{0.134}\\ \footnotesize${\pm 0.009}$}  & \makecell{\textbf{0.118}\\ \footnotesize${\pm 0.010}$}  & \makecell{\textbf{0.100}\\ \footnotesize${\pm 0.012}$}  & \makecell{0.200\\ \footnotesize${\pm 0.038}$}  & \makecell{0.160\\ \footnotesize${\pm 0.029}$}  & \makecell{\textbf{0.134}\\ \footnotesize${\pm 0.027}$}  & \makecell{0.104\\ \footnotesize${\pm 0.013}$}  & \makecell{0.090\\ \footnotesize${\pm 0.012}$}  & \makecell{0.080\\ \footnotesize${\pm 0.014}$}  \\
        & SurvUnc-MLP  & \makecell{0.226\\ \footnotesize${\pm 0.034}$}  & \makecell{0.189\\ \footnotesize${\pm 0.033}$}  & \makecell{0.161\\ \footnotesize${\pm 0.032}$}  & \makecell{\textbf{0.134}\\ \footnotesize${\pm 0.009}$}  & \makecell{\textbf{0.118}\\ \footnotesize${\pm 0.011}$}  & \makecell{\textbf{0.100}\\ \footnotesize${\pm 0.012}$}  & \makecell{\textbf{0.198}\\ \footnotesize${\pm 0.039}$}  & \makecell{\textbf{0.159}\\ \footnotesize${\pm 0.030}$}  & \makecell{\textbf{0.134}\\ \footnotesize${\pm 0.026}$}  & \makecell{\textbf{0.098}\\ \footnotesize${\pm 0.012}$}  & \makecell{\textbf{0.077}\\ \footnotesize${\pm 0.011}$}  & \makecell{\textbf{0.062}\\ \footnotesize${\pm 0.011}$}  \\  \cline{2-14}
        & Improv. & 17.4\%  & 36.3\%  & 52.2\%  & 8.8\%  & 23.4\%  & 37.9\%  & 14.7\%  & 34.3\%  & 47.9\%  & 14.8\%  & 33.0\%  & 47.9\%  \\
        \hline
        \multirow{5}{*}{DeepHit} & MC-Dropout & \makecell{0.300\\ \footnotesize${\pm 0.039}$}  & \makecell{\textbf{0.295}\\ \footnotesize${\pm 0.046}$}  & \makecell{0.305\\ \footnotesize${\pm 0.057}$}  & \makecell{\textbf{0.158}\\ \footnotesize${\pm 0.005}$}  & \makecell{0.153\\ \footnotesize${\pm 0.005}$}  & \makecell{0.150\\ \footnotesize${\pm 0.006}$}  & \makecell{0.279\\ \footnotesize${\pm 0.030}$}  & \makecell{\textbf{0.262}\\ \footnotesize${\pm 0.029}$}  & \makecell{\textbf{0.240}\\ \footnotesize${\pm 0.029}$}  & \makecell{0.123\\ \footnotesize${\pm 0.008}$}  & \makecell{0.125\\ \footnotesize${\pm 0.008}$}  & \makecell{0.125\\ \footnotesize${\pm 0.009}$}   \\
        & Ensemble    & \makecell{\textbf{0.294}\\ \footnotesize${\pm 0.033}$}  & \makecell{0.298\\ \footnotesize${\pm 0.034}$}  & \makecell{\textbf{0.300}\\ \footnotesize${\pm 0.038}$}  & \makecell{\textbf{0.158}\\ \footnotesize${\pm 0.005}$}  & \makecell{\textbf{0.151}\\ \footnotesize${\pm 0.005}$}  & \makecell{0.146\\ \footnotesize${\pm 0.006}$}  & \makecell{0.280\\ \footnotesize${\pm 0.030}$}  & \makecell{\textbf{0.262}\\ \footnotesize${\pm 0.030}$}  & \makecell{0.242\\ \footnotesize${\pm 0.031}$}  & \makecell{0.119\\ \footnotesize${\pm 0.008}$}  & \makecell{0.120\\ \footnotesize${\pm 0.009}$}  & \makecell{0.120\\ \footnotesize${\pm 0.009}$}   \\
        & SurvUnc-RF & \makecell{0.296$^*$\\ \footnotesize${\pm 0.034}$}  & \makecell{0.298$^*$\\ \footnotesize${\pm 0.038}$}  & \makecell{0.303$^*$\\ \footnotesize${\pm 0.044}$}  & \makecell{0.159\\ \footnotesize${\pm 0.005}$}  & \makecell{0.152$^*$\\ \footnotesize${\pm 0.006}$}  & \makecell{\textbf{0.143}\\ \footnotesize${\pm 0.007}$}  & \makecell{0.278$^*$\\ \footnotesize${\pm 0.029}$}  & \makecell{0.269\\ \footnotesize${\pm 0.025}$}  & \makecell{0.271\\ \footnotesize${\pm 0.028}$}  & \makecell{0.115\\ \footnotesize${\pm 0.008}$}  & \makecell{0.109\\ \footnotesize${\pm 0.009}$}  & \makecell{0.102\\ \footnotesize${\pm 0.010}$}   \\
        & SurvUnc-MLP & \makecell{0.299$^*$\\ \footnotesize${\pm 0.035}$}  & \makecell{0.300$^*$\\ \footnotesize${\pm 0.039}$}  & \makecell{0.303$^*$\\ \footnotesize${\pm 0.040}$}  & \makecell{0.160\\ \footnotesize${\pm 0.005}$}  & \makecell{0.152$^*$\\ \footnotesize${\pm 0.006}$}  & \makecell{0.145\\ \footnotesize${\pm 0.007}$}  & \makecell{\textbf{0.276}$^*$\\ \footnotesize${\pm 0.029}$}  & \makecell{0.267\\ \footnotesize${\pm 0.025}$}  & \makecell{0.269\\ \footnotesize${\pm 0.026}$}  & \makecell{\textbf{0.112}\\ \footnotesize${\pm 0.008}$}  & \makecell{\textbf{0.101}\\ \footnotesize${\pm 0.009}$}  & \makecell{\textbf{0.092}\\ \footnotesize${\pm 0.010}$}   \\
         \cline{2-14}
        & Improv. & -0.7\%  & -1.0\%  & -1.0\%  & -0.6\%  & -0.7\%  & 2.1\%  & 1.1\%  & -1.9\%  & -12.1\%  & 5.9\%  & 15.8\%  & 23.3\% \\
        \hline
        \multirow{5}{*}{DSM} & MC-Dropout& \makecell{0.260\\ \footnotesize${\pm 0.038}$}  & \makecell{0.298\\ \footnotesize${\pm 0.046}$}  & \makecell{0.339\\ \footnotesize${\pm 0.054}$}  & \makecell{0.150\\ \footnotesize${\pm 0.009}$}  & \makecell{0.145\\ \footnotesize${\pm 0.010}$}  & \makecell{0.143\\ \footnotesize${\pm 0.011}$}  & \makecell{0.238\\ \footnotesize${\pm 0.046}$}  & \makecell{0.269\\ \footnotesize${\pm 0.058}$}  & \makecell{0.320\\ \footnotesize${\pm 0.070}$}  & \makecell{0.124\\ \footnotesize${\pm 0.017}$}  & \makecell{0.139\\ \footnotesize${\pm 0.018}$}  & \makecell{0.147\\ \footnotesize${\pm 0.019}$}   \\
        & Ensemble & \makecell{0.255\\ \footnotesize${\pm 0.035}$}  & \makecell{0.280\\ \footnotesize${\pm 0.042}$}  & \makecell{0.312\\ \footnotesize${\pm 0.056}$}  & \makecell{0.147\\ \footnotesize${\pm 0.010}$}  & \makecell{0.155\\ \footnotesize${\pm 0.012}$}  & \makecell{0.161\\ \footnotesize${\pm 0.014}$}  & \makecell{0.223\\ \footnotesize${\pm 0.045}$}  & \makecell{0.230\\ \footnotesize${\pm 0.054}$}  & \makecell{0.243\\ \footnotesize${\pm 0.068}$}  & \makecell{0.116\\ \footnotesize${\pm 0.017}$}  & \makecell{0.117\\ \footnotesize${\pm 0.020}$}  & \makecell{0.120\\ \footnotesize${\pm 0.022}$}   \\
        & SurvUnc-RF & \makecell{\textbf{0.208}\\ \footnotesize${\pm 0.030}$}  & \makecell{\textbf{0.172}\\ \footnotesize${\pm 0.027}$}  & \makecell{\textbf{0.143}\\ \footnotesize${\pm 0.023}$}  & \makecell{\textbf{0.135}\\ \footnotesize${\pm 0.010}$}  & \makecell{\textbf{0.118}\\ \footnotesize${\pm 0.011}$}  & \makecell{\textbf{0.103}\\ \footnotesize${\pm 0.013}$}  & \makecell{\textbf{0.190}\\ \footnotesize${\pm 0.041}$}  & \makecell{\textbf{0.147}\\ \footnotesize${\pm 0.030}$}  & \makecell{\textbf{0.132}\\ \footnotesize${\pm 0.028}$}  & \makecell{0.104\\ \footnotesize${\pm 0.015}$}  & \makecell{0.090\\ \footnotesize${\pm 0.016}$}  & \makecell{0.081\\ \footnotesize${\pm 0.017}$}   \\
        & SurvUnc-MLP & \makecell{0.211\\ \footnotesize${\pm 0.032}$}  & \makecell{0.175\\ \footnotesize${\pm 0.029}$}  & \makecell{0.144\\ \footnotesize${\pm 0.024}$}  & \makecell{\textbf{0.135}\\ \footnotesize${\pm 0.010}$}  & \makecell{0.120\\ \footnotesize${\pm 0.011}$}  & \makecell{0.107\\ \footnotesize${\pm 0.013}$}  & \makecell{\textbf{0.190}\\ \footnotesize${\pm 0.040}$}  & \makecell{0.149\\ \footnotesize${\pm 0.029}$}  & \makecell{\textbf{0.132}\\ \footnotesize${\pm 0.027}$}  & \makecell{\textbf{0.095}\\ \footnotesize${\pm 0.014}$}  & \makecell{\textbf{0.075}\\ \footnotesize${\pm 0.012}$}  & \makecell{\textbf{0.064}\\ \footnotesize${\pm 0.012}$}   \\
         \cline{2-14}
        & Improv. & 18.4\%  & 38.6\%  & 54.2\%  & 8.2\%  & 18.6\%  & 28.0\%  & 14.8\%  & 36.1\%  & 45.7\%  & 18.1\%  & 35.9\%  & 46.7\% \\
        \hline
        \multirow{5}{*}{RSF} & MC-Dropout & -& -& -&- & -& -& -&- &- &- &- & -\\
        & Ensemble & \makecell{0.254\\ \footnotesize${\pm 0.039}$}  & \makecell{0.261\\ \footnotesize${\pm 0.046}$}  & \makecell{0.281\\ \footnotesize${\pm 0.053}$}  & \makecell{0.145\\ \footnotesize${\pm 0.011}$}  & \makecell{0.136\\ \footnotesize${\pm 0.012}$}  & \makecell{0.130\\ \footnotesize${\pm 0.014}$}  & \makecell{0.217\\ \footnotesize${\pm 0.040}$}  & \makecell{0.225\\ \footnotesize${\pm 0.050}$}  & \makecell{0.243\\ \footnotesize${\pm 0.062}$}  & \makecell{0.144\\ \footnotesize${\pm 0.010}$}  & \makecell{0.142\\ \footnotesize${\pm 0.012}$}  & \makecell{0.138\\ \footnotesize${\pm 0.013}$}    \\
        & SurvUnc-RF & \makecell{\textbf{0.220}\\ \footnotesize${\pm 0.037}$}  & \makecell{\textbf{0.178}\\ \footnotesize${\pm 0.032}$}  & \makecell{\textbf{0.145}\\ \footnotesize${\pm 0.026}$}  & \makecell{\textbf{0.136}\\ \footnotesize${\pm 0.010}$}  & \makecell{\textbf{0.121}\\ \footnotesize${\pm 0.010}$}  & \makecell{\textbf{0.111}\\ \footnotesize${\pm 0.013}$}  & \makecell{0.189\\ \footnotesize${\pm 0.036}$}  & \makecell{0.152\\ \footnotesize${\pm 0.030}$}  & \makecell{\textbf{0.127}\\ \footnotesize${\pm 0.025}$}  & \makecell{\textbf{0.139}\\ \footnotesize${\pm 0.010}$}  & \makecell{\textbf{0.128}\\ \footnotesize${\pm 0.011}$}  & \makecell{\textbf{0.118}\\ \footnotesize${\pm 0.012}$}  \\
        & SurvUnc-MLP & \makecell{0.239\\ \footnotesize${\pm 0.038}$}  & \makecell{0.216\\ \footnotesize${\pm 0.042}$}  & \makecell{0.191\\ \footnotesize${\pm 0.044}$}  & \makecell{0.139\\ \footnotesize${\pm 0.010}$}  & \makecell{0.127\\ \footnotesize${\pm 0.012}$}  & \makecell{0.120\\ \footnotesize${\pm 0.013}$}  & \makecell{\textbf{0.180}\\ \footnotesize${\pm 0.037}$}  & \makecell{\textbf{0.146}\\ \footnotesize${\pm 0.027}$}  & \makecell{\textbf{0.127}\\ \footnotesize${\pm 0.025}$}  & \makecell{0.141\\ \footnotesize${\pm 0.011}$}  & \makecell{0.131\\ \footnotesize${\pm 0.011}$}  & \makecell{0.120\\ \footnotesize${\pm 0.012}$}   \\
         \cline{2-14}
        & Improv. & 13.4\%  & 31.8\%  & 48.4\%  & 6.2\%  & 11.0\%  & 14.6\%  & 17.1\%  & 35.1\%  & 47.7\%  & 3.5\%  & 9.9\%  & 14.5\%\\
        \hline

        \multirow{4}{*}{BNNSurv} & Bayesian & \makecell{{0.327}\\ \footnotesize${\pm 0.044}$}  & \makecell{{0.348}\\ \footnotesize${\pm 0.054}$}  & \makecell{{0.375}\\ \footnotesize${\pm 0.065}$}  & \makecell{{0.144}\\ \footnotesize${\pm 0.009}$}  & \makecell{{0.137}\\ \footnotesize${\pm 0.010}$}  & \makecell{{0.127}\\ \footnotesize${\pm 0.013}$}  & \makecell{0.268\\ \footnotesize${\pm 0.047}$}  & \makecell{0.297\\ \footnotesize${\pm 0.059}$}  & \makecell{0.345\\ \footnotesize${\pm 0.080}$}  & \makecell{{0.150}\\ \footnotesize${\pm 0.015}$}  & \makecell{{0.144}\\ \footnotesize${\pm 0.018}$}  & \makecell{{0.133}\\ \footnotesize${\pm 0.023}$}   \\
        
        & SurvUnc-RF & \makecell{\textbf{0.289}\\ \footnotesize${\pm 0.038}$}  & \makecell{\textbf{0.266}\\ \footnotesize${\pm 0.038}$}  & \makecell{{0.253}\\ \footnotesize${\pm 0.036}$}  & \makecell{\textbf{0.133}\\ \footnotesize${\pm 0.008}$}  & \makecell{\textbf{0.113}\\ \footnotesize${\pm 0.009}$}  & \makecell{\textbf{0.095}\\ \footnotesize${\pm 0.009}$}  & \makecell{0.229\\ \footnotesize${\pm 0.043}$}  & \makecell{0.179\\ \footnotesize${\pm 0.032}$}  & \makecell{0.154\\ \footnotesize${\pm 0.029}$}  & \makecell{{0.139}\\ \footnotesize${\pm 0.013}$}  & \makecell{{0.121}\\ \footnotesize${\pm 0.012}$}  & \makecell{{0.105}\\ \footnotesize${\pm 0.013}$}   \\
        
        & SurvUnc-MLP  & \makecell{0.292\\ \footnotesize${\pm 0.040}$}  & \makecell{0.268\\ \footnotesize${\pm 0.036}$}  & \makecell{\textbf{0.251}\\ \footnotesize${\pm 0.035}$}  & \makecell{\textbf{0.133}\\ \footnotesize${\pm 0.008}$}  & \makecell{0.114\\ \footnotesize${\pm 0.009}$}  & \makecell{\textbf{0.095}\\ \footnotesize${\pm 0.010}$}  & \makecell{\textbf{0.225}\\ \footnotesize${\pm 0.042}$}  & \makecell{\textbf{0.178}\\ \footnotesize${\pm 0.032}$}  & \makecell{\textbf{0.152}\\ \footnotesize${\pm 0.027}$}  & \makecell{\textbf{0.137}\\ \footnotesize${\pm 0.013}$}  & \makecell{\textbf{0.114}\\ \footnotesize${\pm 0.011}$}  & \makecell{\textbf{0.095}\\ \footnotesize${\pm 0.012}$}  \\ \cline{2-14}
        & Improv. & 11.6\%  & 23.6\%  & 33.1\%  & 7.6\%  & 0.175\%  & 25.2\%  & 16.0\%  & 40.1\%  & 55.9\%  & 8.7\%  & 20.8\%  & 28.6\%\\
        \bottomrule
    \end{tabular}}
\end{table*}

\begin{table*}[htbp]
\centering
\def\arraystretch{1.}
\caption{Misprediction detection results, i.e., Pearson correlation coefficient between uncertainty scores and the absolute difference between predicted median survival and actual event times of samples.}\label{tab:misprediction_time_diff}
\vspace{-10px}
\begin{tabular}{c|cccc|cccc}
\toprule
{\textbf{Datasets}} &  \multicolumn{4}{c|}{\textbf{SUPPORT}} & \multicolumn{4}{c}{\textbf{SEER-BC}} \\
\hline
{\textbf{UQ Methods}} & DeepSurv & DeepHit & DSM & RSF  & DeepSurv  & DeepHit  & DSM & RSF   \\
\hline
MC-Dropout &  -0.136 & -0.097 & -0.008 &  -& -0.245 & 0.209 & -0.677 & - \\ \hline
Ensemble & -0.030 & -0.130 & -0.209 & 0.150 & -0.076 & 0.267 & -0.079 & -0.120  \\
\hline
SurvUnc-RF &  0.479 & 0.028 & 0.415 & 0.503 & 0.643 &0.027 & 0.626 & 0.536 \\ 
SurvUnc-MLP & 0.450 & 0.012 & 0.370 & 0.373 & 0.647 & 0.032 & 0.625 & 0.586\\
\bottomrule
\end{tabular}
\end{table*}

\end{document}